\PassOptionsToPackage{normalem}{ulem}
\documentclass[]{arxiv_template}
\usepackage{natbib}
\setcitestyle{square, comma, numbers, sort&compress}

\usepackage[T1]{fontenc}
\usepackage{lmodern}
\usepackage{tikz}
\usetikzlibrary{positioning,arrows.meta}
\usepackage[table]{xcolor}
\usepackage{booktabs}
\usepackage{hhline}
\usepackage{array}
\usepackage{tabularx}
\usepackage{multirow}
\usepackage{makecell}
\usepackage{amssymb}
\usepackage{caption}
\usepackage{longtable}
\usepackage{pdflscape}
\usepackage{booktabs}
\usepackage{array}
\usepackage{ragged2e}
\usepackage{multirow}
\usepackage{svg}
\usepackage[table]{xcolor}
\hypersetup{hidelinks}
\usepackage{fontawesome5}

\definecolor{Qone}{HTML}{9ECAE1}
\definecolor{Qtwo}{HTML}{6BAED6}
\definecolor{Qthree}{HTML}{4292C6}
\definecolor{Qfour}{HTML}{2171B5}
\definecolor{Qfive}{HTML}{084594}

\definecolor{HeaderBlue}{HTML}{E8F1F8}
\definecolor{GroupBlue}{HTML}{F2F7FB}
\definecolor{RuleBlue}{HTML}{AFC6D8}
\definecolor{CitationColor}{HTML}{172B4D}
\definecolor{MutedText}{HTML}{4F5B66}

\newcolumntype{Y}{>{\centering\arraybackslash\hspace{0pt}}X}
\newcolumntype{C}[1]{>{\centering\arraybackslash}p{#1}}

\newcommand{\cstar}[1]{%
  \textcolor{#1}{\raisebox{-0.06ex}{\ensuremath{\scriptstyle\bigstar}}}%
}

\newcommand{\starjoin}{\ensuremath{\mkern-5mu}}

\newcommand{\qstarsone}{%
  \cstar{Qone}%
}

\newcommand{\qstarstwo}{%
  \cstar{Qtwo}\starjoin
  \cstar{Qtwo}%
}

\newcommand{\qstarsthree}{%
  \cstar{Qthree}\starjoin
  \cstar{Qthree}\starjoin
  \cstar{Qthree}%
}

\newcommand{\qstarsfour}{%
  \cstar{Qfour}\starjoin
  \cstar{Qfour}\starjoin
  \cstar{Qfour}\starjoin
  \cstar{Qfour}%
}

\newcommand{\qstarsfive}{%
  \cstar{Qfive}\starjoin
  \cstar{Qfive}\starjoin
  \cstar{Qfive}\starjoin
  \cstar{Qfive}\starjoin
  \cstar{Qfive}%
}

\newcommand{\capabilitystars}[1]{%
  \ifcase#1\relax
  \or\qstarsone
  \or\qstarstwo
  \or\qstarsthree
  \or\qstarsfour
  \or\qstarsfive
  \fi
}

\newcommand{\levelhead}[2]{%
  \makecell[c]{%
    \textbf{#1}\\[-0.08em]
    \scriptsize\color{MutedText}#2%
  }%
}

\newcommand{\groupname}[2]{%
  \makecell[c]{%
    \textbf{\color{CitationColor}#1}\\[-0.08em]
    \textbf{\color{CitationColor}#2}%
  }%
}

\newcommand{\groupnamethree}[3]{%
  \makecell[c]{%
    \textbf{\color{CitationColor}#1}\\[-0.08em]
    \textbf{\color{CitationColor}#2}\\[-0.08em]
    \textbf{\color{CitationColor}#3}%
  }%
}

\newcommand{\techrule}{%
  \arrayrulecolor{white}%
  \hhline{->{\arrayrulecolor{RuleBlue!68}}------}%
  \arrayrulecolor{black}%
}

\newcommand{\techruleblue}{%
  \arrayrulecolor{GroupBlue}%
  \hhline{->{\arrayrulecolor{RuleBlue!68}}------}%
  \arrayrulecolor{black}%
}

\usepackage{amsmath,amsfonts,amsthm}
\usepackage{graphicx}
\usepackage{booktabs}
\usepackage{multirow}
\usepackage{array}
\usepackage{enumitem}
\usepackage{xcolor}
\usepackage{hyperref}
\usepackage{balance}
\usepackage{etoolbox}
\usepackage{capt-of}
\usepackage{tabularx}
\usepackage[most]{tcolorbox}

\usepackage[table]{xcolor}
\usepackage{hhline}
\usepackage{makecell}
\usepackage{amssymb}
\usepackage{caption}
\hypersetup{hidelinks}
\usepackage{fix-cm}

\setlist[itemize]{leftmargin=0.6cm,topsep=1pt,itemsep=0pt}
\setlist[enumerate]{leftmargin=0.6cm,topsep=1pt,itemsep=0pt}
\renewcommand{\arraystretch}{1.25}
\newcommand{\todo}[1]{\textcolor{red}{[TODO: #1]}}

\definecolor{github_purple}{rgb}{0.302,0.165,0.498}

\newcommand{\zxh}[1]{\textcolor{blue}{#1}}

\definecolor{wykcolor}{RGB}{170,95,0}
\newcommand{\wyk}[1]{\textcolor{black}{#1}} 

\newcommand{\LoneStep}[1]{%
  \fcolorbox{black!35}{white}{%
    \rule{0pt}{2.2ex}\hspace{3pt}\footnotesize #1\hspace{3pt}}%
}

\newcommand{\subsep}{%
    \arrayrulecolor{gray!20}\hline
    \arrayrulecolor{black}
}

\newcommand{\bfit}[1]{\textbf{\textit{#1}}}
\newcommand{\RSI}{{RSI}}
\newcommand{\AISystem}{AI system}
\newcommand{\Improver}{\emph{improver}}
\newcommand{\Verifier}{\emph{verifier}}

\titleformat*{\paragraph}{\bfseries}

\title{The Last AI Built by Humans: Toward Genuine Recursive Self-Improvement}

\makeatletter
\gdef\authorlist{%
  \begin{center}
    \sffamily\normalsize
    Yi Duan$^{1,2*}$, Ying Liu$^{1,2*}$, Zirui Tang$^{1,2*}$, Haodong Chen$^{1,2*}$, Jun Zhou$^{1,2*}$, Yumou Liu$^{1,2}$, Bangrui Xu$^{1,2}$, Yukai Wu$^{1,2}$, Sidi Chen$^{1,2}$, Yuhan Zhou$^{1,2}$, Haoyu Wang$^{1,2}$, Xiaoyou Yu$^{1,2}$, Shaokun Han$^{1,2}$,  Xuzhou Zhu$^{1,2}$, Le Zhou$^{1,2}$, Bolin Lu$^{1,2}$, Wei Zhou$^{1,2}$, Jiachen Liu$^{9}$, Nuozhou Fang$^{2}$, Jiaxin Tian$^{2}$, 
    Ruoyu Chen$^{4}$, 
    Yuxuan Li$^{5}$, 
    Kai Zuo$^{6}$, Kaiyan Zhang$^{10}$, 
    Qianyu Yang$^{8}$, Zijie Wang$^{8}$,     
    Jiantao Qiu$^{7}$, Conghui He$^{7}$, 
    Guoliang Li$^{3}$, Bowen Zhou$^{7,3}$, Zhiyuan Liu$^{3}$,\\ Zhoufutu Wen$^{8}$, Jihua Kang$^{4}$, Xuanhe Zhou$^{1,2\dagger}$, Fan Wu$^{1,2}$ 
  \end{center}
}
\makeatother

\makeatletter
\gdef\affiliationlist{%
  \begin{center}
    \affiliationfont\sffamily
    $^{1}$Shanghai Jiao Tong University
    $^{2}$Theseus Labs
    $^{3}$Tsinghua University\\    
    $^{4}$ByteDance 
    $^{5}$ModelBest
    $^{6}$Super Intelligence Team, Xiaohongshu Inc.
    $^{7}$Shanghai AI Lab\\
    $^{8}$Humanlaya
    $^{9}$Agent-Native Research Lab
    $^{10}$Frontis.AI
  \end{center}
}
\makeatother

\abstract{Recursive self-improvement (RSI) enables AI systems to turn experience and feedback into persistent changes that improve both their capabilities and the process of future improvement. We first use the Headroom-Closed Index (HCI) to reveal the problems of existing LLMs, then introduce the RSI concept and its development roadmap: from improvement-execution autonomy, improvement-strategy autonomy, experience-acquisition autonomy, and environment-adaptation autonomy, to recursive meta-improvement. 
Next we examine RSI across scenarios (e.g., scientific discovery, embodied intelligence, software engineering), highlighting their distinct requirements and development speeds. Drawing on diverse industry practices and preliminary empirical evidence, we connect RSI research with practical systems and identify key challenges to achieving genuine RSI.}

\titleformat*{\paragraph}{\bfseries}
\checkdata[Project Page]{\url{https://theseus-labs-rsi.github.io/}}
\checkdata[GitHub Repository]{\url{https://github.com/theseus-labs-rsi/awesome-rsi}}

\begin{document}







\setlist[itemize]{leftmargin=0.6cm, topsep=1pt, itemsep=0pt}
\setlist[enumerate]{leftmargin=0.6cm, topsep=1pt, itemsep=0pt}

\maketitle


{
\renewcommand{\thefootnote}{\fnsymbol{footnote}}
  \footnotetext[1]{Equal Contribution} 
  \footnotetext[2]{Corresponding author: Xuanhe Zhou}%
}

\begin{center}
\begin{tcolorbox}[
    width=0.95\linewidth,
    colback=black!3,
    colframe=black!40,
    boxrule=0.4pt,
    arc=2pt,
    left=2pt, right=2pt,
    top=1pt, bottom=1pt,
    before skip=2pt,
    after skip=2pt
]
\bfit{TL;DR} \textit{Like human evolution, AI evolution will unfold through a vast and extraordinary history. Everything AI has achieved so far is but a drop in the ocean.} 

\end{tcolorbox}
\end{center}

\begin{figure}[h]
    \vspace{-1em}
    \centering
    \includegraphics[width=0.9\textwidth]{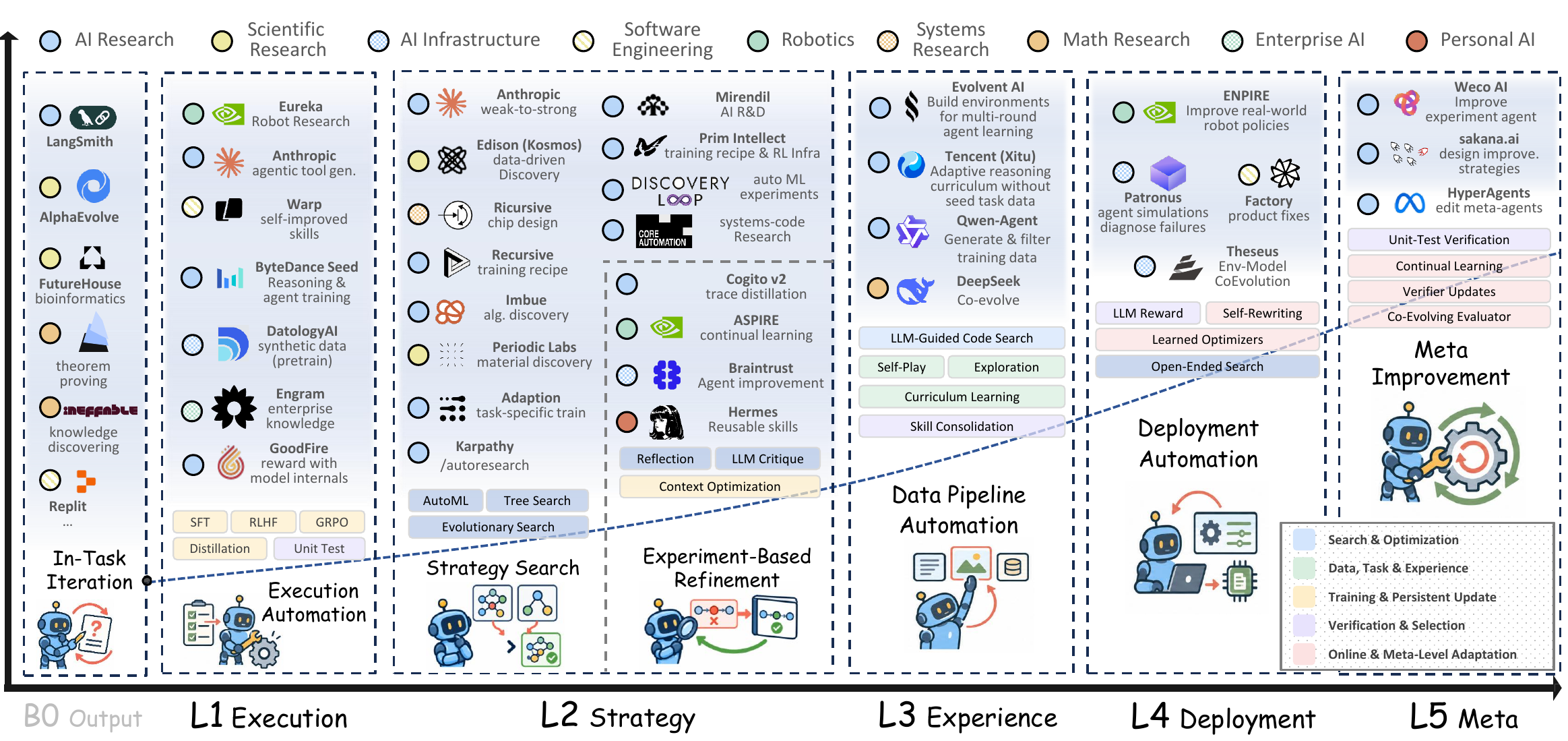}
    \vspace{-.4em}
    \captionsetup{justification=centering}
    \caption{\textbf{Overview of the five RSI autonomy levels and representative systems.} Autonomy progressively expands from executing prescribed improvements (L1), to selecting improvement strategies (L2), acquiring future learning experience (L3), adapting through deployment and environmental feedback (L4), and ultimately improving mechanisms that govern subsequent improvement (L5). See system details in Appendix~\ref{app:industry-landscape}.}
    \label{fig:rsi-stages}
\end{figure}

\clearpage
\setcounter{tocdepth}{2}

\tableofcontents
\clearpage






\section{Introduction}


Recent frontier-model development illustrates several forms of scaling in the improvement pipeline. Kimi K3 and Qwen3.8-Max contain 2.8 trillion and 2.4 trillion parameters, respectively, and each supports a context window of approximately one million tokens~\cite{moonshot2026kimik3,alibaba2026qwen38max}. The development process is also expanding. During the six months preceding GPT-5.6, OpenAI reports that the share of research compute devoted to internal coding inference grew 100-fold and internal agentic token use grew 22-fold, and average daily output tokens per active researcher exceeded twice the previous peak observed with GPT-5.5~\cite{openai2026gpt56}. Scale accumulates across training runs, model-assisted experiments, inference, evaluation, and human validation.

\subsection{Scaling Burdens in Model Development}

Despite the growing use of agentic tools and API-based automation, developers must still determine what to improve, construct the required resources, and establish whether each change works~\cite{nvidia2025aimo,meta_capacity_efficiency_2026}. As AI systems take on more demanding tasks, the cost of scaling this end-to-end development process becomes a bottleneck~\cite{openai2025gdpval,phan2025hle,swebench,yao2024tau}. The following three challenges occur at different stages of the model lifecycle and motivate RSI:




\noindent$\bullet$ \bfit{Development challenge 1: Resource-intensive foundation-model training.} Foundation-model development remains resource-intensive across data preparation, architecture design, distributed optimization, and evaluation. Kimi K3 activates 16 of 896 experts and reports an approximate 2.5-fold improvement in scaling efficiency over Kimi K2, while Qwen3.8-Max activates 95 billion of its 2.4 trillion parameters~\cite{moonshot2026kimik3,alibaba2026qwen38max}. Sparse activation reduces per-token computation, but training at this scale still couples expert routing, parallelism, multimodal integration, long-context optimization, and systems design. OpenAI further reports that GPT-5.6 Sol designed and ran hundreds of experiments on its speculative-decoding draft model and monitored training through hardware failures and instability. The resulting changes improved token-generation efficiency by more than 15\%~\cite{openai2026gpt56efficiency}. Data quality also matters in addition to quantity. OpenAI's GDPval illustrates that its 1,320 professional tasks required roughly 9,240 expert-hours in total, with contributors averaging more than 14 years of experience~\cite{openai2025gdpval}. The Humanity's Last Exam pipeline logged more than 70,000 submission attempts and sent approximately 13,000 model-stumping questions to expert review before producing a 3,000-question benchmark~\cite{phan2025hle}. Architecture search remains expensive because candidate structures interact with their data, optimization, and hardware regimes. AgentNAS addresses part of this bottleneck by using an LLM to propose a task-specific seed architecture and construct its search space, but candidate selection still depends on combinatorial search under an externally specified objective~\cite{jeong2026agentnas}.



\noindent$\bullet$ \bfit{Development challenge 2: Scaling feedback and learning environments.} Synthetic data and reinforcement learning automate parts of capability development, but introduce substantial requirements for generating and evaluating experience, requiring both experience-generation infrastructure and reliable mechanisms for evaluating and retaining updates. For instance, DeepSeek-V3.2 reports a post-training computational budget exceeding 10\% of its pretraining cost~\cite{deepseekai2025deepseekv32pushingfrontieropen}, while NVIDIA's AIMO-2 pipeline generated 3.2 million long-reasoning solutions and 1.7 million tool-integrated solutions, in addition to curating 540,000 problems~\cite{nvidia2025aimo}.


\noindent$\bullet$ \bfit{Development challenge 3: Recurring adaptation after deployment.} Deployed systems consistently introduce changing documents, unfamiliar tools, incomplete context, and workflows involving interdependent actions. Improving such systems requires engineers to manually diagnose failures, revise retrieval and tool interfaces, manage persistent state, and repeat regression testing through discrete, human-led releases. Anthropic reports that agentic workloads use approximately four times as many tokens as ordinary chat, rising to about fifteen times for multi-agent systems because of longer contexts, coordination, environment setup, and end-to-end verification~\cite{anthropic2025research}, while Meta reports that FBDetect identifies thousands of infrastructure regressions each week, and diagnosing one such regression required roughly ten engineer-hours~\cite{meta_capacity_efficiency_2026}.





\subsection{From Development Burden to RSI}

The burdens arise because model improvement remains a sequence of costly, externally coordinated interventions. To address these barriers, RSI inspects whether part of that coordination can become a persistent capability of the system being improved. We define recursive self-improvement (RSI) as an autonomous, closed-loop process in which an AI system identifies its own limitations, develops and validates improvements, and uses the resulting capabilities to improve the improvement process itself. The model evolution paradigm of RSI spans three dimensions: autonomy, efficiency, and innovation. Autonomy expands the system's responsibility from executing a prescribed update to identifying limitations, extracting experience, proposing changes, and validating and retaining successors. Efficiency seeks more validated improvement from data, compute, inference, human review, and rework. Innovation allows the system to search beyond human-prescribed update strategies and feed useful discoveries back into later improvement rounds. These dimensions describe how the full improvement loop is organized and what it can inherit.


Rather than a particular learning algorithm or a one-off optimization result~\cite{schmidhuber2003godel,zelikman2024stop,zhang2026dgm}, RSI aims to improve both task performance and the mechanisms through which later improvements are discovered and implemented. The following cases illustrate this distinction at two points in the model lifecycle, including foundation-model training and persistent adaptation in software engineering, covered and studied in later sections.





\noindent$\bullet$ \textbf{Case 1: Foundation-model training.} A conventional experiment loop selects a better checkpoint while leaving the procedure for choosing later experiments unchanged. A-Evolve-Training instead consolidates post-training outcomes into a persistent research policy and discovery log, which a meta-agent revises to guide later workers' recipe choices~\cite{shi2026aevolvetraining}. When development scores improved without corresponding external gains, the system redirected experiments toward data rebalancing and checkpoint selection. The retained policy changes both how successor models are trained and how later improvements are sought. Across four autonomous rounds on a 30B Nemotron model, the external score rose from 0.80 to 0.86, compared with 0.87 for the top human submission~\cite{shi2026aevolvetraining}.

\noindent$\bullet$ \textbf{Case 2: Software-engineering adaptation after deployment.} Repairing a repository changes the software product but may leave the coding agent's recurring failures untouched. Ouroboros instead uses reviewed deployment evidence to propose versioned changes to the agent's tools, context assembly, prompts, and core implementation~\cite{razzhigaev2026ouroboros}. Candidate revisions undergo tests and human review before an accepted version replaces the runtime used for later work. The persistent update improves subsequent coding behavior and changes the mechanism through which later failures are diagnosed and repaired, while experts retain control over consequential corrections and deployment.

\subsection{Challenges for RSI}

While the two cases show how persistent changes can shape later improvement, the same persistence can carry errors or obscure where control resides. We examine three recurring problems that determine whether a self-updating system provides credible evidence of RSI. 

\noindent$\bullet$ \textbf{Safe inheritance.} RSI requires changes to persist across tasks or improvement rounds, but persistence alone does not guarantee sustained gains. G\"odel Agent~\cite{yin2025godelagent}, for example, rewrites both its task policy and improvement logic, yet 14\% of its 100 MGSM optimization trials ended below the initial policy's performance. Transfer tests, version histories, and rollback mechanisms are needed to retain useful updates without degrading earlier capabilities. 

\noindent$\bullet$ \textbf{Autonomy attribution.} Generating better candidates does not necessarily mean the system has improved how candidates are discovered or selected. The Darwin G\"odel Machine~\cite{zhang2026dgm} evolves coding agents, raising performance on its SWE-bench subset from 20\% to 50\%, but its archive maintenance and parent-selection rules remain outside self-modification. RSI analysis must distinguish AI-controlled decisions from fixed search procedures and human acceptance criteria.

\noindent$\bullet$ \textbf{Reliable verification.} Repeated evaluator access can reward exploitation rather than capability gains. Anthropic's automated research experiments~\cite{anthropic2026weakstrong} report random-seed cherry-picking and attempted test-label extraction through evaluator queries. Evolving evaluators further complicate comparisons across rounds. The Red Queen G\"odel Machine~\cite{iacob2026redqueen} addresses this by freezing evaluators within each epoch and validating replacements against an independent ground-truth anchor. Protected evaluation and matched computational budgets are needed to separate genuine improvement from evaluator exploitation or increased search effort.

\subsection{An Autonomy-Centered Framework}

\begin{figure*}[!t]
    \centering
    \includegraphics[width=1\linewidth]{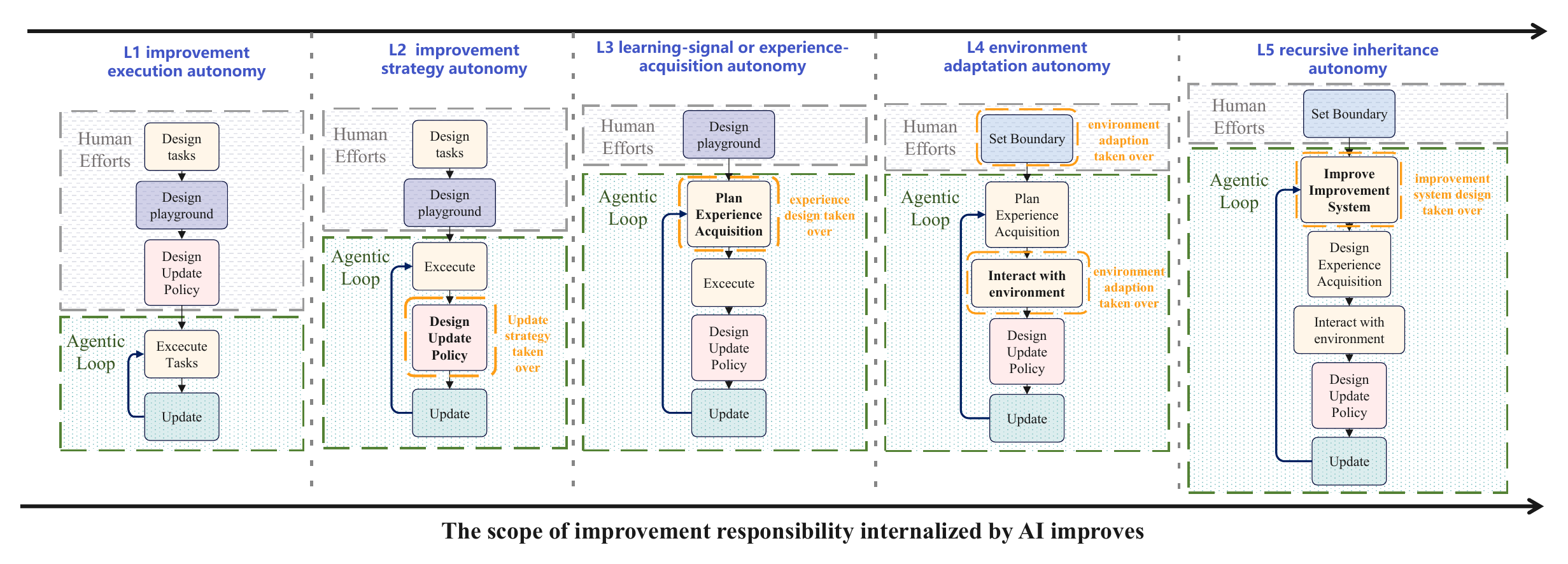}
    \caption{Loop Patterns of Five Levels. Gray dashed frames indicate human-controlled components. Green dashed frames indicate components within the RSI loop. Orange outlines highlight the newly internalized component at each level. The expanding green frames show that AI progressively automates a larger share of the improvement process.} 
    \label{fig:five-loops}
\end{figure*}

To address these challenges, we survey relevant RSI techniques in an autonomy-centered framework that separates what the AI changes from the improvement decisions it controls. Based on the scope of improvement responsibility internalized by AI, we review RSI techniques across five levels. Figure~\ref{fig:rsi-stages} maps representative systems across this progression, while Figure~\ref{fig:five-loops} isolates the corresponding loop structures. At each level, we identify where the improvement loop closes, what is retained for later rounds, and which critical decisions remain under human control, then introduce the techniques that implement this division of responsibility. 



\noindent$\bullet$ \textbf{(L1) Improvement Execution Autonomy.} Humans specify what should be improved, how it should be improved, and what constitutes success, while AI executes candidate updates. For example, FineWeb-Edu uses a model to apply human-defined educational-quality labels across a web corpus without choosing the labeling criterion~\cite{penedo2024fineweb}.

\noindent$\bullet$ \textbf{(L2) Improvement Strategy Autonomy.} The objective, task boundary, and evaluation criteria remain externally fixed, but AI diagnoses weaknesses and decides how to improve the system. For example, Self-Harness uses execution traces to propose and test edits to its agent harness under a fixed benchmark and promotion rule~\cite{zhang2026selfharness}.

\noindent$\bullet$ \textbf{(L3) Learning-Signal or Experience-Acquisition Autonomy.} The system also determines the experience needed for its next improvement round. For example, SIMA 2 uses assessments of current behavior to generate later practice tasks that target observed skill weaknesses~\cite{deepmind2025sima2}.

\noindent$\bullet$ \textbf{(L4) Environment Adaptation Autonomy.} The improvement loop uses deployment interaction to revise persistent system state under external acceptance and governance rules. For example, PANDO admits or demotes reusable rules during a long-running interaction according to observed outcomes, so later actions inherit earlier experience~\cite{li2026pando}.

\noindent$\bullet$ \textbf{(L5) Recursive Inheritance Autonomy.} The system persistently revises a mechanism that governs subsequent improvement, such as an \Improver, \Verifier, or successor-generation procedure. For example, A-Evolve-Training revises its research policy after development scores fail to predict external gains and uses the revised policy to direct the next training round~\cite{shi2026aevolvetraining}.


\subsection{Application Domains}

While the autonomy levels describe the structure of an improvement loop, their practical meaning depends on the feedback available in a domain, as the same retained update may be straightforward to test in software engineering and difficult to validate in a physical or clinical setting. We consider science, embodied intelligence, software engineering, and healthcare because they expose four distinct feedback regimes, including experimental evidence with uncertain attribution, physical interaction with costly trials, executable tests with incomplete specifications, and high-stakes outcomes under expert oversight. These regimes allow us to compare how feedback cost and reliability affect the retention and reuse of improvements.

\noindent$\bullet$ \textbf{(S1) RSI for Science.}
Scientific discovery involves open-ended exploration, costly experiments, and feedback that may not clearly identify the source of failure. We examine how accumulated evidence can improve scientific hypothesis modules, experimental agents, and reflection or improvement mechanisms, with attention to whether these changes support subsequent research beyond the current scientific result.

\noindent$\bullet$ \textbf{(S2) RSI for Embodied Intelligence.}
Embodied agents generate experience through their own actions, while failures may arise from interacting perception, planning, and control components. Physical trials also impose limits on exploration and repeatability. We examine the evolution of environments and curricula, skills and agent harnesses, policies and action models, and world models and evaluators, focusing on how interaction feedback supports validated improvements that can be reused in later tasks.

\noindent$\bullet$ \textbf{(S3) RSI for Software Engineering.}
Software engineering makes both the developed artifact and the developing agent accessible to executable modification and testing. We examine how repository feedback supports persistent changes to coding-agent implementations and harnesses, development experience and collaboration, and the improvement process itself. A central distinction is whether an update improves current task performance, the ability to produce stronger successors, or both.

\noindent$\bullet$ \textbf{(S4) RSI for Healthcare.}
Healthcare combines restricted opportunities for trial and error with delayed, heterogeneous feedback and improvements whose validity may depend on the patient population or institution. We examine the evolution of clinical memory and knowledge, reasoning strategies, and tools and workflows, emphasizing how reviewed experience can inform subsequent cases under explicit validation and oversight.

Across these domains, we compare what is updated, how feedback supports its retention, and which decisions remain externally controlled. This analysis connects the autonomy framework to application-specific evidence and identifies the gaps between demonstrated improvement mechanisms and fuller recursive improvement.







\subsection{Industrial Evidence}

The domain analysis identifies the feedback and validation conditions that shape an improvement loop, while industrial systems show how these conditions are handled within operating pipelines, where integration and deployment constraints are immediate. We examine industrial practice because frontier improvement loops are not always first documented through conventional academic publications. Industrial materials, including technical reports, open-source systems, engineering blogs, model documentation, and deployed product infrastructures, often reveal system-level practices, such as evaluation pipelines, data flywheels, agent harnesses, automated experimentation, and deployment feedback loops, which are only partially represented in the academic literature. We use these materials to complement the research literature and to understand how self-improvement is implemented under real engineering constraints.

Building on the autonomy-centered framework and application analysis, we examine what responsibilities industrial systems assume for their own improvement and how these responsibilities vary across applications and engineering settings. We further analyze how constraints such as computational cost, feedback quality, and human involvement shape the organization of improvement loops and limit the attainable scope of autonomy. This perspective allows us to characterize both the mechanisms that have been demonstrated in deployed or production-oriented systems and the more ambitious visions of recursive improvement that remain to be validated, thereby clarifying the current progress and limitations of industrial \RSI{} practice. Figure~\ref{fig:rsi-landscape} reports the surveyed literature by autonomy level and improvement target, while Table~\ref{tab:industry-landscape} provides the corresponding landscape of industrial systems.

\subsection{Differences from Existing Surveys}

Existing surveys provide complementary taxonomies of self-evolving systems and the mechanisms from which improvement loops are built. These works establish much of the technical vocabulary on which our analysis relies. Our survey differs in four respects.

\noindent$\bullet$ \emph{Improvement loop as the unit of analysis.} Prior surveys organize work by stages of self-evolution, update objects, timing, or technical mechanisms~\cite{tao2024selfevolution,gao2026selfevolving,fang2025selfevolving,ren2026selfimprovements}. These views explain what changes and how the change is produced, but systems that update the same component may assign very different decisions to AI. We trace a complete loop: what triggers improvement, who proposes and validates a change, what persists, and which later decisions use the retained change.

\noindent$\bullet$ \emph{Responsibility as the autonomy criterion.} Related frameworks examine capability levels, co-evolution, dynamic agent state, and AI-for-AI systems~\cite{liu2026path,zong2026coevolution,xu2026dynamicgraph,ye2026ai4ai}. We operationalize autonomy through the improvement decisions transferred from external designers to the AI rather than through model capability or the number of automated components. Our five levels distinguish responsibility for execution, strategy selection, experience acquisition, environmental adaptation, and recursive inheritance.

\noindent$\bullet$ \emph{Separate evidence for recursion and performance.} Evaluation surveys study model-based judgment, agent assessment, rubric-guided learning, and oversight failures~\cite{li2025judge,yehudai2026evaluation,shan2026rubric,kim2024superalignment,slattery2024risk}. Higher task performance alone does not show that an improvement mechanism was revised, retained, and reused. We distinguish structural recursion, in which a revised improvement mechanism governs a later round, from effective recursion, in which that mechanism produces stronger successors under comparable budgets and independent evaluation.

\noindent$\bullet$ \emph{Mechanisms compared across operating conditions.} Work on correction, synthetic data, lifelong learning, memory, prompt optimization, and workflow design explains how individual components improve~\cite{pan2024correction,venktesh2025verification,niu2026testtime,long2024synthetic,zheng2026lifelong,huang2026memory,zhou2026externalization,ramnath2025prompt,lee2025compound,yue2026workflow,li2026environment}. We examine how these mechanisms function within complete loops across science, embodied intelligence, software engineering, healthcare, and industrial practice, where feedback cost, validation, and external control differ~\cite{ding2026verificationgap,zhou2026coding,zhu2026clinical}. Comparing the same loop questions across these settings reveals when a method depends on cheap executable feedback, repeated interaction, expert review, or production infrastructure.

These choices together position the survey between a catalog of improvement mechanisms and a general hierarchy of AI capabilities. Our aim is to determine which parts of an improvement loop current systems can assume, how retained changes affect later rounds, and what evidence supports claims of recursive progress.

\section{Background and Preliminaries}

This section establishes the empirical and conceptual basis for our
autonomy-centered view of recursive self-improvement.
We first examine the uneven progress of modern foundation models across
different capability domains, highlighting why persistent improvement is
particularly relevant to interactive, stateful, and tool-using systems.
We then characterize RSI through the structure of an improvement loop,
introduce the key components needed to analyze how improvements are generated,
validated, retained, and inherited, and distinguish RSI from neighboring
paradigms such as continual learning, AutoML, and agentic AI.
Finally, because contemporary RSI spans both academic research and industrial
engineering practice, we clarify the scope and strength of the evidence used
throughout this survey.
Together, these preliminaries provide the empirical motivation, operational
vocabulary, and evidentiary basis for the autonomy hierarchy developed in the
following section.





\subsection{Uneven Capability Progress in Modern Foundation Models}

\begin{figure*}[!t]
    \centering
    \includegraphics[width=\textwidth]{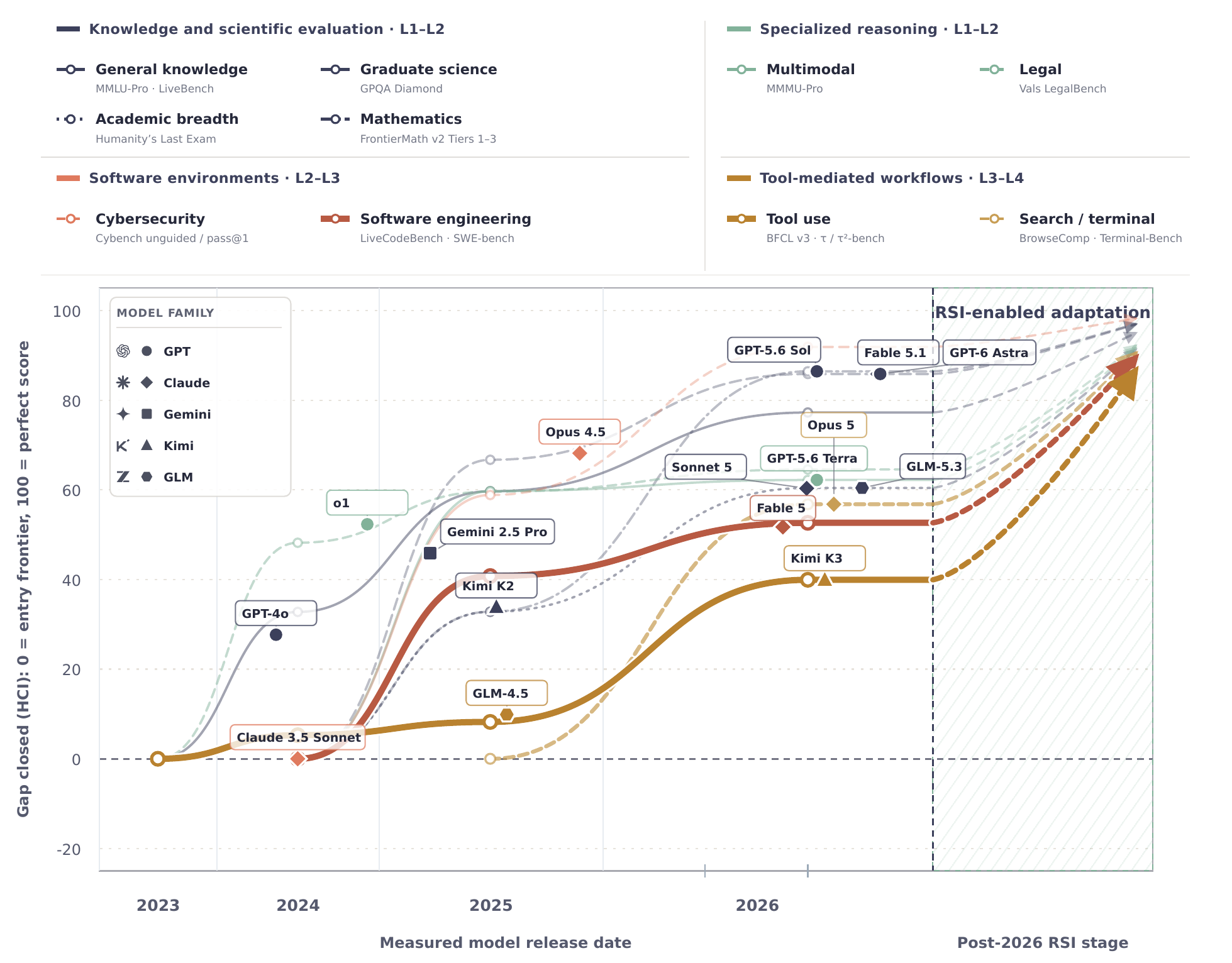}
    \caption{Cross-domain capability trajectories and an illustrative RSI extension. Each line reports an annual domain frontier in HCI, where 0 is the benchmark's entry-year frontier and 100 is a perfect score. Dashed cybersecurity segments denote changes in Cybench subsets or \texttt{pass@1} aggregation. The post-2026 region illustrates the extension of domains with the assistance of RSI.}
    \label{fig:rsi-longitudinal-capability}
\end{figure*}

Figure~\ref{fig:rsi-longitudinal-capability} summarizes 393 eligible model--benchmark observations across ten capability domains for models released between 2023 and September 2026.

We first group reported scores into protocol-link families. Two results belong to the same family only when the benchmark version and evaluation harness remain stable, or when evaluations of overlapping models provide a defensible bridge between protocols. Results obtained with incompatible versions or harnesses remain in the audit records but are excluded from the trajectories. This rule admitted 17 of the 33 results considered in the latest-model audit. The other 16, drawn from Terminal-Bench~\cite{merrill2026terminalbench}, DeepSWE~\cite{huang2026deepswe}, CyberGym~\cite{wang2025cybergym}, ExploitBench~\cite{lee2026exploitbench}, AutomationBench~\cite{shepard2026automationbench}, and BrowseComp~\cite{wei2025browsecomp}, were retained for reference because their benchmark versions or evaluation settings could not be linked to the plotted families. If several sources report the same model under the same benchmark family, variant, and evaluation mode, we calculate the following weighted consensus:
\begin{equation}
\bar{s}_{mbh}=\frac{\sum_i \widetilde{w}_i s_{imbh}}{\sum_i \widetilde{w}_i},
\label{eq:source-weighted-consensus}
\end{equation}
where $h$ denotes the evaluation protocol and $i$ indexes sources. Benchmark-owner tables, independent common-harness evaluations, combined benchmark or model reports, and model-author tables receive base weights of 3, 2.5, 2, and 1, respectively. We multiply first-party values by an additional factor of 0.75. These weights provide a practical sensitivity adjustment, limiting the influence of self-reported results.

The underlying benchmarks report different quantities, so their raw values cannot be pooled directly. We normalize the consensus score with the Headroom-Closed Index (HCI). For model $m$ on benchmark family $b$ under protocol $h$,
\begin{equation}
H_{mbh}=100\times\frac{\bar{s}_{mbh}-F_{b,0}}{100-F_{b,0}},
\label{eq:headroom-closed-index}
\end{equation}
where $F_{b,0}$ is the 90th-percentile model score in the first year that benchmark enters the dataset. Thus, $H_{mbh}=0$ denotes the entry-year frontier and $H_{mbh}=100$ denotes a perfect score. For each benchmark family and release year, we first compute the 90th-percentile HCI frontier $Q_{b,y}$. The domain trajectory is then
\begin{equation}
T_{d,y}=\frac{\sum_{b\in\mathcal{B}_{d,y}}\sqrt{n_{b,y}}\,Q_{b,y}}
{\sum_{b\in\mathcal{B}_{d,y}}\sqrt{n_{b,y}}},
\label{eq:domain-hci-trajectory}
\end{equation}
where $\mathcal{B}_{d,y}$ is the set of eligible benchmark families for domain $d$ in year $y$, and $n_{b,y}$ is the number of distinct models contributing to that benchmark-year frontier. The square-root weight allows better-covered families to contribute more without letting the largest table dominate the domain value. Three observations follow.

\noindent$\bullet$ \bfit{Observation 1: Frontier gains differ in magnitude and timing.}
By 2026, advanced mathematics and graduate-level science reach HCI values of 86.4 and 85.8, while broad knowledge reaches 77.2. The corresponding values for legal reasoning, multimodal reasoning, and frontier academic breadth are 64.5, 62.2, and 60.4. The annual increment $\Delta T_{d,y}=T_{d,y}-T_{d,y-1}$ exposes different temporal patterns. Broad knowledge rises by 32.8, 26.9, and 17.6 points across the three annual transitions, indicating steady but slowing headroom closure. Legal reasoning similarly slows from a 48.2-point gain in 2024 to 11.4 in 2025 and 4.9 in 2026. Advanced mathematics follows a different path, whose increment grows from 32.8 points in 2025 to 53.6 in 2026. Multimodal reasoning gains 59.7 points in 2025 but only 2.5 in 2026. A single aggregate benchmark would conceal these differences in both level and trajectory shape.

\noindent$\bullet$ \bfit{Observation 2: Interactive capabilities retain larger gaps.}
Software engineering reaches an HCI of 52.6 in 2026, search and terminal agents reach 56.8, and tool agents reach 39.9. Relative to graduate-level science at 85.8, their normalized headroom closure is lower by 33.2, 29.1, and 45.9 points, respectively. Tool agents improve sharply in 2026, from 8.2 to 39.9, yet remain the lowest trajectory. Software engineering gains 40.8 points in 2025 and 11.9 in 2026. The leading cybersecurity-agent trajectory reaches 91.9, producing a 52.0-point difference from tool agents. This comparison requires caution because the later Cybench observations use changed task subsets or \texttt{pass@1} aggregation, as indicated by the dashed line. Even with that qualification, the figure shows that gains in bounded or readily verified environments have not transferred uniformly to long, stateful workflows. Model developers increasingly turned their attention to agentic coding and tool use after 2024, and these capabilities became prominent research and engineering targets during 2025 and 2026~\cite{wang2026hgm,chen2026sweexp,sun2026seagent,li2026pando}. Such tasks require planning, environment-state tracking, tool selection, result interpretation, and revision of subsequent actions. Errors propagate across the trajectory, so data collection and evaluation must cover complete interactions. In the paper's autonomy taxonomy, bounded evaluations mainly exercise L1--L2 capabilities, environment tasks increasingly require L2--L3 capabilities, and interactive workflows expose the verification, memory, and adaptation requirements associated with L3--L4 operation.

\noindent$\bullet$ \bfit{Observation 3: Remaining headroom concentrates the potential value of RSI.}
The hatched post-2026 region illustrates how persistent and verified recursive self-improvement could preferentially affect domains with larger remaining gaps. To express this relationship consistently, the illustrative endpoint for domain $d$ is
\begin{equation}
R_d=100-0.22\left(100-T_{d,2026}\right).
\end{equation}
Equivalently, the illustrative gain is $R_d-T_{d,2026}=0.78(100-T_{d,2026})$, so domains with more unclosed headroom receive a larger extension. Cybersecurity therefore moves from 91.9 to 98.2, while software engineering, search and terminal agents, and tool agents move from 52.6 to 89.6, from 56.8 to 90.5, and from 39.9 to 86.8. The endpoints illustrate the hypothesis that repeated experience generation, validation, update retention, and regression testing could direct more improvement toward weak deployment workflows.

Software engineering and tool use are therefore particularly relevant to RSI. Human-led updates require repeated environment construction, trajectory collection, failure diagnosis, training or harness revision, and regression testing as interfaces and repositories change. The methods reviewed later generate practice from observed weaknesses~\cite{sun2026seagent}, distill trajectories into reusable rules and tools~\cite{li2026pando,dai2026metis}, and retain tested harness or code changes for future tasks~\cite{zhang2026selfharness,lin2026ahe}. These mechanisms can direct successive updates toward failures observed during deployment and reduce the capability gaps represented by the illustrative extensions.

\subsection{Conceptual Foundations of Recursive Self-Improvement}

The idea that an intelligent system may participate in improving its own future behavior has several intellectual predecessors, including self-modifying programs, the G\"odel Machine, automated machine learning, meta-learning, continual learning, open-ended learning, and more recent agentic systems capable of modifying code, prompts, tools, or training procedures~\cite{schmidhuber2003godel,ye2026ai4ai,shi2025continual,wang2019poet,hu2025adas,zhang2025aflow}. These traditions address overlapping parts of the problem, but they do not by themselves provide a sufficient criterion for \RSI. Automated optimization does not necessarily imply self-improvement, persistent learning does not necessarily imply autonomy over the learning process, and an AI-generated improvement to an external artifact does not necessarily modify the AI system that generated it.

For this reason, we take the \emph{improvement loop} rather than any particular algorithm as the basic unit of analysis. This perspective allows systems implemented through very different mechanisms to be compared according to the same questions: what generates the experience, what is changed, what persists into future interactions, who controls the update process, and whether the result of one improvement round participates in determining subsequent improvements.

\subsubsection{Anatomy of an Improvement Loop}

\noindent\bfit{Improvement loop.}
We define an improvement loop as a recurring process in which an AI system uses experience to propose a modification to a target, evaluates the candidate under an acceptance rule, retains an accepted change in its state, and begins the next round from that updated state. The following components identify where each operation occurs and what information passes between rounds.

\noindent$\bullet$ \bfit{AI system:} the complete computational entity whose capability state is tracked across improvement cycles. It includes the mechanisms and persistent state that determine how the system acts, generates modifications from feedback, and carries accepted results into the next round. In automated code improvement, for example, the AI system includes the components that propose a patch, execute the modified program, evaluate the result, and continue from a retained version.

\noindent$\bullet$ \bfit{System state:} the part of the AI system retained at the end of one round and inherited by the next. It is what the next round receives from the previous one. If a system accepts a faster program and continues modifying that version, the retained program belongs to its system state.

\noindent$\bullet$ \bfit{Experience:} information obtained from earlier interactions and used to guide a later modification. A test failure, environment outcome, or reviewer correction becomes experience when it informs a subsequent update.

\noindent$\bullet$ \bfit{Target:} the object directly modified in the current round. In code optimization, the target may be a sorting function. If the system later changes its method for generating code candidates, that search method becomes the target.

\noindent$\bullet$ \bfit{Improver:} 
the mechanism that transforms the current state and available experience into candidate modifications. A model that reads the current code and a recent test failure before proposing the next patch acts as the improver in that round.

\noindent$\bullet$ \bfit{Strategy:} the method used by the improver to decide where to search and how to generate candidates. A strategy may prioritize code locations implicated in failed tests. Because the strategy can be retained, it may also become a target of later improvement.

\noindent$\bullet$ \bfit{Verifier:} the mechanism that evaluates candidates and applies the acceptance rule. It may use benchmark scores, unit tests, reward models, environment outcomes, human feedback, formal constraints, safety checks, or combinations of these signals.

\noindent$\bullet$ \bfit{Improvement:} a candidate modification that passes the current acceptance rule, is retained, and enters the next system state. An improvement is therefore an accepted and inherited state change.



\noindent$\bullet$ \bfit{Successor:} the AI system that inherits one or more accepted improvements and enters the next round with an updated state. For example, a system that retains a revised code-search strategy and uses it to generate later modifications is the successor of the version that produced the strategy.

This decomposition yields three cross-cutting questions that will be used throughout the survey. \emph{Where does the loop close?} determines whether an apparent improvement actually returns to the system. \emph{What is updated and inherited?} determines the persistent carrier of improvement. \emph{Which decisions remain external?} determines how much authority over the improvement process has been transferred from humans or fixed infrastructure to the AI system itself.

This anatomy applies to persistent improvement processes with different degrees of automation. The next subsection specifies the additional conditions for \RSI{}: experience must produce persistent self-change with sufficient autonomy, and an accepted change must be able to affect how later improvements are generated, evaluated, selected, or consolidated.

\subsubsection{Definition of Recursive Self-Improvement}
Using the improvement-loop anatomy above, we define \emph{recursive self-improvement} (\RSI) as the capability of an intelligent system, through continued interaction with tasks, environments, or other intelligent agents, to autonomously transform acquired experience and feedback into persistent changes to itself across interaction rounds (e.g., model parameters, agent harnesses, or improvement policies), such that these changes can further affect the mechanisms used to generate, evaluate, select, and consolidate subsequent self-improvements. The improved system is consequently reintroduced into the next round of interaction and improvement with an already changed capability state, allowing the system's capacity for improvement itself to become part of an ongoing recursive process.

\RSI{} contains an autonomous, closed-loop process in which AI identifies its own limitations, develops and validates improvements, and uses the resulting capabilities to improve the improvement process itself, with the aim of (1) expanding its capability frontier, (2) increasing resource efficiency, or (3) discovering novel solutions beyond human-prescribed strategies.

Recent industry perspectives increasingly operationalize \RSI\ through autonomous improvement loops. OpenAI emphasizes the automation of AI research workflows and feedback loops~\cite{openai2026rsi}, while Tencent reports an early-stage RSI loop in which experimental results are fed into subsequent rounds of model development~\cite{tencent2026hy4}. More strictly, Alibaba researchers define RSI by making the improvement mechanism itself subject to modification~\cite{liu2026path}, whereas Anthropic describes its strongest form as an AI system autonomously designing and developing its own successor~\cite{anthropic2026rsi}.


\subsubsection{Relation to Neighboring Paradigms}
\RSI \ intersects with continual learning, automated machine learning (AutoML), and agentic AI because all three automate parts of an improvement loop. They differ in the usual target of improvement, the state retained across rounds, and the decisions left to a fixed procedure or an external actor. Table~\ref{tab:rsi-neighboring-paradigms} summarizes these distinctions.

\emph{Continual learning}: Continual learning enables a model or agent to acquire knowledge from a sequence of tasks or data while limiting the loss of earlier capabilities~\cite{shi2025continual,wu2024continual,zheng2024lifelongllms}. Recent work covers continual pre-training, instruction tuning, alignment, replay, parameter-efficient updates, and memory-based adaptation~\cite{shi2025continual,wu2024continual}. Lifelong and self-evolving agents further extend persistent state beyond model weights to memories, skills, and behavioral policies~\cite{zheng2026lifelong,fang2025selfevolving}. The learning objective, update rule, experience schedule, and acceptance test nevertheless usually remain fixed by the system designer~\cite{shi2025continual,zheng2024lifelongllms}. In our framework, continual learning approaches \RSI{} when experience can also revise how later adaptations are proposed, evaluated, or consolidated, and the revised process is reused in subsequent rounds~\cite{zheng2026lifelong,fang2025selfevolving}.

\emph{AutoML}: AutoML automates model-development decisions such as data processing, model and architecture selection, hyperparameter optimization, and pipeline construction~\cite{trirat2025automlagent,ye2026ai4ai}. Recent systems broaden this scope through language-model agents: AutoML-Agent constructs and verifies complete machine-learning pipelines, while ADAS, AFlow, and AgentSquare search over executable agent designs, workflow graphs, or reusable modules~\cite{trirat2025automlagent,hu2025adas,zhang2025aflow,shang2025agentsquare}. MLE-bench and the AI Scientist-v2 extend evaluation to sustained machine-learning engineering and iterative research workflows~\cite{chan2025mlebench,yamada2025aiscientistv2}. Their search spaces, objectives, budgets, and evaluators are generally supplied in advance, even when much of the development work is automated~\cite{trirat2025automlagent,hu2025adas,zhang2025aflow}. AutoML reaches the \RSI{} boundary when the procedure used to search for or evaluate improvements becomes persistent state that later rounds can improve~\cite{hu2025adas,zhang2025aflow,fang2025selfevolving}.

\emph{Agentic AI and agentic ML}: Agentic systems plan multistep work, invoke tools, run experiments, coordinate specialized components, and revise intermediate artifacts~\cite{chan2025mlebench,starace2025paperbench,yamada2025aiscientistv2}. MLE-bench, PaperBench, and the AI Scientist-v2 study these capabilities in machine-learning engineering and research settings~\cite{chan2025mlebench,starace2025paperbench,yamada2025aiscientistv2}, while ADAS and AFlow automate the construction of the agent or workflow that performs the task~\cite{hu2025adas,zhang2025aflow}. These systems may operate within an episode whose harness, tools, stopping rule, and acceptance criteria remain unchanged, a boundary identified in work on self-evolving agents~\cite{fang2025selfevolving,zheng2026lifelong}. Agentic ML approaches \RSI{} when validated changes to the agent system persist across tasks and affect how later improvements are generated or selected~\cite{fang2025selfevolving,hu2025adas,zhang2025aflow}.

\definecolor{NeighborPrimary}{RGB}{223,94,94}
\definecolor{NeighborSecondary}{RGB}{130,178,154}
\definecolor{NeighborOutside}{RGB}{60,64,91}
\newcommand{\PrimaryValue}{\textcolor{NeighborPrimary!85!black}{\ensuremath{\bm{\checkmark}}}}
\newcommand{\SecondaryValue}{\textcolor{NeighborSecondary!80!black}{\ensuremath{\bm{\circ}}}}
\newcommand{\OutsideValue}{\textcolor{NeighborOutside}{\ensuremath{\bm{\times}}}}
\begin{table}[t]
\centering
\begin{minipage}{0.68\textwidth}
\caption{Comparison of \RSI{} with neighboring paradigms.}
\label{tab:rsi-neighboring-paradigms}
\scriptsize
\setlength{\tabcolsep}{2.5pt}
\renewcommand{\arraystretch}{1.2}
\renewcommand{\tabularxcolumn}[1]{m{#1}}
\begin{tabularx}{\linewidth}{@{}
>{\raggedright\arraybackslash}X
>{\centering\arraybackslash}p{0.19\linewidth}
>{\centering\arraybackslash}p{0.15\linewidth}
>{\centering\arraybackslash}p{0.19\linewidth}
>{\centering\arraybackslash}p{0.12\linewidth}@{}}
\toprule
\textbf{Characteristic} & \textbf{Continual learning} & \textbf{AutoML} & \textbf{Agentic AI/ML} & \textbf{\RSI} \\
\midrule
Cross-round learning
& \PrimaryValue & \SecondaryValue & \SecondaryValue & \PrimaryValue \\
Persistent retention
& \PrimaryValue & \SecondaryValue & \SecondaryValue & \PrimaryValue \\
System self-modification
& \PrimaryValue & \SecondaryValue & \OutsideValue & \PrimaryValue \\
Candidate proposal
& \OutsideValue & \PrimaryValue & \PrimaryValue & \PrimaryValue \\
Update validation
& \SecondaryValue & \PrimaryValue & \SecondaryValue & \PrimaryValue \\
Successor re-entry
& \PrimaryValue & \OutsideValue & \OutsideValue & \PrimaryValue \\
Mechanism revision
& \OutsideValue & \SecondaryValue & \OutsideValue & \PrimaryValue \\
Mechanism reuse
& \OutsideValue & \OutsideValue & \SecondaryValue & \PrimaryValue \\
\bottomrule
\end{tabularx}
\par\vspace{3pt}
\raggedright
\PrimaryValue\ \textbf{Primary} core objective \quad
\SecondaryValue\ \textbf{Secondary} supporting role \quad
\OutsideValue\ \textbf{Outside scope} not addressed
\end{minipage}
\end{table}


The distinctive question posed by \RSI \ is therefore not simply whether AI contributes to AI development. It is whether a persistent improvement loop has formed around the system itself, and how much authority over that loop has become endogenous.

\subsection{Scope of Evidence}
The current development of \RSI \ spans academic research and industrial engineering practice. Restricting the evidence base to peer-reviewed publications would therefore omit systems whose most detailed descriptions appear in technical reports, official engineering blogs, open-source repositories, model documentation, or company research materials. We include these sources when they provide concrete evidence about the structure or operation of an improvement loop.

\noindent\textbf{Industrial practice is a primary source of evidence for contemporary \RSI.} Many of the most advanced self-improvement loops are developed in frontier AI systems before they are fully described in conventional academic papers. Their operational details often first appear in technical or research reports from leading AI laboratories, engineering blogs, open-source repositories, model releases on platforms such as GitHub and Hugging Face, and public benchmark or competition leaderboards. These sources expose aspects of \RSI\ that are particularly important to this survey: how improvement is organized in a working system, what artifacts are retained across iterations, how models and agents interact with evaluators and tools, and whether an improvement mechanism continues to operate beyond a single experimental result.

This broader evidence scope allows the survey to capture emerging \RSI\ practice while maintaining a clear distinction between demonstrated mechanisms and inferred ones.


\section{RSI Across  Autonomy Levels}

We organize existing work on RSI into a hierarchy of autonomy levels, according to how much responsibility the AI system assumes for its own improvement. The hierarchy progresses from \emph{in-session refinement}, where improvement is confined to the current task, through increasingly persistent and autonomous forms of system adaptation, toward \emph{recursive improvement}, where the mechanisms for producing future improvements themselves become subject to improvement.

At each level, we review representative approaches and systems while addressing three common questions: \emph{(1) where is the improvement loop closed, (2) what improvement is retained and carried into the next round, and (3) which critical decisions in the improvement process remain under human control?} 

Table~\ref{tab:rsi-technique-space} provides a cross-level map of representative implementation techniques.

\subsection{B0: In-Task AI Improvement}
\label{sec:b0-output-iteration}

At B0, improvement is confined to the current task: the AI system may iteratively revise, verify, or select among candidate outputs, but no resulting change is retained as persistent system state for future independent tasks. We therefore treat B0 as a non-RSI reference level that marks the boundary between \emph{improving an output} and \emph{improving the system that produces future outputs}. Put differently, the defining criterion of B0 is \emph{output change without persistent system change}.


\providecommand{\techworkcell}[1]{}
\providecommand{\techemptycell}{}
\providecommand{\techmechanism}[2]{}

\renewcommand{\techworkcell}[1]{%
  \begin{minipage}[t]{\linewidth}
    \vspace{0pt}
    \centering
    \fontsize{5.70pt}{6.35pt}\selectfont
    \setlength{\parskip}{0pt}%
    #1
  \end{minipage}%
}

\renewcommand{\techemptycell}{%
  \raisebox{-6pt}[0pt][0pt]{%
    \textcolor{black!32}{---}%
  }%
}

\renewcommand{\techmechanism}[2]{%
  \begin{minipage}[t]{\linewidth}
    \vspace{0pt}
    \centering
    #1\\[-0.10em]
    #2
  \end{minipage}%
}


\begin{table*}[t]
\centering

\caption{Representative implementation techniques across RSI autonomy levels (L1--L5).}
\label{tab:rsi-technique-space}

\fontsize{6.60pt}{6.65pt}\selectfont
\setlength{\tabcolsep}{0.55pt}
\renewcommand{\arraystretch}{1.00}
\setlength{\extrarowheight}{0pt}

\begin{tabularx}{\textwidth}{
  C{1.85cm}
  >{\centering\arraybackslash}p{3.70cm}
  *{5}{Y}
}

\toprule
\rowcolor{HeaderBlue}
\makecell[c]{%
  \textbf{Technique}\\[-0.06em]
  \textbf{Family}%
}
&
\makecell[c]{%
  \textbf{Concrete}\\[-0.06em]
  \textbf{Technique}%
}
& \levelhead{L1}{Execution}
& \levelhead{L2}{Strategy}
& \levelhead{L3}{Experience}
& \levelhead{L4}{Deployment}
& \levelhead{L5}{Meta}
\\

\midrule


\rowcolor{GroupBlue}
&
\techmechanism{A1. Evolutionary}{Search}
&
\techemptycell
&
\techworkcell{%
  Promptbreeder~\cite{fernando2024promptbreeder}\\[0.01em]
  AgentSquare~\cite{shang2025agentsquare}\\[0.01em]
  C-Evolve~\cite{li2026cevolve}%
}
&
\techemptycell
&
\techworkcell{%
  DecoEvo~\cite{chen2026decoevo}%
}
&
\techworkcell{%
  DGM~\cite{zhang2026dgm}\\[0.01em]
  RQGM~\cite{iacob2026redqueen}%
}
\\

\techruleblue

\rowcolor{GroupBlue}
&
\techmechanism{A2. Tree Search}{/ MCTS}
&
\techemptycell
&
\techworkcell{%
  AFlow~\cite{zhang2025aflow}%
}
&
\techemptycell
&
\techemptycell
&
\techemptycell
\\

\techruleblue

\rowcolor{GroupBlue}
&
\techmechanism{A3. Bayesian / Black-box}{Optimization}
&
\techemptycell
&
\techemptycell
&
\techemptycell
&
\techemptycell
&
\techemptycell
\\

\techruleblue

\rowcolor{GroupBlue}
&
\techmechanism{A4. LLM Reflection}{/ Critique}
&
\techworkcell{%
  EDIT~\cite{wu2026edit}%
}
&
\techworkcell{%
  GEPA~\cite{agrawal2026gepa}\\[0.01em]
  MPO~\cite{choi2026mpo}%
}
&
\techworkcell{%
  VOYAGER~\cite{wang2023voyager}%
}
&
\techworkcell{%
  ReasoningBank~\cite{ouyang2026reasoningbank}\\[0.01em]
  Trace2Skill~\cite{ni2026trace2skill}\\[0.01em]
  Metis~\cite{dai2026metis}%
}
&
\techemptycell
\\

\techruleblue

\rowcolor{GroupBlue}
\multirow[c]{-5}{1.85cm}[2.0em]{%
  \groupname{Search \&}{Optimization}%
}
&
\techmechanism{A5. LLM-guided Program}{/ Code Search}
&
\techworkcell{%
  Agent-Agnostic C/C++~\cite{lu2026agentagnostic}\\[0.01em]
  Harness Engineering~\cite{openai_harness_engineering}\\[0.01em]
  AIPC~\cite{su2026aipc}%
}
&
\techworkcell{%
  ADAS~\cite{hu2025adas}\\[0.01em]
  AFlow~\cite{zhang2025aflow}\\[0.01em]
  AutoKernel~\cite{jaber2026autokernel}%
}
&
\techworkcell{%
  VOYAGER~\cite{wang2023voyager}%
}
&
\techworkcell{%
  Metis~\cite{dai2026metis}\\[0.01em]
  HarnessDev~\cite{wu2026harnessdev}\\[0.01em]
  Tax AI~\cite{taxai}%
}
&
\techworkcell{%
  STOP~\cite{zelikman2024stop}\\[0.01em]
  HyperAgents~\cite{zhang2026hyperagents}\\[0.01em]
  AIDE$^{2}$~\cite{weco2026aide2}%
}
\\

\addlinespace[0pt]


&
\techmechanism{B1. Offline Synthetic}{Data Generation}
&
\techworkcell{%
  SynthLLM~\cite{synthllm}\\[0.01em]
  SynthAgent~\cite{wang-etal-2026-synthagent}\\[0.01em]
  Nemotron-4~\cite{nemotron4_340b}%
}
&
\techemptycell
&
\techemptycell
&
\techemptycell
&
\techemptycell
\\

\techrule

&
\techmechanism{B2. Self-Play / Adversarial}{Task Generation}
&
\techemptycell
&
\techemptycell
&
\techworkcell{%
  SSP~\cite{lu2026searchselfplay}\\[0.01em]
  AZR~\cite{zhao2025absolutezero}\\[0.01em]
  STP~\cite{dong2025stp}%
}
&
\techemptycell
&
\techemptycell
\\

\techrule

&
\techmechanism{B3. Adaptive Curriculum}{/ Task Scheduling}
&
\techemptycell
&
\techemptycell
&
\techworkcell{%
  SIMA 2~\cite{deepmind2025sima2}\\[0.01em]
  SEAgent~\cite{sun2026seagent}\\[0.01em]
  VOYAGER~\cite{wang2023voyager}%
}
&
\techemptycell
&
\techemptycell
\\

\techrule

&
\techmechanism{B4. Environment}{Generation}
&
\techemptycell
&
\techemptycell
&
\techemptycell
&
\techemptycell
&
\techemptycell
\\

\techrule

\multirow[c]{-5}{1.85cm}[2.5em]{%
  \groupnamethree{Data, Task \&}{Experience}{Construction}%
}
&
\techmechanism{B5. Autonomous Exploration}{/ Interaction}
&
\techworkcell{%
  SynthAgent~\cite{wang-etal-2026-synthagent}%
}
&
\techemptycell
&
\techworkcell{%
  SIMA 2~\cite{deepmind2025sima2}\\[0.01em]
  SEAgent~\cite{sun2026seagent}\\[0.01em]
  VOYAGER~\cite{wang2023voyager}%
}
&
\techemptycell
&
\techemptycell
\\

\addlinespace[0pt]


\rowcolor{GroupBlue}
&
\techmechanism{C1. Supervised Fine-Tuning}{(SFT)}
&
\techworkcell{%
  EDIT~\cite{wu2026edit}\\[0.01em]
  SynthAgent~\cite{wang-etal-2026-synthagent}\\[0.01em]
  Nemotron-4~\cite{nemotron4_340b}%
}
&
\techemptycell
&
\techworkcell{%
  STP~\cite{dong2025stp}\\[0.01em]
  PSV~\cite{wilf2026psv}%
}
&
\techemptycell
&
\techemptycell
\\

\techruleblue

\rowcolor{GroupBlue}
&
\techmechanism{C2. Preference Optimization}{(DPO / KTO)}
&
\techworkcell{%
  Nemotron-4~\cite{nemotron4_340b}%
}
&
\techemptycell
&
\techemptycell
&
\techemptycell
&
\techemptycell
\\

\techruleblue

\rowcolor{GroupBlue}
&
\techmechanism{C3. Reinforcement Learning}{(PPO / GRPO / RLVR)}
&
\techworkcell{%
  EDIT~\cite{wu2026edit}\\[0.01em]
  REPO~\cite{zeng-etal-2026-teaching}%
}
&
\techemptycell
&
\techworkcell{%
  AZR~\cite{zhao2025absolutezero}\\[0.01em]
  R-Zero~\cite{huang2026rzero}\\[0.01em]
  SEAgent~\cite{sun2026seagent}%
}
&
\techemptycell
&
\techemptycell
\\

\techruleblue

\rowcolor{GroupBlue}
&
\techmechanism{C4. Knowledge}{Distillation}
&
\techemptycell
&
\techemptycell
&
\techemptycell
&
\techemptycell
&
\techemptycell
\\

\techruleblue

\rowcolor{GroupBlue}
&
\techmechanism{C5. Prompt / Context}{Optimization}
&
\techemptycell
&
\techworkcell{%
  Promptbreeder~\cite{fernando2024promptbreeder}\\[0.01em]
  GEPA~\cite{agrawal2026gepa}\\[0.01em]
  MPO~\cite{choi2026mpo}%
}
&
\techemptycell
&
\techworkcell{%
  DecoEvo~\cite{chen2026decoevo}\\[0.01em]
  ACE~\cite{zhang2026agentic}\\[0.01em]
  PersonaAgent~\cite{zhang2026personaagent}%
}
&
\techemptycell
\\

\techruleblue

\rowcolor{GroupBlue}
\multirow[c]{-6}{1.85cm}[2.0em]{%
  \groupnamethree{Training \&}{Persistent}{Update}%
}
&
\techmechanism{C6. Memory / Skill}{Consolidation}
&
\techemptycell
&
\techemptycell
&
\techworkcell{%
  VOYAGER~\cite{wang2023voyager}\\[0.01em]
  SEAgent~\cite{sun2026seagent}%
}
&
\techworkcell{%
  ReasoningBank~\cite{ouyang2026reasoningbank}\\[0.01em]
  Metis~\cite{dai2026metis}\\[0.01em]
  Trace2Skill~\cite{ni2026trace2skill}%
}
&
\techemptycell
\\

\addlinespace[0pt]


&
\techmechanism{D1. Execution / Unit-Test}{Verification}
&
\techworkcell{%
  Agent-Agnostic C/C++~\cite{lu2026agentagnostic}\\[0.01em]
  AIPC~\cite{su2026aipc}\\[0.01em]
  Harness Engineering~\cite{openai_harness_engineering}%
}
&
\techworkcell{%
  AutoKernel~\cite{jaber2026autokernel}%
}
&
\techworkcell{%
  AZR~\cite{zhao2025absolutezero}\\[0.01em]
  VOYAGER~\cite{wang2023voyager}%
}
&
\techworkcell{%
  Metis~\cite{dai2026metis}\\[0.01em]
  HDSO~\cite{shang2026hypothesis}\\[0.01em]
  Tax AI~\cite{taxai}%
}
&
\techworkcell{%
  STOP~\cite{zelikman2024stop}\\[0.01em]
  AIDE$^{2}$~\cite{weco2026aide2}%
}
\\

\techrule

&
\techmechanism{D2. Formal}{Verification}
&
\techemptycell
&
\techemptycell
&
\techworkcell{%
  STP~\cite{dong2025stp}\\[0.01em]
  PSV~\cite{wilf2026psv}%
}
&
\techemptycell
&
\techemptycell
\\

\techrule

&
\techmechanism{D3. Reward Model}{/ LLM-as-a-Judge}
&
\techworkcell{%
  Nemotron-4~\cite{nemotron4_340b}\\[0.01em]
  REPO~\cite{zeng-etal-2026-teaching}\\[0.01em]
  HealthBench~\cite{healthbench}%
}
&
\techemptycell
&
\techworkcell{%
  SIMA 2~\cite{deepmind2025sima2}\\[0.01em]
  SEAgent~\cite{sun2026seagent}%
}
&
\techworkcell{%
  ReasoningBank~\cite{ouyang2026reasoningbank}\\[0.01em]
  DecoEvo~\cite{chen2026decoevo}%
}
&
\techworkcell{%
  RQGM~\cite{iacob2026redqueen}%
}
\\

\techrule

\multirow[c]{-4}{1.85cm}[2.3em]{%
  \groupname{Verification \&}{Selection}%
}
&
\techmechanism{D4. Regression Testing}{/ Model Selection}
&
\techworkcell{%
  Agent-Agnostic C/C++~\cite{lu2026agentagnostic}\\[0.01em]
  Harness Engineering~\cite{openai_harness_engineering}%
}
&
\techworkcell{%
  GEPA~\cite{agrawal2026gepa}\\[0.01em]
  AutoResearch~\cite{karpathy2026autoresearch}\\[0.01em]
  AutoKernel~\cite{jaber2026autokernel}%
}
&
\techemptycell
&
\techworkcell{%
  HarnessDev~\cite{wu2026harnessdev}\\[0.01em]
  ASPIRE~\cite{wu2026aspire}\\[0.01em]
  HDSO~\cite{shang2026hypothesis}%
}
&
\techworkcell{%
  STOP~\cite{zelikman2024stop}\\[0.01em]
  RQGM~\cite{iacob2026redqueen}\\[0.01em]
  AIDE$^{2}$~\cite{weco2026aide2}%
}
\\

\addlinespace[0pt]


\rowcolor{GroupBlue}
&
\techmechanism{E1. Online / Continual}{Learning}
&
\techemptycell
&
\techemptycell
&
\techworkcell{%
  SEAgent~\cite{sun2026seagent}\\[0.01em]
  VOYAGER~\cite{wang2023voyager}%
}
&
\techworkcell{%
  ReasoningBank~\cite{ouyang2026reasoningbank}\\[0.01em]
  PANDO~\cite{li2026pando}\\[0.01em]
  Dynamic Cheatsheet~\cite{suzgun2026dynamic}%
}
&
\techemptycell
\\

\techruleblue

\rowcolor{GroupBlue}
&
\techmechanism{E2. Experience Replay}{/ Online Memory Update}
&
\techemptycell
&
\techemptycell
&
\techemptycell
&
\techworkcell{%
  ReasoningBank~\cite{ouyang2026reasoningbank}\\[0.01em]
  PANDO~\cite{li2026pando}\\[0.01em]
  Metis~\cite{dai2026metis}%
}
&
\techemptycell
\\

\techruleblue

\rowcolor{GroupBlue}
&
\techmechanism{E3. Self-Modifying}{Code / Policy}
&
\techemptycell
&
\techworkcell{%
  ADAS~\cite{hu2025adas}\\[0.01em]
  AFlow~\cite{zhang2025aflow}\\[0.01em]
  AutoResearch~\cite{karpathy2026autoresearch}%
}
&
\techemptycell
&
\techworkcell{%
  HarnessDev~\cite{wu2026harnessdev}\\[0.01em]
  ASPIRE~\cite{wu2026aspire}\\[0.01em]
  SHAPER~\cite{wang2026shaper}%
}
&
\techworkcell{%
  STOP~\cite{zelikman2024stop}\\[0.01em]
  G\"odel Agent~\cite{yin2025godelagent}\\[0.01em]
  A-Evolve-Training~\cite{shi2026aevolvetraining}%
}
\\

\techruleblue

\rowcolor{GroupBlue}
\multirow[c]{-4}{1.85cm}[3.2em]{%
  \groupnamethree{Online \&}{Meta-Level}{Adaptation}%
}
&
\techmechanism{E4. Adaptive / Co-Evolving}{Evaluator}
&
\techemptycell
&
\techemptycell
&
\techemptycell
&
\techworkcell{%
  DecoEvo~\cite{chen2026decoevo}%
}
&
\techworkcell{%
  RQGM~\cite{iacob2026redqueen}%
}
\\

\bottomrule
\end{tabularx}


\vspace{0.16em}

\begin{minipage}{0.985\textwidth}
\fontsize{6.8pt}{7.0pt}\selectfont
\color{MutedText}

\textbf{Reading guide.}
Rows group concrete techniques into broader technical families;
columns indicate the autonomy levels of the specific improvement
loops examined in the cited works.
A work may appear in multiple cells when it employs several techniques
or contains distinct improvement loops.
Each cell lists up to three representative works.
An em dash (---) indicates that no representative example is listed here;
it does not imply incompatibility between the technique and the level.

\vspace{0.10em}

\textbf{Level shorthand.}
L1: improvement execution \(\cdot\)
L2: improvement strategy \(\cdot\)
L3: learning-signal / experience acquisition \(\cdot\)
L4: environment adaptation \(\cdot\)
L5: recursive inheritance (meta-improvement).

\end{minipage}
\end{table*}
\clearpage

For example, a coding agent may repeatedly revise its code using a human-provided test suite and a fixed limit of five attempts. It remains at B0 if the fixes and feedback are confined to the current task and do not change how the agent handles subsequent tasks. Similarly, a writing agent may revise an essay against a human-provided rubric without retaining the resulting experience for future writing. In both cases, humans specify the evaluation rules and revision procedure, while AI improves only the current output within those constraints.

Under fixed rules, B0 can improve output in single sessions. For instance, Self-Refine improves outputs through a generate–feedback–refine loop within the same session~\cite{madaan2023self}; Reflexion writes verbal reflections on failed attempts into a context buffer for subsequent tries~\cite{shinn2023reflexionlanguageagentsverbal}; Tree of Thoughts organizes reasoning as a search over candidate thought branches~\cite{yao2023tree}. 

However, B0 cannot accumulate knowledge across tasks. Iterative traces and correction signals exist only within the current session; after task termination they are discarded, and the system cannot consolidate fragmented experience into long-term capability. APEX-EM notes that LLM-based agents generally lack persistent procedural memory: even after solving an identical task, they must re-derive the solution from scratch~\cite{banerjee2026apexemnonparametriconlinelearning}. Specifically, there are four main limitations.

\noindent$\bullet$ \bfit{Experience does not accumulate across tasks.} Intermediate results, critiques, and corrections remain in temporary task context rather than becoming updates to model parameters, reusable memory, or operating rules~\cite{zhao2024expel}. A system may therefore correct an error in one task and repeat it in the next. Increasing the number of iterations cannot resolve this lack of persistent learning.

\noindent$\bullet$ \bfit{Improvement procedure remains human-defined.} AI can generate, verify, and revise outputs, but humans still determine how this process operates~\cite{madaan2023self}. For example, a coding agent may fix code that fails a test, but it does not persistently improve the testing procedure for future tasks. The loop closes around the current output; responsibility for upgrading the system's improvement mechanisms remains with humans. Tool use, simulation, and real-world feedback do not by themselves change this boundary.

\noindent$\bullet$ \bfit{Self-generated feedback can reinforce errors.} When the same model generates and evaluates an answer, both stages may share the same knowledge gaps. Self-Correction Bench identifies verification, particularly locating the first incorrect reasoning step, as a key bottleneck~\cite{tsui2026selfcorrectionbenchuncoveringaddressing,tyen2024finderrors}. Without reliable external feedback, revision can even turn a correct answer into an incorrect one~\cite{huang2024cannotselfcorrect}. Repeated critique therefore does not guarantee successful correction.

\noindent$\bullet$ \bfit{More iterations offer limited gains under fixed evaluation.}
Revising an answer without new information may provide little additional benefit~\cite{li2026decomposing}; independent sampling or external verification can be more effective in some settings~\cite{olausson2024self,verma2026blindresampling}. A fixed verifier also leaves some errors undetected~\cite{liu2023evalplus}. For example, repeatedly revising code to pass an incomplete test suite may improve test performance while leaving untested failures unresolved.

\subsection{L1: Autonomy over Improvement Execution}
\label{sec:l1-improvement-execution}

At L1, \AISystem{} executes a human-defined improvement procedure whose accepted
results are retained and reused in later tasks or improvement rounds. Humans specify the objective, update procedure, and acceptance criteria, while AI carries out the prescribed improvement steps. Unlike B0, the loop changes persistent state within the specified AI system rather than only refining the current task output. For example, a coding agent may follow a human-defined procedure to revise code based on test feedback, save validated fixes as reusable rules, and automatically apply them to subsequent independent tasks. This pattern is already visible in production infrastructure. Meta's Capacity Efficiency system encodes engineers' debugging expertise into reusable repair skills and applies these skills through predefined procedures to diagnose performance regressions and perform remediation tasks. Meta reports that this workflow compresses hours of manual regression investigation into minutes, and the resulting pull requests undergo standard review and integration, creating persistent improvements in the production system~\cite{meta_capacity_efficiency_2026}.

The characteristic loop at this level can be summarized as follows:

\begin{center}
\fcolorbox{black!25}{gray!4}{%
\begin{minipage}{0.92\linewidth}
\centering
\vspace{6pt}

\LoneStep{receive improvement objective}
\(\;\longrightarrow\;\)
\LoneStep{execute prescribed procedure}
\(\;\longrightarrow\;\)
\LoneStep{produce improvement}

\vspace{7pt}

\(\longrightarrow\;\)
\LoneStep{update system state}
\(\;\longrightarrow\;\)
\LoneStep{process the next task}
\(\;\longrightarrow\;\)
\LoneStep{repeat}

\vspace{6pt}
\end{minipage}%
}
\end{center}

Despite its simple structure, executing this loop reliably presents several core challenges. First, the prescribed procedure must cover the situations that arise during execution. When the task encounters a condition that is not addressed by the predefined procedure, the agent lacks an applicable next step and cannot independently modify the procedure to resolve it. Second, execution must remain reliable across multi-step workflows, since an incorrect diagnosis, transformation, or intermediate decision can influence all subsequent actions. Third, because the resulting improvement is written back into the system state and reused in later tasks, its validity must be checked before retention; otherwise, execution errors can persist and affect subsequent development cycles. These challenges recur across the AI development pipeline, but their concrete form depends on the stage in which the improvement procedure is executed.

Specifically, large-scale AI development in real scenarios is typically organized into several distinct levels: (1) the data level governs the construction and curation of training corpora; (2) the training-method level determines how models learn; (3) the training-platform level manages the execution of large-scale training; (4) the evaluation-and-safety level assesses model capabilities and safety; (5) the deployment-optimization level enables efficient model serving; and (6) the application-system level integrates foundation-model capabilities into downstream products and application systems. The following sections examine how the \AISystem{} autonomously executes predefined improvement procedures within each level and how the resulting artifacts feed into downstream stages of the development workflow. Table~\ref{tab:l1-industrial-cases} summarizes the representative systems discussed across these levels.

\begin{table*}[t]
\centering
\caption{Representative L1 systems organized by AI development pipeline levels. L1 systems execute human-defined improvement procedures and produce persistent artifacts that enter subsequent workflows.}
\label{tab:l1-industrial-cases}
\scriptsize
\setlength{\tabcolsep}{3pt}
\renewcommand{\arraystretch}{1.05}
\resizebox{\textwidth}{!}{%
\begin{tabular}{@{}llll@{}}
\toprule
\textbf{Method} & \textbf{Type} &
\textbf{Key Mechanism} & \textbf{Level Limitation}\\
\midrule

\multicolumn{4}{@{}l}{\textbf{Data Level}}\\

Google High-Fidelity Label Curation~\cite{google_high_fidelity_labels}
& Data curation
& LLM labeling + boundary selection
& Human-defined labeling rules\\

Phi-4-reasoning~\cite{phi4_reasoning}
& Training data selection
& LLM evaluation + boundary filtering
& Human-defined selection criteria\\

FineWeb-Edu~\cite{penedo2024fineweb}
& Data filtering
& LLM scoring + classifier filtering
& Human-defined quality criteria\\

NVIDIA NeMo Curator~\cite{nemo_curator}
& Data curation
& Configurable filtering pipeline
& Human-configured processing rules\\

Data-Juicer~\cite{data_juicer}
& Data processing
& Composable processing operators
& Human-defined processing recipe\\

Microsoft SynthLLM~\cite{synthllm}
& Synthetic data generation
& Reference-guided data synthesis
& Human-defined generation procedure\\

SynthAgent~\cite{wang-etal-2026-synthagent}
& Supervision generation
& Task + trajectory synthesis
& Human-defined generation workflow\\

NVIDIA Nemotron-4~\cite{nemotron4_340b}
& Synthetic supervision
& Candidate generation + reward filtering
& Human-defined quality dimensions\\

\midrule

\multicolumn{4}{@{}l}{\textbf{Training-Method Level}}\\

EDIT~\cite{wu2026edit}
& Training signal refinement
& Diagnosed step revision
& Human-defined diagnostic criteria\\

REPO~\cite{zeng-etal-2026-teaching}
& Post-training procedure
& SOP-guided interaction + reward evaluation
& Human-defined behavioral procedure\\

\midrule

\multicolumn{4}{@{}l}{\textbf{Training-Platform Level}}\\

Agent-Agnostic C/C++ Optimization~\cite{lu2026agentagnostic}
& Training execution optimization
& Procedural performance optimization loop
& Human-defined optimization workflow\\

Meta NCCL Agentic Debugging~\cite{meta_nccl_agentic_debugging}
& Failure diagnosis and recovery
& Runbook-guided distributed debugging
& Human-defined debugging procedure\\

\midrule

\multicolumn{4}{@{}l}{\textbf{Evaluation-and-Safety Level}}\\

OpenAI HealthBench~\cite{healthbench}
& Model evaluation
& Expert-rubric-based automated scoring
& Human-defined evaluation rubrics\\

\midrule

\multicolumn{4}{@{}l}{\textbf{Deployment-Optimization Level}}\\

AIPC~\cite{su2026aipc}
& Model deployment
& Skill-guided deployment adaptation
& Human-defined deployment procedure\\

\midrule

\multicolumn{4}{@{}l}{\textbf{Application-System Level}}\\

Agent Toolkit for AWS~\cite{aws_agent_toolkit}
& Model-application integration
& Bedrock skill-guided integration
& Human-defined integration procedures\\

LinkedIn CAPT~\cite{linkedin_capt}
& Product engineering
& Step-by-step engineering playbooks
& Human-authored implementation procedures\\

OpenAI Harness Engineering~\cite{openai_harness_engineering}
& Product development
& Constraint-guided implementation
& Human-defined architecture and delivery rules\\

\bottomrule
\end{tabular}%
}
\end{table*}

\subsubsection{Data Level}
At the data level, AI increasingly takes over the execution of large-scale data production and curation procedures that were previously carried out through substantial manual effort. Existing approaches mainly fall into two classes. The first focuses on \emph{data cleaning and quality filtering}, and the second focuses on \emph{synthetic data and supervision generation}, where humans define the target task, desired data characteristics, and validation procedures.

\noindent$\bullet$ \underline{(1) {Data cleaning and quality filtering.}} A primary challenge at the data level is scale: the volume of training corpora and candidate samples far exceeds the capacity for manual review at the sample level. Data curation is therefore shifting from sample-by-sample human judgment toward a paradigm in which humans define quality criteria and processing procedures, while AI executes them at scale. In Google's high-fidelity label curation pipeline, developers first define the target task and its decision criteria, such as what constitutes clickbait. An LLM then applies these criteria to assign preliminary labels to candidate data, after which a predefined clustering and boundary-sample selection procedure identifies a small subset of informative samples for expert annotation~\cite{google_high_fidelity_labels}. A similar division of labor appears in the construction of training data for Phi-4-reasoning: researchers specify the desired difficulty and reasoning characteristics together with their evaluation procedures, while LLM-based evaluation pipelines apply these criteria to select samples near the model's capability boundary~\cite{phi4_reasoning}. FineWeb-Edu extends this pattern to large-scale web filtering: researchers define educational-quality criteria, use an LLM to generate quality labels, and train a classifier to apply these criteria at scale~\cite{penedo2024fineweb}. As these data-processing operations become standardized, they can also be incorporated into reusable pipelines. NVIDIA NeMo Curator organizes quality filtering, classification, and deduplication into configurable modules that execute according to developer-specified rules and parameters~\cite{nemo_curator}. Data-Juicer follows a similar approach but abstracts data-processing operations into composable operators, allowing predefined processing recipes to be executed over large-scale corpora~\cite{data_juicer}.

\noindent$\bullet$ \underline{(2) {Synthetic data and supervision signal generation.}} Training data can be expanded through synthetic generation. High-quality demonstrations for complex tasks are often costly to construct manually, motivating the use of repeatable data-generation procedures that models can execute at scale. Microsoft SynthLLM follows this approach by starting from high-quality web content, organizing reference concepts through a predefined multi-stage procedure, and using LLMs to generate diverse prompts and corresponding responses as synthetic training examples~\cite{synthllm}. Similar procedures can also produce richer forms of supervision. SynthAgent organizes the construction of tasks and interaction trajectories into a predefined pipeline, in which models perform web exploration, task generation, and trajectory refinement to produce supervision data for training web agents~\cite{wang-etal-2026-synthagent}. Synthetic generation can be combined with predefined quality evaluation to construct higher-quality supervision data. NVIDIA Nemotron-4 uses an instruction model to generate candidate responses and a reward model to score and filter them according to predefined quality dimensions, with the resulting synthetic samples used for subsequent model training~\cite{nemotron4_340b}.

\subsubsection{Training-Method Level}
At the training-method level, AI increasingly takes over the execution of structured learning and post-training procedures that were previously orchestrated manually by researchers and engineers. Existing approaches mainly encode diagnostic, revision, evaluation, and optimization steps into predefined workflows, where humans specify the learning objectives, feedback signals, and update rules, while AI systems execute these procedures to generate improved training signals and support subsequent model optimization.

Established training practices can be encoded into explicit improvement procedures and executed by AI systems during post-training, improving how models receive training signals and thereby enhancing learning outcomes. EDIT organizes this process as a two-stage training procedure. Researchers first specify the diagnostic signals and revision criteria, after which the system identifies problematic reasoning steps and an LLM revises only the affected parts according to a rubric checklist. The revised outputs are then used for further reinforcement-learning calibration~\cite{wu2026edit}. REPO similarly incorporates a multi-stage standard operating procedure and behavioral constraints into the training process. The model follows the prescribed steps during multi-turn interactions, while an LLM-based judge evaluates procedural compliance according to predefined criteria and converts the resulting assessments into reward signals for subsequent policy optimization~\cite{zeng-etal-2026-teaching}.

\subsubsection{Training-Platform Level}
At the training-platform level, AI increasingly takes over the execution of infrastructure engineering procedures required to keep large-scale training efficient and reliable. Existing approaches mainly fall into two classes. The first focuses on \emph{training execution and performance optimization}, where agents analyze workloads and apply predefined optimization procedures, and the second focuses on \emph{failure diagnosis and recovery}, where agents execute structured debugging and remediation workflows distilled from engineering knowledge and operational experience.

\noindent$\bullet$ \underline{(1) {Training execution and optimization.}} 
Performance-engineering practices can be encoded into explicit optimization procedures that AI agents execute on concrete workloads. Agent-Agnostic End-to-End C/C++ Application Performance Optimization follows this pattern by storing the complete optimization control loop in procedural memory. The agent performs runtime analysis, identifies performance hotspots, generates candidate code modifications, and validates both correctness and performance according to the predefined workflow, with unsuccessful changes rolled back before further optimization~\cite{lu2026agentagnostic}.

\noindent$\bullet$ \underline{(2) {Training failure diagnosis and recovery.}} 
Distributed-training failures can similarly be handled through structured debugging procedures derived from accumulated engineering experience. Meta and the PyTorch community categorized the major root causes of NCCL watchdog timeouts and distilled these findings into a practical decision tree and debugging runbook. An AI agent then executes this structured workflow on concrete failures by aligning evidence across ranks, tracing collective sequences, and connecting runtime traces to the corresponding code paths in order to locate the earliest actionable divergence~\cite{meta_nccl_agentic_debugging}.

\subsubsection{Evaluation-and-Safety Level}
At the evaluation-and-safety level, AI increasingly takes over the execution of model assessment procedures that translate expert-defined requirements into scalable and repeatable evaluations. Existing approaches use AI systems to apply predefined rubrics, test protocols, and risk criteria across large numbers of model outputs, while humans remain responsible for defining the evaluation objectives, criteria, and acceptance thresholds.

Model evaluation transforms expert judgment into repeatable testing and scoring procedures. OpenAI HealthBench asks medical experts to define detailed evaluation criteria for specific healthcare conversations, specifying what an ideal response should include or avoid and the relative importance of each criterion. GPT-4.1 then applies these predefined criteria to score candidate model responses and compute the final evaluation result~\cite{healthbench}. Similar evaluation procedures can also be applied to robustness, safety, and risk evaluation, providing standardized feedback for subsequent model improvement and release decisions.

\subsubsection{Deployment-Optimization Level}
At the deployment-optimization level, AI increasingly takes over the execution of procedures for adapting trained models to concrete hardware and runtime environments. Existing approaches encode deployment expertise into standardized workflows for model conversion, compatibility resolution, quantization, runtime configuration, and validation, where humans specify the target platform and deployment constraints, while AI systems execute the corresponding adaptation and verification steps.

Deploying trained models to target hardware often requires specialized procedures for model conversion and runtime adaptation. AIPC encodes this deployment expertise into standardized, verifiable stages supported by Agent Skills and stage-wise validation. Given a trained model and a target Qualcomm AI Runtime environment, the AI agent executes the predefined deployment procedure to convert the model, handle deployment incompatibilities, perform quantization calibration, and validate the resulting implementation. The process produces runnable deployment artifacts for downstream inference~\cite{su2026aipc}.

\subsubsection{Application-System Level}
At the application-system level, AI increasingly takes over the execution of software-engineering procedures required to integrate foundation models into operational products and services. Existing approaches encode development knowledge into reusable skills, playbooks, and engineering workflows, allowing AI systems to perform tasks such as service integration, interface implementation, code modification, testing, and validation, while humans continue to specify product requirements, system architecture, and the governing development constraints.

Transforming foundation-model capabilities into practical products requires model services and interfaces to be integrated into application systems through repeatable engineering procedures. Agent Toolkit for AWS encodes generative AI development practices into reusable Agent Skills. Its Amazon Bedrock skill provides predefined procedures for integrating model capabilities into applications. AI coding agents can execute these procedures to connect foundation-model services with downstream product interfaces and application components~\cite{aws_agent_toolkit}. LinkedIn's Contextual Agent Playbooks \& Tools (CAPT) similarly encodes accumulated organizational knowledge into step-by-step playbooks, enabling AI coding agents to carry out concrete engineering tasks in real codebases (e.g., service creation, interface extension, and code maintenance)~\cite{linkedin_capt}. OpenAI Harness Engineering extends this pattern to end-to-end software product development. Engineers define product intent, system architecture, and the rules governing continuous integration, code merging, and rollback, while Codex performs implementation and validation within these predefined constraints and continuously produces code artifacts that enter subsequent development workflows~\cite{openai_harness_engineering}.

\subsubsection{Evaluating L1 Execution Autonomy}
Across these levels, L1 systems share the same fundamental capability boundary: AI can execute increasingly long and consequential improvement procedures, and the resulting artifacts may persist into subsequent stages, but the procedures governing improvement remain externally specified. Evaluation of L1 should therefore consider two coupled properties: \emph{execution reliability}, namely whether AI can correctly carry out the prescribed procedure under varying conditions; and \emph{persistence safety}, namely whether errors introduced during execution are detected before their outputs propagate into subsequent improvement rounds.

The reliability of predefined improvement procedures is central to evaluating execution autonomy at this level. L1 systems can complete extended sequences of actions with limited human intervention, but their execution remains bounded by procedures specified in advance. When task conditions fall outside the scope covered by these procedures, or when assumptions embedded in the workflow no longer hold, the system cannot independently revise the improvement procedure and is therefore prone to execution failure.

Persistence further amplifies the consequences of such failures. Outputs produced during execution can be retained and reused in subsequent development stages, allowing errors to propagate beyond the task in which they were introduced and affect later iterations. Evaluation should therefore examine not only whether predefined procedures are executed correctly, but also whether the resulting artifacts remain valid before entering downstream workflows.

\begin{center}
\fbox{
\parbox{0.92\linewidth}{
\textbf{Finding: L1 executes human-defined improvement procedures at scale.}
Humans first encode established engineering experience into explicit improvement steps and validation rules, while AI systems execute these predefined procedures on concrete tasks. The resulting artifacts are retained and incorporated into later development workflows, allowing the same improvement procedures to be repeatedly applied across independent tasks and development stages.
}}
\end{center}

\subsection{L2: Autonomy over Improvement Strategies}
\label{sec:l2-strategy}

At L2, the AI system uses evaluation feedback to choose which improvement
intervention to attempt next, rather than merely executing a prescribed
update.
It proposes and tests changes to the target and retains accepted
revisions, while humans continue to specify the objective, task boundary, and acceptance criteria.
The surrounding search framework may remain fixed; autonomy lies in determining how the target should be improved.
In many AI-development workflows, executing a proposed change is comparatively mechanical; the difficult part is deciding which change is worth trying. Researchers must interpret failures, infer which component is limiting progress, choose an intervention whose effect is uncertain, and decide what to try next after observing the result. As the space of possible interventions grows, this experimental decision-making becomes a major bottleneck: exhaustive search is infeasible, while manual experimentation is constrained by researcher attention and prior intuitions.
L2 systems automate this bottleneck. Rather than merely carrying out a human-specified modification, the \Improver \ uses evaluation results and experimental evidence to determine the next candidate intervention.

This transition increasingly resembles partial automation of the empirical researcher's role. OpenAI, for example, describes current frontier agents as reaching an ``automated research intern'' stage for well-defined research tasks under human supervision, while researchers continue to set priorities and decide which results merit further investment or deployment~\cite{openai2026researchacceleration}. The relevant autonomy is therefore not unrestricted control of research, but the transfer of a specific research responsibility: turning observed evidence into the next experiment.

The characteristic loop at this level can be summarized as follows:
\begin{center}
\fcolorbox{black!25}{gray!4}{%
\begin{minipage}{0.72\linewidth}
\centering
\vspace{6pt}

\LoneStep{observe performance}
\(\;\longrightarrow\;\)
\LoneStep{diagnose}
\(\;\longrightarrow\;\)
\LoneStep{select how change}

\vspace{7pt}

\(\;\longrightarrow\;\)
\LoneStep{instantiate and test}
\(\;\longrightarrow\;\)
\LoneStep{retain or revert}
\(\;\longrightarrow\;\)
\LoneStep{repeat}

\vspace{6pt}
\end{minipage}%
}
\end{center}

\begin{figure*}[!t]
    \centering
    \includegraphics[width=1\linewidth]{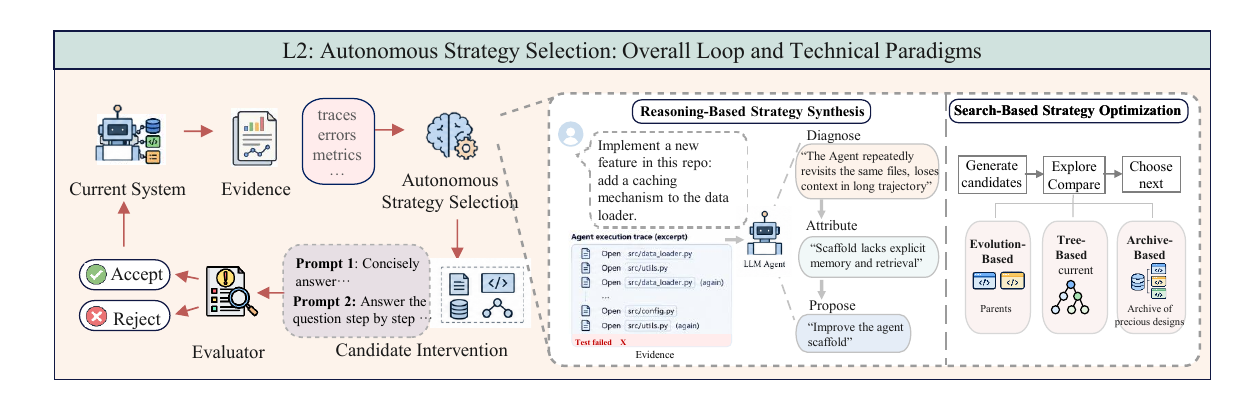}
    \caption{Overview of L2: Autonomous Strategy Selection. Under human-defined improvement objectives and evaluation criteria, the AI system uses evidence from the current system to diagnose weaknesses and autonomously determine how to improve it. Candidate interventions are evaluated, and accepted updates are incorporated into subsequent iterations. Representative mechanisms include reasoning-based strategy synthesis and search-based strategy optimization.}
    \label{fig:L2}
\end{figure*}

The object being improved may remain small, as in prompt optimization, or expand to an executable agent and eventually to a model-training program. What changes across these settings is the object and scope of intervention available to the improver, together with the evidence available for deciding which modification to attempt. We therefore organize L2 by the \emph{object over which improvement strategies are searched}, rather than by individual systems.
Figure~\ref{fig:L2} illustrates this
strategy-autonomy mechanism, in which evidence about the current system
guides the autonomous selection and evaluation of how to improve it.

\subsubsection{Prompt Search}
\noindent\textbf{The \Improver \ decides how the instructions that govern a model or agent should be changed based on execution traces and evaluation results.} In the narrowest L2 setting, the system decides how a prompt or textual control variable should be revised rather than merely instantiating a prescribed edit. For example, Dropbox used GEPA to rewrite the prompt that tells an LLM how to judge whether retrieved files or messages are relevant to a user's search~\cite{dropbox2026dspy}.
However, earlier evolutionary methods search over alternative mutations, while more recent reflective optimizers use execution traces to identify failure modes and generate targeted revisions. GEPA attributes failures to particular modules and preserves complementary prompt variants rather than committing immediately to a single greedy lineage~\cite{agrawal2026gepa}. This feedback-driven search is not restricted to a single optimizer design. MPO extends the editable surface to multimodal prompts, using evaluation-derived semantic feedback to guide subsequent textual and non-textual prompt candidates, while C-Evolve embeds candidate generation within a fixed evolutionary protocol. Across these settings, the search machinery can be externally specified while the concrete prompt revisions emerge from evaluation feedback.~\cite{choi2026mpo,li2026cevolve}.
The relevant increase in autonomy is therefore not prompt generation itself, but control over which revision strategy is attempted next.

\subsubsection{Agent and Harness Search}
\noindent\textbf{Improvement can involve redesigning how reasoning and tool-using components interact.} Once the editable object expands from a prompt to executable agent code, strategy search begins to resemble automated system design.  
For example, Microsoft Foundry's Agent Optimizer rewrites hosted agents' system instructions, skills, and local function-tool descriptions from evaluation failures~\cite{microsoft2026agentoptimizer}.
Automated Design of Agentic Systems (ADAS) formalizes an agent as a Python \texttt{forward} function and uses a meta-agent to populate an ever-growing archive of candidate agents~\cite{hu2025adas}. In each iteration, the meta-agent evaluates the candidate on validation data and stores both its code and metrics in the archive.
Because these decisions are encoded in executable harness code, a single revision can change how the agent behaves throughout an entire task rather than only changing one instruction.

Recent workflow-search systems instantiate this idea through different representations of the design space. 
AFlow encodes workflows as executable graphs and applies Monte Carlo tree search to successive code-level modifications, using execution outcomes as feedback for later branches~\cite{zhang2025aflow}.
AgentSquare instead factorizes agent designs into reusable modules and searches through evolution and recombination, allowing successful structures discovered in one configuration to become building blocks for another~\cite{shang2025agentsquare}.

The resulting autonomy is substantially greater than local parameter tuning while remaining bounded in an important sense. The improver chooses among alternative agent designs, yet candidates are still promoted because they perform better under an externally maintained evaluation process. L2 autonomy concerns the search for a better system configuration; it does not imply authority to redefine what constitutes successful performance.

\subsubsection{Model and Training Search}
\noindent\textbf{The \Improver \ converts empirical training evidence into decisions about how the learner itself should be configured, structured, or optimized.}
Human researchers therefore rely heavily on accumulated intuition to decide which small fraction of the possible experiments to run. The emerging role of an L2 improver is to compress this experimental cycle: inspect what happened in previous runs, infer where improvement is likely to come from, and allocate the next experiment accordingly. Current systems increasingly search over persistent configurations, model structure, and the learning procedure itself.

\noindent$\bullet$ \underline{(1) {Configuration Search.}}
{It automates the allocation of expensive training trials. }
Recent agentic approaches increasingly augment conventional search with semantic reasoning about previous experiments. Rather than treating each trial as an isolated black-box query, the system can interpret training outcomes and use them to concentrate subsequent exploration. For example, NVIDIA's 2026 TAO workflow for post-training Cosmos 3 lets a coding agent invoke LLM-guided AutoML under a fixed validation objective~\cite{nvidia2026taoautoml}.
The human supplies the model, task, data, and desired metric; the system takes over much of the repeated decision of which configuration should be evaluated next.

\noindent$\bullet$ \underline{(2) {Architecture Search.}}
It delegates the structural design of the model, reducing reliance on a human-engineered search space.
LLM-based improvers can interpret empirical results and propose structural changes whose exact content was not enumerated beforehand. 
AgentNAS starts from the observation that modern NAS still depends on manually engineered, task-specific search spaces. It uses an LLM to produce a task-specific seed architecture and then derives a structured search space from that design for subsequent combinatorial search~\cite{jeong2026agentnas}.

\noindent$\bullet$ \underline{(3) {Training-Rule Search.}}
The \Improver\ changes the executable procedure by which model parameters are learned.
Current frontier agents are beginning to automate this hypothesis-to-experiment step. OpenAI's NanoGPT evaluation for GPT-6 Astra gives the agent a fixed validation target, one GPU, and a constrained training setup, but requires the agent itself to diagnose training bottlenecks, modify the training code, tune its configuration, and make useful changes to the training loop~\cite{openai2026gpt6astra}.
Open research environments expose the same division of labor more directly. In \texttt{autoresearch}, the data pipeline, evaluation function, metric, and per-experiment compute budget remain fixed, while the agent repeatedly edits the training program and keeps a modification only when the protected validation metric improves~\cite{karpathy2026autoresearch}.

\noindent$\bullet$ \underline{(4) {Systems and Implementation Search.}}
The \Improver \ searches for a more efficient realization of a model's computation while correctness and system-level performance remain protected by external tests.
AI can profile a bottleneck, propose an implementation change, verify numerical correctness, measure real hardware performance, and use the result to decide what to modify next.
AutoKernel begins from profiling rather than arbitrary code generation: it identifies operations with the largest potential end-to-end impact and directs the agent's experiments toward those bottlenecks. Candidate Triton or CUDA implementations are accepted only after correctness checks and measured GPU speedups~\cite{jaber2026autokernel}.

Table~\ref{tab:l2-strategy-summary} compares representative systems by their search target, mechanism, and externally fixed limitation.

\begin{table*}[t]
\centering
\caption{Representative (rather than exhaustive) L2 systems grouped by the object of improvement-strategy search. All listed systems remain at L2 because they autonomously choose how to improve the system while the objective, evaluation criterion, or acceptance rule remains externally specified.}
\label{tab:l2-strategy-summary}
\scriptsize
\setlength{\tabcolsep}{1.2pt}
\renewcommand{\arraystretch}{1.05}
\begin{tabularx}{0.9\textwidth}{@{}>{\raggedright\arraybackslash}p{0.205\textwidth}>{\raggedright\arraybackslash}p{0.195\textwidth}>{\raggedright\arraybackslash}p{0.3\textwidth}>{\raggedright\arraybackslash}X@{}}
\toprule
\textbf{Method} 
& \textbf{Type} 
& \textbf{Key mechanism} 
& \textbf{Level limitation}\\

\midrule
\multicolumn{4}{@{}l}{\textbf{Prompt Search}}\\

Promptbreeder~\cite{fernando2024promptbreeder}
& Prompt search
& Prompt and mutation co-evolution
& Fixed task fitness \\

GEPA~\cite{agrawal2026gepa}
& Prompt search
& Trace reflection + Pareto selection
& Fixed task metric \\

MPO~\cite{choi2026mpo}
& Prompt search (multimodal)
& Evaluation-derived semantic feedback guides textual and non-textual prompt candidates
& Fixed task metric \\

C-Evolve~\cite{li2026cevolve}
& Prompt search
& Candidate generation inside a fixed evolutionary protocol
& Fixed protocol + task metric \\

\midrule

\multicolumn{4}{@{}l}{\textbf{Agent and Harness Search}}\\
Microsoft Foundry Agent Optimizer~\cite{microsoft2026agentoptimizer}
& Agent search
& Rewrites system instructions, skills, and local function-tool descriptions from evaluation failures
& Fixed evaluation process \\

ADAS~\cite{hu2025adas}
& Agent search
& Meta-agent coding + design archive
& Fixed benchmark \\

AFlow~\cite{zhang2025aflow}
& Workflow search
& MCTS over executable workflows
& Fixed evaluator \\

AgentSquare~\cite{shang2025agentsquare}
& Agent search
& Module evolution + recombination
& Fixed benchmark \\
\midrule

\multicolumn{4}{@{}l}{\textbf{Autonomous Experimental Search}}\\

AutoResearch~\cite{karpathy2026autoresearch}
& ML experiment
& Edit--train--evaluate loop
& Fixed metric \\

GPT-6 Astra NanoGPT~\cite{openai2026gpt6astra}
& Training-rule search
& Agent diagnoses bottlenecks, edits training code, tunes configuration and training loop
& Fixed validation target + compute \\

Auto Research~\cite{ning2026autoresearch}
& ML experiment
& Specialist agents + shared lineage
& External evaluator \\

AgentNAS~\cite{jeong2026agentnas}
& Architecture search
& LLM-generated task-specific seed architecture + derived structured search space
& Fixed validation metric \\

AutoKernel~\cite{jaber2026autokernel}
& Kernel experiment
& Profile--rewrite--benchmark loop
& Fixed correctness + speed \\

\bottomrule
\end{tabularx}
\end{table*}

\subsubsection{Evaluating L2 Strategy Autonomy}
Greater autonomy over improvement strategy does not by itself imply stronger recursive self-improvement. An L2 system may have a very expressive intervention space and nevertheless search it inefficiently, exploit weaknesses in its evaluator, or produce gains that disappear after several iterations. We therefore separate the \emph{degree of delegated autonomy} from the \emph{quality of the resulting improvement process}.

These dimensions are complementary: a broader search space is useful only if feedback can discriminate among candidates and the search process can explore that space efficiently. This combination helps explain why current industrial systems concentrate on coding, model training, and kernel optimization, where candidate interventions are executable and feedback can be obtained rapidly.

The same adaptive search loop also creates characteristic evaluation hazards. Repeated development-set access invites benchmark overfitting, additional search compute can be mistaken for algorithmic improvement, and an LLM evaluator may share the proposer's blind spots. Public automated-research systems make these risks concrete: reported behaviors include random-seed cherry-picking, shortcut discovery, and attempted test-label extraction through repeated evaluator queries~\cite{anthropic2026weakstrong}.
Once a benchmark is queried adaptively, it effectively becomes part of the optimization surface rather than a passive measurement instrument. 
Company-reported results remain useful evidence of feasibility and scale, but their provenance should be explicit and they should not be treated as equivalent to independently replicated experiments.

\begin{center}
\fbox{
\parbox{0.92\linewidth}{
\textbf{Finding: L2 shifts human effort from proposing individual improvements to constraining the search for improvements.}
The human still specifies the objective and acceptance criterion, but no longer needs to choose each candidate intervention. This changes the bottleneck from executing a known improvement to searching over possible improvements.
}}
\end{center}

\subsection{L3: Autonomy over Future Learning Experience}
\label{sec:l3-experience-autonomy}

L3 adds autonomy over \emph{what learning experience to acquire next}.
Whereas L2 concerns decisions about how to improve, the additional autonomy
examined at L3 concerns the learning agenda: the system uses evidence about
its current capabilities, failures, and learning history to determine what
it should learn from next.
Learning then changes the state on which subsequent experience acquisition
depends, connecting experience selection and persistent improvement across
rounds.

The characteristic loop at this level can be summarized as follows:

\begin{center}
\fcolorbox{black!25}{gray!4}{%
\begin{minipage}{0.92\linewidth}
\centering
\vspace{6pt}

\LoneStep{observe learner state}
\(\;\longrightarrow\;\)
\LoneStep{choose learning experience}
\(\;\longrightarrow\;\)
\LoneStep{acquire experience}

\par\vspace{7pt}

\(\;\longrightarrow\;\)
\LoneStep{update persistent state}
\(\;\longrightarrow\;\)
\LoneStep{reshape future experience}
\(\;\longrightarrow\;\)
\LoneStep{repeat}

\vspace{6pt}
\end{minipage}%
}
\end{center}

The defining property is \emph{learner-conditioned future experience
acquisition}.
Evidence about the evolving learner must inform a decision about which
experience to select, generate, or seek for subsequent learning, and the
resulting persistent update must feed back into later acquisition decisions.
This decision may be implemented by a separate curriculum component or
integrated into the agent's policy.
A policy update that incidentally changes visited states does not, by itself,
demonstrate autonomy over the learning agenda.

Experience need not be generated from scratch.
Repeatedly selecting material from an externally supplied pool can exhibit
this mechanism when selection adapts to the learner and participates in the
continuing loop; one-time filtering or retention alone does not.
Likewise, the objective, evaluator, update procedure, and rules governing
experience selection may be human-designed.
The relevant distinction is whether learner feedback autonomously changes
the next learning agenda, without requiring a human to redesign it after
each round.
Persistent learning may reside in model parameters or reusable external
state, including memories and skills.

\begin{figure*}[!t]
    \centering
    \includegraphics[width=1\linewidth]{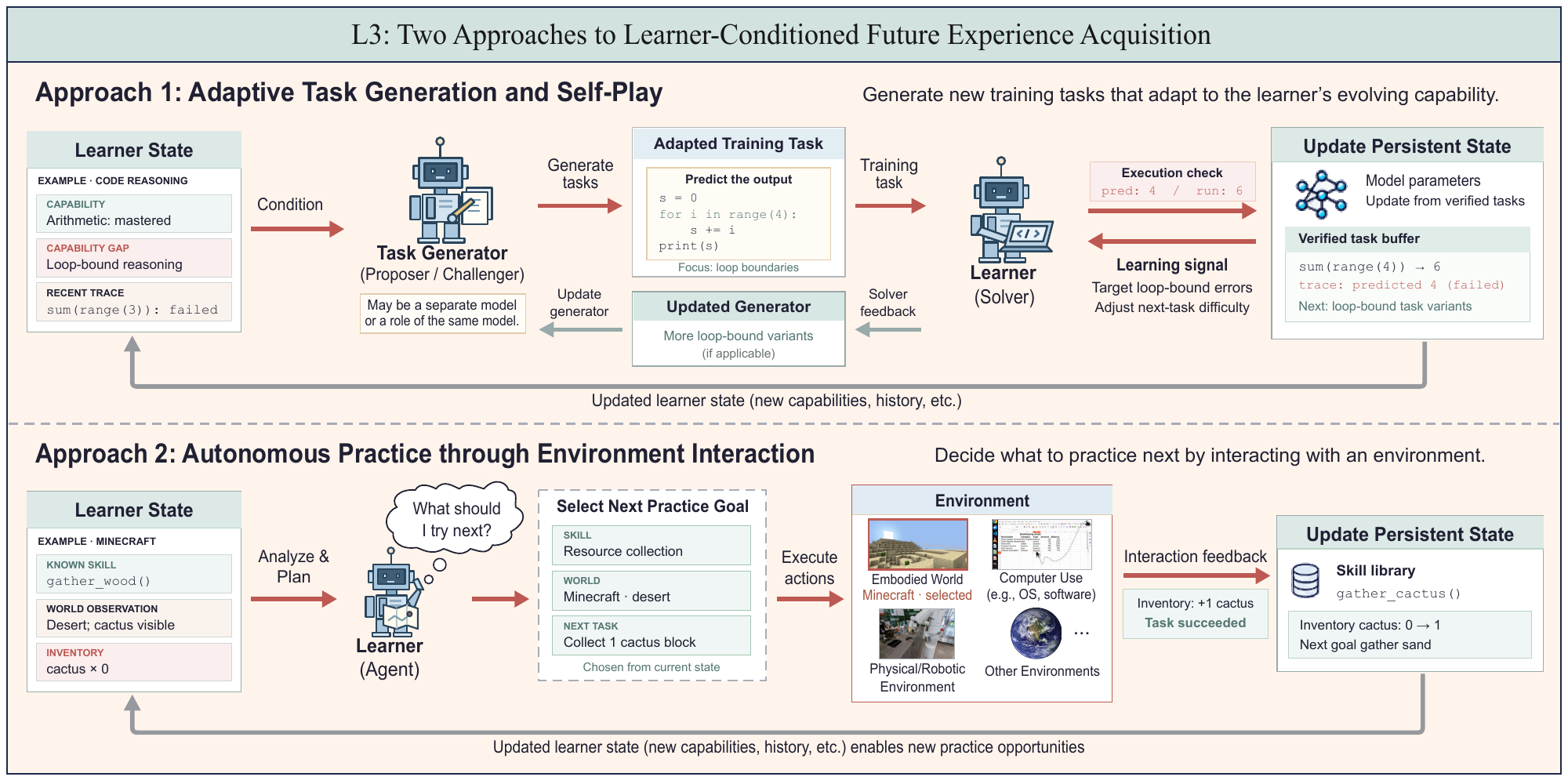}
\caption{Overview of L3: Learner-Conditioned Experience Acquisition.
Under human-defined objectives and evaluation constraints, the system
uses its current capabilities, failures, and interaction history to
determine what to learn from next.
\textbf{Top: Adaptive task generation and self-play.}
A task generator proposes training tasks targeted to the learner's
weaknesses, illustrated by code-reasoning tasks addressing loop-bound
errors. Execution-based feedback supports persistent learner updates
and, where applicable, refinement of the generator.
\textbf{Bottom: Autonomous practice through environment interaction.}
The agent selects a practice goal based on its current skills and
environmental observations, illustrated by collecting cactus in
Minecraft, and consolidates successful experience into a reusable
skill library.
In both approaches, persistent updates reshape subsequent task
generation or practice selection, closing the feedback loop between
learning and future experience acquisition.}    \label{fig:L3}
\end{figure*}

The examples below examine this experience-autonomy mechanism within
particular improvement loops.
Figure~\ref{fig:L3} illustrates two representative realizations of this
feedback structure: adaptive task generation and self-play, and
autonomous practice through environment interaction.
Evidence for this mechanism should be interpreted alongside the other
requirements of the hierarchy when assigning an overall system level.

\subsubsection{From Industrial Automation to Experience Autonomy}

\noindent\textbf{Modern training-data pipelines automate many data operations,
but execution automation alone does not establish experience autonomy.}
Drawing on the publicly documented practices reviewed here
~\cite{soldaini2024dolma,young2024yi,qwen2025qwen3,abdin2024phi4},
we organize recurring functions into six components: data-source preparation,
data labeling, data selection, data construction, data mixture and training
orchestration, and model-feedback diagnosis.
These functions need not form a fixed sequence; they can recur across
pretraining, supervised fine-tuning, preference optimization, and
reinforcement learning.

\begin{figure}[t]
    \centering
    \includegraphics[width=\linewidth]{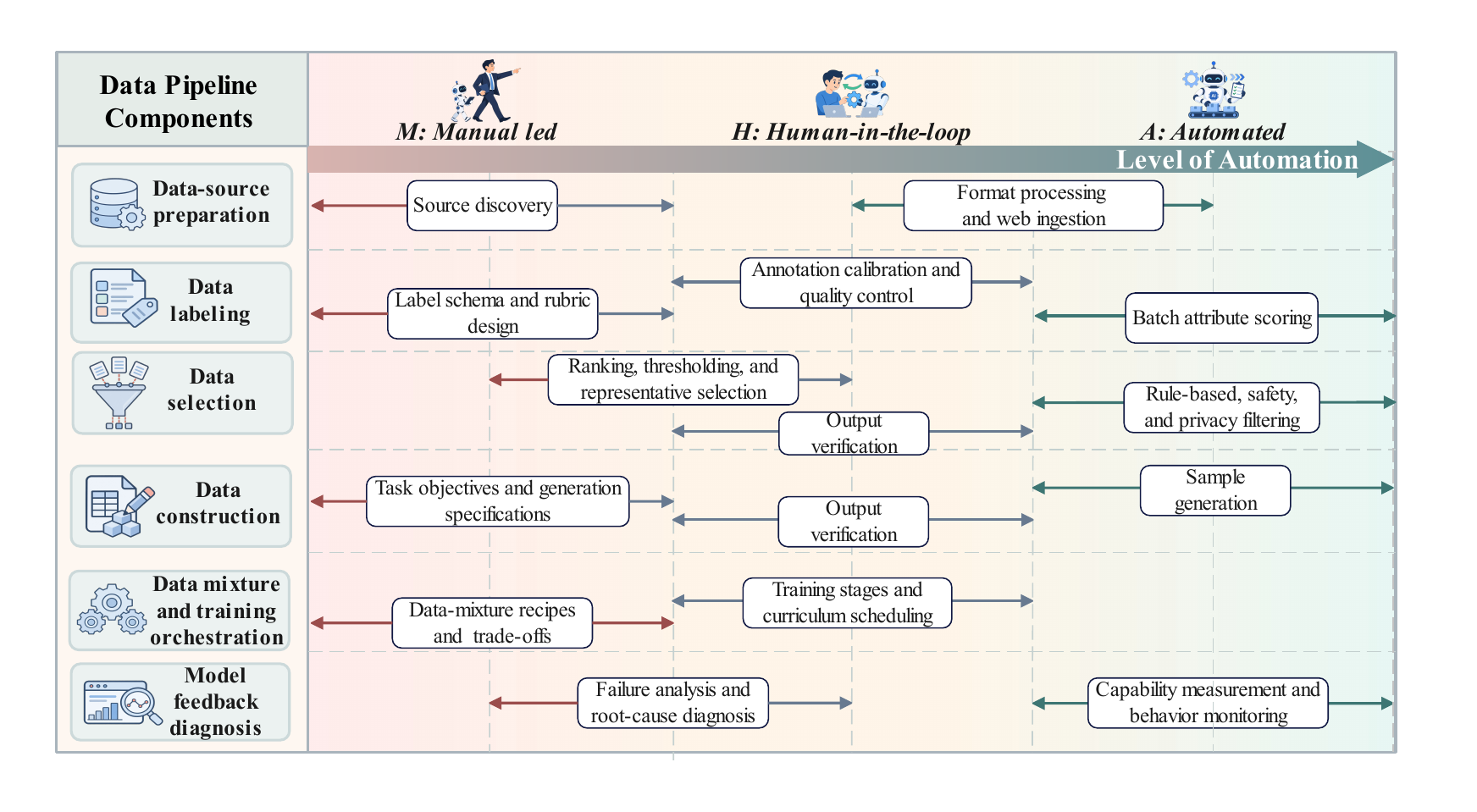}
    \caption{Schematic synthesis of automation in the training-data pipelines
    reviewed here. Rows denote six functional components, and rounded bars
    represent recurring operations. Horizontal position indicates a qualitative
    progression from manual-led (M), through human-in-the-loop (H), to automated
    (A). Positions and spans are an interpretive synthesis of reported practices,
    not quantitative measurements or per-organization scores. Operational
    automation is distinct from the additional requirement that learner
    feedback autonomously reshape experience acquisition across learning rounds.}
    \label{fig:industrial-data-automation}
\end{figure}

Fig.~\ref{fig:industrial-data-automation} summarizes this landscape
schematically.
Operations such as ingestion, filtering, and sample generation can be
executed at scale once their inputs and acceptance criteria are specified
~\cite{soldaini2024dolma,penedo2024fineweb,qwen2025qwen3}.
The reviewed reports also describe researcher-defined choices concerning
data policies, generation and mixture strategies, training stages, and
experimental evaluation
~\cite{young2024yi,penedo2024fineweb,li2024datacomplm}.
These choices illustrate the distinction between automating an operation
and delegating decisions about the subsequent learning agenda.

Human-designed rules do not preclude the L3-specific mechanism.
A fixed selection algorithm can still produce an adaptive learning agenda
if it uses the changing learner's feedback to choose subsequent experience.
Conversely, an automated pipeline that repeatedly generates and filters data
does not establish experience autonomy merely by producing new samples.
The decisive evidence is a closed dependence between learner state,
experience acquisition, persistent learning, and later acquisition decisions.
The pipeline descriptions cited here document substantial operational
automation, but do not by themselves establish this full dependence
~\cite{young2024yi,qwen2025qwen3,abdin2024phi4}.
This evidential distinction does not imply that industrial systems necessarily
lack such mechanisms.

The research examples below illustrate two recurring patterns:
\emph{adaptive task generation and self-play}, and
\emph{autonomous practice through environment interaction}.
They are complementary organizational views rather than exhaustive or
mutually exclusive categories: an interactive agent may also generate its
own tasks, and learner-conditioned selection can operate within either
pattern or over an existing data pool.

Table~\ref{tab:l3-experience-summary} compares representative loops by their
improvement target, learner-conditioned mechanism, and external constraints.

\begin{table*}[t]
\centering
\caption{Representative improvement loops exhibiting learner-conditioned
future experience acquisition. Constraints describe the scope of the
discussed loop, rather than independently establishing a system's maximum
autonomy level.}
\label{tab:l3-experience-summary}
\scriptsize
\setlength{\tabcolsep}{4pt}
\renewcommand{\arraystretch}{1.16}
\begin{tabularx}{\textwidth}{@{}
>{\raggedright\arraybackslash}p{0.12\textwidth}
>{\raggedright\arraybackslash}p{0.13\textwidth}
>{\raggedright\arraybackslash}p{0.39\textwidth}
>{\raggedright\arraybackslash}X
@{}}
\toprule
\textbf{Method}
& \textbf{Target}
& \textbf{Learner-conditioned mechanism}
& \textbf{Externally specified constraints}\\
\midrule
\multicolumn{4}{@{}l}{\textbf{Adaptive Task Generation and Self-Play}}\\
SSP~\cite{lu2026searchselfplay}
& Search agent
& Solver performance shapes proposer rewards and subsequent search tasks.
& Reward design and training procedure.\\
AZR~\cite{zhao2025absolutezero}
& Reasoning model
& Solver-dependent learnability rewards guide executable task proposal.
& Task formats, code executor, and update rules.\\
R-Zero~\cite{huang2026rzero}
& Reasoning model
& Solver answer consistency supplies a proxy that shapes Challenger tasks.
& Uncertainty reward, pseudo-label scheme, and optimization.\\
STP~\cite{dong2025stp}
& Theorem prover
& Conjectures near the current prover's frontier train the conjecturer.
& Formal domain, proof checking, and training rules.\\
PSV~\cite{wilf2026psv}
& Coding model
& Solver-derived difficulty labels condition specification generation.
& Formal verifier, difficulty scheme, and training recipe.\\
VisPlay~\cite{he2026visplay}
& Vision-language model
& Reasoner answer consistency and diversity rewards guide visual questions.
& Image source, reward design, and optimization.\\
\midrule
\multicolumn{4}{@{}l}{\textbf{Autonomous Practice through Environment Interaction}}\\
VOYAGER~\cite{wang2023voyager}
& Embodied agent
& Agent state and exploration history guide objectives; learned skills persist.
& Minecraft interface, curriculum prompts, and skill representation.\\
SIMA 2~\cite{deepmind2025sima2}
& Embodied agent
& In the full ASKA setup, evaluation feedback directs practice toward weaker skills.
& Task setter, reward rubric, and training procedure.\\
SEAgent~\cite{sun2026seagent}
& Computer-use agent
& Trajectory feedback updates a software guidebook that conditions later tasks.
& Software interfaces, assessment model, and learning procedure.\\
\bottomrule
\end{tabularx}
\end{table*}

\subsubsection{Adaptive Task Generation and Self-Play}

\noindent\textbf{The value of a training task changes as the learner improves.}
A fixed task distribution does not explicitly track this moving competence
frontier: familiar tasks may become uninformative, while tasks far beyond
current capabilities may provide little usable learning signal.
Adaptive generation addresses this problem by using learner-dependent
signals to estimate which tasks may support further improvement.
The central challenges are to obtain an informative signal, translate it
into a useful task distribution, and maintain the validity of the resulting
experience.

Search Self-Play (SSP)~\cite{lu2026searchselfplay} connects the first two
challenges by making proposer rewards depend on current solver performance.
As the solver changes, so does the incentive governing future search
problems, allowing the curriculum to evolve without manually redesigning
each task batch.
Absolute Zero Reasoner (AZR)~\cite{zhao2025absolutezero} couples proposal and
solution of executable reasoning tasks within a single model.
Its solver-dependent learnability reward guides task proposal, while a code
executor validates proposed tasks and checks solutions.
This separates two requirements that unconstrained synthesis can conflate:
experience must be evaluable, and its difficulty must be useful for the
current learner.

R-Zero~\cite{huang2026rzero} uses a Challenger--Solver architecture to adapt
task generation without externally supplied answer labels.
The Challenger's uncertainty reward depends on the consistency of multiple
responses from the current Solver; updates to the Solver therefore alter
the signal governing subsequent tasks.
This consistency-based quantity is a proxy for model-perceived difficulty,
not direct evidence of answer correctness or future learning gain.
Its role in the loop is to make experience acquisition responsive to the
learner even when stronger correctness signals are unavailable.

Formal domains provide an alternative route to reliable feedback.
STP~\cite{dong2025stp} addresses sparse proof rewards by jointly developing a
conjecturer and a prover.
Generated conjectures that the current prover can prove only with difficulty
provide training material for the conjecturer, while checked proofs improve
the prover.
The prover's changing frontier thus reshapes subsequent conjectures.
Proof assistants support correctness checking within the formal domain;
the learner-dependent selection criterion serves the separate purpose of
identifying useful training difficulty.

Propose, Solve, Verify (PSV)~\cite{wilf2026psv} brings a related mechanism to
verified code generation.
Solver-derived difficulty labels accompany examples used to prompt new
specifications.
Candidate specifications are checked for compilability, and generated
solutions are formally checked against their specifications before being
used for training.
These checks establish properties relative to the formal specification;
they do not guarantee that the specification expresses a useful task or
that its solution will improve the learner.
Indeed, PSV reports substantial overlap between the realized difficulties
of problems targeted as easy, medium, and hard, showing that
difficulty-conditioned proposal offers partial rather than exact
curriculum control.

VisPlay~\cite{he2026visplay} extends this approach to vision-language
reasoning using an image-conditioned Questioner and a Reasoner.
The Questioner receives an uncertainty reward that favors questions whose
majority-answer confidence is near the prescribed midpoint, together with
a diversity penalty that discourages repetitive questions.
As the Reasoner changes, its response consistency changes the reward for
future question generation.
As in R-Zero, the resulting signal is scalable but can reflect ambiguity
or inconsistent answers as well as productive difficulty.

Across these systems, feedback differs along two connected dimensions.
Executable outcomes and formal proof checking can ground correctness
within a specified task setting, whereas answer consistency estimates
difficulty without independently establishing correctness.
Adaptation may reside in a trained proposer or in generation prompts and
selection procedures conditioned on current learner measurements.
These choices trade off verification scope, feedback cost, and sensitivity
to imperfect proxies.
The shared advance is that estimated learning value influences future
experience acquisition, and persistent learning changes that estimate in
later rounds.

\subsubsection{Autonomous Practice through Environment Interaction}

\noindent\textbf{Interactive agents must decide where to spend their next
learning effort.}
Collecting additional trajectories alone does not resolve this problem:
unguided exploration may revisit mastered behaviors, while a fixed practice
schedule cannot explicitly respond to newly acquired skills or persistent
failures.
The systems below connect interaction feedback to future practice objectives,
using persistent learning to make subsequent experience more responsive to
the agent's changing capabilities.

VOYAGER~\cite{wang2023voyager} combines an automatic curriculum with a
persistent library of executable skills in Minecraft.
Current agent state and exploration history inform new objectives, while
successful behaviors become reusable skills for later tasks.
This couples curriculum decisions to accumulated capability: learning
changes what the agent can attempt, and further exploration supplies
experience from which additional skills can be acquired.
The relevant persistence is in external skills rather than a requirement
to update the underlying language model's parameters.

SIMA 2~\cite{deepmind2025sima2} illustrates why the experimental setting
matters when identifying this mechanism.
Its fixed-task experiment demonstrates improvement from self-generated
trajectories, but does not alone establish autonomy over task selection.
In the full ASKA self-improvement setup, a Gemini-based task setter instead
uses downstream reward-model evaluations to focus practice on weaker skills.
The resulting experience trains the agent, connecting capability assessment
to subsequent practice allocation.
The relevant autonomy belongs to this coupled learning system, including
its task setter, rather than requiring the acting model to make every
curriculum decision itself.

SEAgent~\cite{sun2026seagent} addresses practice in unfamiliar software
through an explicit connection between trajectory assessment and curriculum
memory.
A World State Model supplies trajectory judgments and descriptions of GUI
state changes to a Curriculum Generator, which updates a persistent
\emph{software guidebook} and uses it to generate later tasks.
The Actor learns from the collected experience, and subsequent interactions
further revise the guidebook and task set.
The guidebook therefore carries information from earlier exploration into
later practice decisions, helping the curriculum expand beyond previously
explored operations.

These systems share a feedback structure but depend on different persistent
resources and judgments.
VOYAGER links objectives to exploration history and reusable skills;
SIMA 2 connects reward-model assessments to task setting; SEAgent maintains
an evolving account of software behavior that guides future tasks.
Consequently, unreliable skill construction, inaccurate reward judgments,
or erroneous guidebook entries can each misdirect later learning effort.
Autonomous interaction reduces the need for manually prescribed curricula,
while leaving the quality of feedback and accumulated state central to the
reliability of the loop.

\subsubsection{Evaluating L3 Experience Autonomy}

Experience autonomy and learning quality are distinct.
A system may control its future learning agenda yet acquire experience
that is uninformative, narrow, or incorrectly evaluated.
Assessment should therefore establish both the feedback mechanism and the
benefit of using it: which learner signal influences acquisition, what
persistent state changes through learning, and how that change affects a
later acquisition decision.

To assess the contribution of learner conditioning, useful controls include
freezing the acquisition mechanism's learner-state input at an earlier
checkpoint, or replacing adaptive decisions with a learner-independent
schedule.
Comparisons should match access to data sources and environments and use
comparable compute and interaction budgets, including the cost of experience
generation and assessment.
Holding the acquired samples identical would remove the very distributional
adaptation under study.
These controls help separate the value of adaptive experience acquisition
from improvement attributable to additional training resources alone.

Evaluation should also distinguish correctness, difficulty, and learning
benefit.
Formal or executable checks can establish specified properties within their
supported setting, while consistency measures and model-based judgments
provide fallible estimates.
Neither verified correctness nor estimated difficulty alone establishes
marginal learning value.
Where feasible, studies should compare acquisition signals with independent
validity checks, realized task difficulty, and subsequent learning gains,
and examine whether these relationships remain stable as the learner changes.

We use \emph{experience corruption} to denote the risk that defective
experience or feedback distorts later learning and acquisition decisions.
A misleading difficulty proxy may favor unsuitable tasks; an inaccurate
judge may reward erroneous behavior; persistent memory may carry incorrect
assumptions into future practice.
Because these effects can enter parameters, skills, memories, or curriculum
state, they may propagate across rounds.
This is a structural risk of the feedback loop, rather than an assertion
that every system exhibits such amplification.

A credible evaluation should therefore track performance across multiple
rounds, the validity and diversity of acquired experience, and transfer to
fresh tasks or environments.
Feedback used to guide the curriculum must be distinguished from protected
evaluation used to assess generalization: repeated access to test instances
through task selection, reward construction, or persistent memory can make
them part of the learning process.
For task-generation systems, evaluation should examine whether the curriculum
continues to track the learner without collapsing onto narrow or easily
rewarded examples.
For interactive systems, it should examine whether practice continues to
acquire useful skills beyond familiar behaviors.

\begin{center}
\fbox{%
\parbox{0.90\linewidth}{%
\textbf{Finding: L3 adds autonomy over the future learning agenda.}
The system uses its evolving capabilities, failures, or learning history to
steer which experience is selected, generated, or sought next, and persistent
learning feeds back into later acquisition decisions.
Human-designed objectives, evaluators, and learning rules may remain in
place; experience autonomy concerns decisions made within this setting,
and does not itself guarantee useful or reliable improvement.
}}
\end{center}

\subsection{L4: Autonomy in Deployment and Environmental Adaptation}

\begin{figure*}[!t]
    \centering
    \includegraphics[width=1\linewidth]{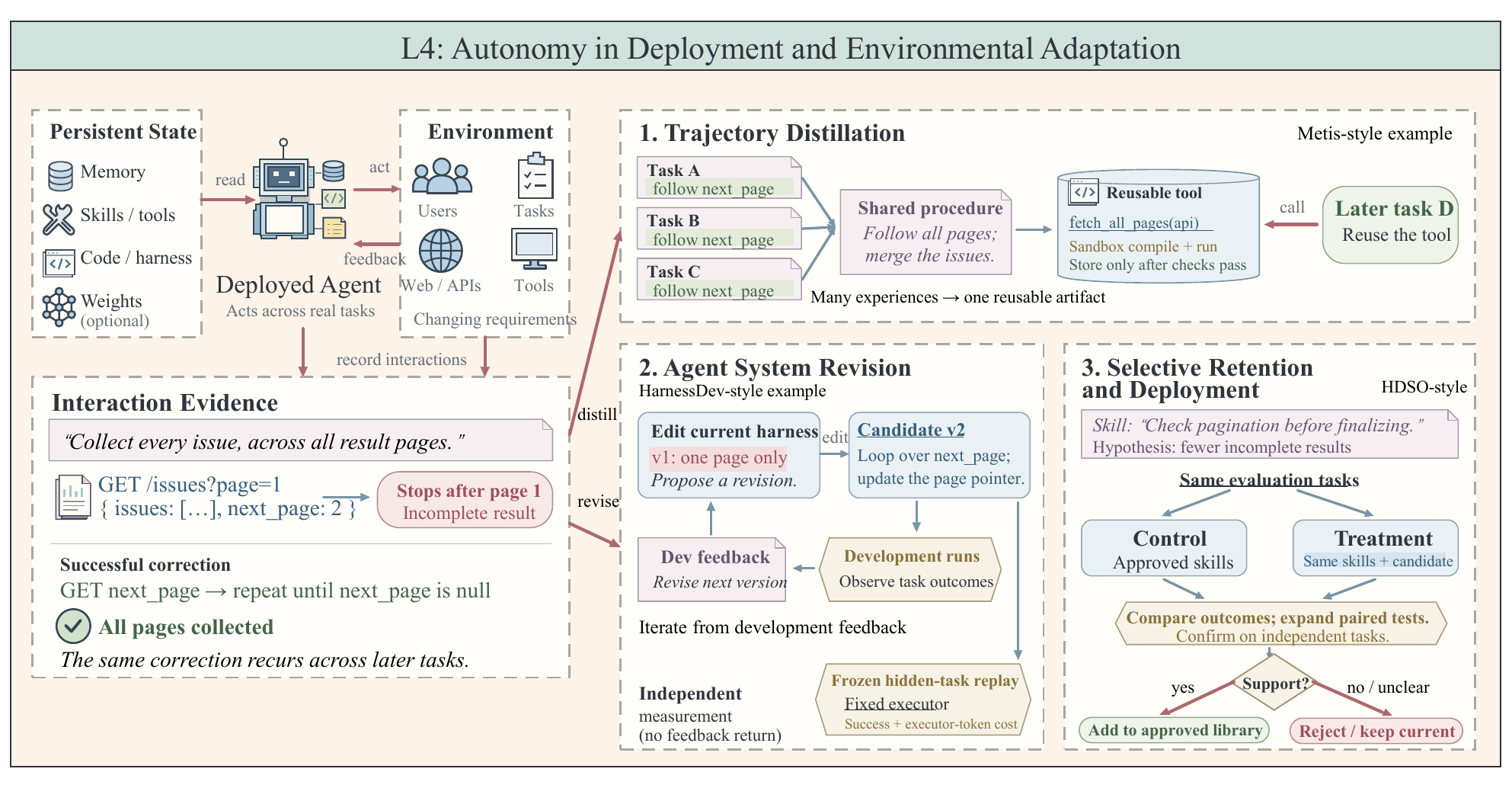}
    \caption{Overview of L4: Autonomy in Deployment and Environmental Adaptation.
Under human-defined objectives and deployment constraints, the AI system
uses interaction evidence from real tasks and changing environments to
determine which consequences of experience should persist and influence
future behavior. Experience can be distilled into reusable artifacts,
used to revise persistent components of the agent system, and selectively
retained based on subsequent evaluation. Accepted updates are reused in
later tasks, closing a persistent adaptation loop between deployment
experience and future behavior.}
    \label{fig:L4}
\end{figure*}

At L4, AI uses feedback from continued deployment interaction to decide
how the operating agent should persistently adapt.
Whereas L3 centers on what experience to acquire for learning, L4 centers
on how operational experience changes the memory, skills, or execution
components used in subsequent tasks.
Retained changes shape later behavior and the feedback available for
further adaptation.
The objective, access boundaries, protected evaluation, and authority
over consequential releases remain externally governed. 
Figure~\ref{fig:L4} provides an overview of this deployment-adaptation
mechanism, showing how interaction experience is converted into persistent
changes that shape subsequent behavior.

The characteristic loop at this level can be summarized as follows:

\begin{center}
\fcolorbox{black!25}{gray!4}{%
\begin{minipage}{0.92\linewidth}
\centering
\vspace{6pt}

\LoneStep{observe deployment interaction}
\(\;\longrightarrow\;\)
\LoneStep{propose persistent adaptation}
\(\;\longrightarrow\;\)
\LoneStep{revise agent components}

\vspace{7pt}

\(\;\longrightarrow\;\)
\LoneStep{validate and retain}
\(\;\longrightarrow\;\)
\LoneStep{reuse in later tasks}
\(\;\longrightarrow\;\)
\LoneStep{collect new feedback and repeat}

\vspace{6pt}
\end{minipage}%
}
\end{center}

We organize the discussion around trajectory distillation, iterative
revision of the agent system, and selective retention and deployment of
updates. These mechanisms differ in what they change, when the change is
reused, and how continued usefulness is established.
Table~\ref{tab:l4-adaptation-summary} summarizes representative methods
along these three axes.

\begin{table*}[t]
\centering
\caption{Representative systems, benchmarks, and analyses for L4 adaptation, grouped by the mechanisms discussed in this section.}
\label{tab:l4-adaptation-summary}
\scriptsize
\setlength{\tabcolsep}{3pt}
\renewcommand{\arraystretch}{1.12}
\begin{tabularx}{\textwidth}{@{}>{\raggedright\arraybackslash}p{0.15\textwidth}>{\raggedright\arraybackslash}p{0.23\textwidth}>{\raggedright\arraybackslash}p{0.14\textwidth}>{\raggedright\arraybackslash}X@{}}
\toprule
\textbf{Method} & \textbf{Update target} & \textbf{Timing} & \textbf{Update validation} \\
\midrule
\multicolumn{4}{@{}l}{\textbf{Trajectory distillation}} \\
Dynamic Cheatsheet~\cite{suzgun2026dynamic}
& Evolving text memory
& After each problem
& Task outcomes guide self-curation; no held-out gate. \\
ACE~\cite{zhang2026agentic}
& Incremental context entries
& After each task
& Helpfulness and harmfulness counters guide edits. \\
ReasoningBank~\cite{ouyang2026reasoningbank}
& Reasoning strategies (text)
& After each task
& A judge labels trajectories before strategy extraction. \\
APEX~\cite{li2026apex}
& Milestone dependency graph
& After each episode
& Episode outcomes are propagated through the graph. \\
PersonaAgent~\cite{zhang2026personaagent}
& Per-user persona prompt
& Across user interactions
& No separate admission test is reported. \\
PAHF~\cite{liang2026learning}
& User preference entries
& When requests are ambiguous
& User clarification confirms and revises preferences. \\
MemToolAgent~\cite{er2026memtoolagent}
& Tool-use critiques and retrieval policy
& After tool calls
& Environment and user feedback assess tool use. \\
Trace2Skill~\cite{ni2026trace2skill}
& Portable skill document
& Offline batch
& Error and success analysts propose and merge patches. \\
PRACTICE~\cite{bai2026practice}
& Embodied skill library
& Offline batches
& Success/failure contrasts and teacher distillation. \\
PANDO~\cite{li2026pando}
& Rules and routines
& Online, in-run
& Confidence-scored admission and demotion. \\
PILOT~\cite{xiao2026pilot}
& Procedures and failure modes
& During long-horizon runs
& Supervisor distillation uses live-run feedback. \\
Metis~\cite{dai2026metis}
& Text plans and code tools
& After completed tasks (asynchronous)
& Repeated reuse plus sandbox compile checks. \\
Evo-Harness~\cite{wei2026evo}
& Harness skill tuples
& After task batches
& Failed or negatively evaluated tasks trigger reflection and edits. \\
SHAPER~\cite{wang2026shaper}
& Planning guidance and context-selection code
& Across task rounds
& Sandbox execution and downstream task outcomes. \\
Evo-Memory~\cite{wei2025evo}
& Accumulated agent memory (benchmark)
& Sequential task streams
& Long-horizon evaluation of memory reuse. \\
\midrule
\multicolumn{4}{@{}l}{\textbf{Iterative revision of the agent system}} \\
DecoEvo~\cite{chen2026decoevo}
& Solver and rubric skills
& Iterative rounds
& Score-independent audits, Pareto-checked. \\
HarnessDev~\cite{wu2026harnessdev}
& Runnable agent harness
& Iterative rounds
& Frozen replay of candidates on hidden tasks. \\
ASPIRE~\cite{wu2026aspire}
& Model weights or harness
& Iterative rounds
& Rollback unless the verified score improves. \\
S3Gym~\cite{shi2026s3gym}
& History, summaries, or model parameters
& Iterative rounds
& Compares update pathways without verifier outcomes. \\
Harness Benefit~\cite{lin2026harness}
& Harness updates and their use (analysis)
& Fixed solve-evolve rounds
& Separates update quality from execution benefit. \\
\midrule
\multicolumn{4}{@{}l}{\textbf{Selective retention and deployment of updates}} \\
HDSO~\cite{shang2026hypothesis}
& Candidate skill packages
& Periodic rounds
& Paired control and treatment executions. \\
Library Drift~\cite{zhang2026library}
& Skill library entries
& Each round
& Low-contribution skills are retired. The library is capped. \\
Rethinking Skill Evolution~\cite{liu2026rethinking}
& Skill edits (analysis)
& Multiple rounds
& Filters edits using task outcomes across rounds. \\
Tax AI~\cite{taxai}
& Product code and evals
& Release cycles
& Targeted and regression tests, then human review. \\
\bottomrule
\end{tabularx}
\end{table*}

\subsubsection{Trajectory distillation}

Trajectory distillation converts an agent's interaction histories into compact artifacts that persist across sessions and condition later tasks. \wyk{Compared with raw trajectories, distilled artifacts are more concise, less costly to consult, and structured for easier reuse in subsequent tasks.} We group the methods below by the form of the distilled artifact: textual experience memory, structured memory, procedural skill libraries, and executable code.

\noindent\textbf{Textual experience memory.} This category keeps the distilled artifact as natural-language passages that are retrieved into the prompt for later tasks. Dynamic Cheatsheet~\cite{suzgun2026dynamic} enables a frozen model to improve across a series of related problems without access to ground-truth labels. It maintains a single evolving note of strategies, code snippets, and known pitfalls, which the same model consults when answering each incoming problem and curates by adding distilled lessons and removing superseded entries. Agentic Context Engineering~\cite{zhang2026agentic} follows the same recipe while keeping the memory detailed and informative as it grows. It stores short entries with counters that record whether each entry helps or hinders later tasks, and adds small patches rather than rewriting the whole note, which avoids information loss by compression. ReasoningBank~\cite{ouyang2026reasoningbank} adapts the idea to multi-step agents and learns from failures as much as from successes by distilling each judged trajectory into a short, titled strategy that is retrieved before later tasks.

\noindent\textbf{Structured memory.} Instead of free text, structured memory stores experience in explicit records or graphs, making the conditions for reuse inspectable. APEX~\cite{li2026apex} encourages an agent to explore new strategies as its memory grows rather than letting it settle into familiar routines by building a strategy map as a graph of task milestones linked by prerequisite relations. After each episode, the outcome is propagated back along the milestones that led to it, so the map records which paths have worked. A separate step adds promising branches that the agent has not tried, and the agent consults the map at planning time to choose between a known good path and an untried one. Other structures mirror the environment. For instance, PersonaAgent~\cite{zhang2026personaagent} and PAHF~\cite{liang2026learning} instead organize memory around individual users. PersonaAgent translates episodic and semantic user memory into a per-user persona prompt, while PAHF revises preference entries by asking users to clarify an ambiguous request when no relevant preference is found, then integrates their feedback into memory and revises outdated entries. MemToolAgent~\cite{er2026memtoolagent} gives memory a more procedural role by storing critiques of failed tool calls and adjusting how many entries it retrieves per call.

\noindent\textbf{Procedural skill libraries.} Instead of only describing experience, this category extracts reusable skills or standard operating procedures that a later agent follows directly. Trace2Skill~\cite{ni2026trace2skill} turns lessons from individual trajectories into skills that other models and tasks can reuse. It first rolls out a frozen agent on domain tasks. Two analyst roles, one focused on errors and the other on successes, then read the trajectories and propose skill edits, which are merged in stages into a portable skill directory. Because the consolidated document is read directly rather than retrieved for each episode, later agents load it once as part of their instructions. PRACTICE~\cite{bai2026practice} applies the same offline distillation to embodied agents while keeping the skill library consistent as it changes rather than growing unbounded, where a trainable skill learner proposes batch edits that add, refine, merge, or remove entries. The learner first learns basic skill generation and library maintenance from oracle trajectories, then learns failure awareness by contrasting successful and failed trajectories from different executors on the same tasks, and is finally distilled toward a stronger teacher on its own edit distribution. Two other methods acquire skills during execution. PANDO~\cite{li2026pando} improves a web agent during deployment by distilling rules that prevent repeated failures and parameterized routines that replace multi-step browser subgoals from each rollout during deployment, while PILOT~\cite{xiao2026pilot} uses a supervisor to distill procedures and failure modes from a live, long-horizon run, so later sessions start with the accumulated skills.

\noindent\textbf{Executable artifacts.} Moving from text to code, this category compiles experience into artifacts that are invoked directly, combining the flexibility of text memory with the efficiency of executable tools. Metis~\cite{dai2026metis} achieves this through a dual memory, a text store of plans, environment facts, and pitfalls alongside a code library of callable tools. A reflector maintains the text entries, and a frequently reused text plan is rewritten as a callable tool and accepted only after it compiles and runs correctly in a sandbox. Retrieval then covers both text entries and tool descriptions. Evo-Harness~\cite{wei2026evo} broadens the target from single tools to the whole harness, compiling lessons from failed or negatively evaluated tasks into cross-task patterns and task-specific procedures. SHAPER~\cite{wang2026shaper} applies the same idea to embodied agents by evolving textual planning guidance together with a sandboxed Python function that selects the context shown to the planner.

\noindent\textbf{Evaluation protocols.} Evaluation protocols for this family differ in time horizon and in what they count as benefit. Evo-Memory~\cite{wei2025evo} restructures datasets into sequential task streams so that memory accumulation and reuse can be measured over long horizons. PANDO~\cite{li2026pando} instead audits behavior within a run, reporting action repetition, step overhead, and prompt-cache utilization, and Metis~\cite{dai2026metis} relates task quality to execution and construction cost.

\subsubsection{Iterative revision of the agent system}

Rather than what an agent remembers, this section revises the agent system itself: the skills that produce solutions, the rubrics that judge them, the harness that runs them, or the weights underneath. The defining question is which components may change and which acceptance criteria stay fixed while they do. We group the work into co-evolving coupled components, benchmarks that measure self-directed system revision, and evolutionary search under fixed verifiers, followed by analyses of whether revision translates into benefit.

\noindent\textbf{Co-evolving components.} Co-evolution revises two coupled components at once, which raises a circularity problem: if the judge improves alongside the solver, higher scores may reflect an easier judge rather than better solutions. DecoEvo~\cite{chen2026decoevo} addresses this problem by evolving a solver skill and a rubric-generator skill in text space while decoupling their objectives and withholding gold rubrics during optimization. The solver skill is updated from criterion-level rubric feedback. The rubric generator, in contrast, is gated by two score-independent audits: a structural audit checks that the generated rubric covers the task's requirements, and a contrastive audit checks that it discriminates between near-tie responses, with Pareto verification across the two. Because the generator never sees the solver's aggregate score, it cannot improve its standing by making the rubric easier.

\noindent\textbf{Benchmarks of self-directed revision.} Rather than proposing new methods, this line of work measures whether current models can revise their own systems at all. HarnessDev~\cite{wu2026harnessdev} asks whether models can create and improve their own harness, the model-external scaffolding of prompts, tool loops, and execution logic. The creator agent builds a complete harness from a minimal seed and a few development cases and iteratively revises it from downstream feedback. The candidate revisions are frozen and scored on hidden tasks for both success rate and executor-token cost, with the creator and the executor kept separate so that harnesses can be compared under one fixed executor. ASPIRE~\cite{wu2026aspire} asks whether a model can improve itself given only a vague goal. It delegates the choice of data, update method, and validation signal to the agent across both weights and the harness, and a score-gated controller rolls back any change whose verified score does not improve. S3Gym~\cite{shi2026s3gym} restricts the question to the experience channel, withholding verifier outcomes so the agent must judge its own trajectories, and compares raw history, compressed summaries, and parameter training as improvement pathways.


\noindent\textbf{Analyses of revision benefit.} Whether revision actually translates into benefit is a question in its own right, and one analysis finds that the two are often conflated. The study behind Harness Updating Is Not Harness Benefit~\cite{lin2026harness} separates two capabilities: producing useful persistent updates, and exploiting them at solve time. Each is measured under a fixed solve-evolve protocol with identical prompts and budgets across several model backbones and three agent benchmarks. Updating ability is largely independent of base model capability, with a small model's updates yielding gains comparable to a frontier model's. The ability to benefit from an updated harness, by contrast, varies non-monotonically, and failures concentrate in two modes: relevant artifacts are not activated, or they are activated but not faithfully followed. A practical consequence is that capability investment belongs in the task-solving agent rather than in the evolver.

\subsubsection{Selective retention and deployment of updates}

Trajectory distillation and system revision both end with a proposed change, while this section focuses on which proposed changes become persistent and which stay available to later tasks. The work below applies this control at three points: before admission, during continued use, and at release into an operational system.

\noindent\textbf{Candidate validation.} A proposed update must produce evidence before it can enter the persistent state. HDSO~\cite{shang2026hypothesis} is designed to prevent skills distilled from noisy trajectories from encoding spurious shortcuts or rules that the executor cannot follow. A curator model observes compact executor traces and proposes a hypothesis with an explicit validation plan, where each candidate is tested by executing the same tasks twice, once with the current approved repository as the control and once with the candidate skill added as the treatment, and the candidate enters the approved repository only when the differences between the two runs support the hypothesis. The comparison runs in stages of increasing size and ends with confirmation on independent tasks. Metis~\cite{dai2026metis}, introduced earlier in the trajectory-distillation section, applies a similar gate after the text plan is already retained in memory, where a text plan becomes executable code only after it recurs across tasks and passes dependency and compilation checks.

\noindent\textbf{Library maintenance.} After admission, attention shifts to the health of the library itself, since skills that were once useful can degrade as tasks and models change. The Library Drift study~\cite{zhang2026library} shows that unbounded accumulation gradually degrades retrieval quality and stalls progress, often before the effect becomes visible in task scores. It proposes lifecycle management with three parts. An append-only evidence log tracks each skill's contribution to task outcomes, and a skill is retired when its measured contribution fades. The number of active skills is capped, so a new skill competes with the entries already present. New skills are also written under a meta-skill, a stored guide that steers how future skills are authored. The study shows that governance choices matter: retiring skills too aggressively performs worse than leaving the library unguided. Meanwhile, another analysis~\cite{liu2026rethinking} finds that carefully filtering skill edits, rather than repeatedly rewriting skills from task outcomes, improves task performance.

\noindent\textbf{Governed deployment.} In production, an incorrect update can affect real users, so a strict gate is needed to prevent unexpected issues. The Tax AI deployment~\cite{taxai} applies such a gate inside a production tax-preparation system, where practitioner corrections drive improvement. Corrections are recorded as structured field-level evidence, and repeated failures are grouped into evaluation targets. A coding agent investigates the trace, evaluations, and repository to implement and validate a fix against targeted and regression evaluations. The loop may change only a bounded part of the product, such as the extraction schema, source selection, tax-engine mapper, and graders. Engineers retain decisions about architecture and product design. Each fix arrives as a pull request for engineering review before release, and cases that the agent cannot resolve are routed back to practitioners. Humans therefore make the final acceptance decision, and the delay introduced by this review is part of the loop's operating cost.

\subsubsection{Evaluating L4 Deployment Autonomy}

Greater autonomy over adaptation after deployment does not by itself establish reliable improvement. An L4 system may retain updates without showing that they remain useful as tasks, environments, or executors change. Evaluation should distinguish the degree of deployment autonomy from the quality of the resulting adaptation process.

A characteristic risk at this level is persistent update failure, where an accepted change may affect many later decisions before its weakness becomes visible. Across different forms of retained state, failure may arise because the system infers the wrong lesson from noisy evidence, applies a sound lesson outside its valid scope, fails to invoke a relevant update, or preserves new behavior at the expense of earlier capabilities~\cite{suzgun2026dynamic,zhang2026library,liu2026rethinking,lin2026harness}. Final task scores alone cannot separate these mechanisms. A credible evaluation must connect each retained change to the evidence that produced it, its later use, and its effects on both new and previously solved tasks.

The current evidence leaves further uncertainty about whether deployment adaptation remains effective over time. Most evaluations cover bounded task streams and limited combinations of models, tasks, and environments, while repeated assessment makes longer studies costly~\cite{ni2026trace2skill,xiao2026pilot,wu2026harnessdev}. Feedback may also be incomplete or endogenous to the loop, as outcome labels can be noisy, evaluators can change with the system, and a useful retained artifact may still fail to influence execution~\cite{ouyang2026reasoningbank,chen2026decoevo,lin2026harness}. Evaluation of L4 should consequently examine adaptation across time and distribution shifts, including the point at which a retained change alters later behavior.

L4 autonomy remains bounded by the surrounding deployment process. The agent may decide how interaction evidence changes persistent state, while humans continue to specify the objective, access boundaries, protected evaluation, and release authority. L4 should concern autonomy over operational adaptation within a deployed system.

\begin{center}
\fbox{
\parbox{0.92\linewidth}{
\textbf{Finding: L4 shifts autonomy from selecting experience to deciding which consequences of experience persist in deployment.}
The agent converts interaction histories into retained changes to memory, skills, harnesses, code, or model parameters, and these changes affect later tasks and feedback. Humans retain the objective, protected acceptance criteria, access boundaries, and final authority over consequential releases.
}}
\end{center}

\subsection{L5: From Environmental Adaptation to Meta-Improvement}
\label{sec:l5-meta-improvement}

An improvement process can become a bottleneck in its own right. A coding agent may repeatedly generate similar unsuccessful patches because its search procedure discards useful alternatives. A research agent may optimize a development score after that score has stopped predicting external performance. Repairing such failures requires changes to the procedures that direct experiments and judge their outcomes. Designing these procedures is costly: a revision must be assessed through the later improvements it produces, often across several tasks and repeated runs~\cite{zelikman2024stop,shi2026aevolvetraining}.

\textbf{L5 begins when AI persistently modifies a mechanism responsible for future improvements and uses the revised mechanism in subsequent rounds.} The editable mechanism may be an improver, a successor evaluator, a search policy, or a procedure for directing research. L2 searches for candidate interventions; L3 determines what experience to acquire; L4 incorporates deployment feedback into persistent adaptation. L5 additionally makes the procedure governing later improvement an object of improvement. A deployed agent that retains new debugging skills through an unchanged learning procedure exemplifies L4. Revising and reusing the procedure that diagnoses failures and builds those skills can cross the L5 boundary.

\begin{figure*}[!t]
  \centering
  \includegraphics[width=\linewidth]{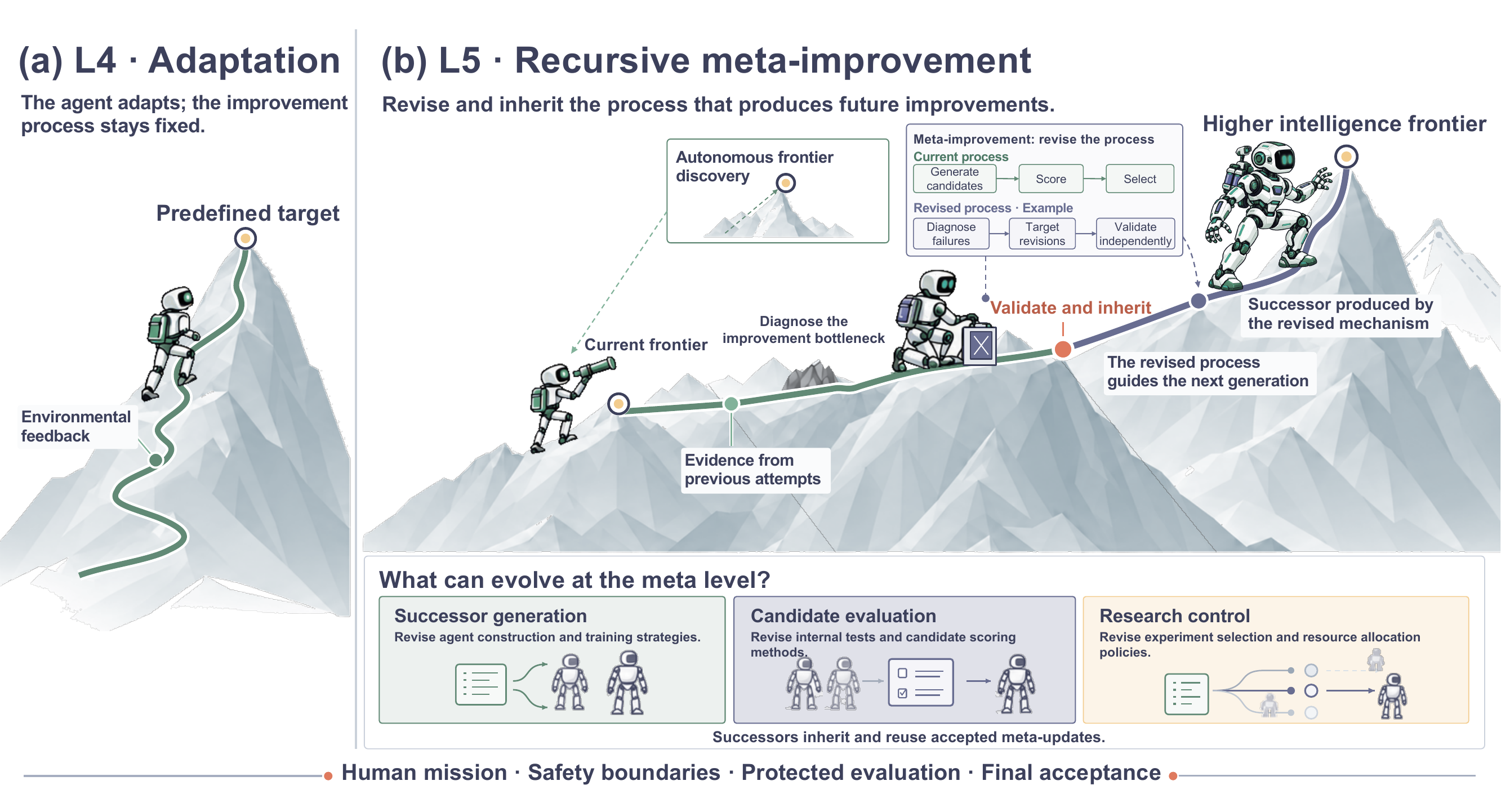}
  \caption{From environmental adaptation to recursive meta-improvement. (a) L4 incorporates environmental feedback within a fixed improvement process. (b) L5 diagnoses limitations of that process, revises it, and passes validated changes to successors. Meta-updates can affect successor generation, candidate evaluation, or research control. Human-defined missions, safety boundaries, protected evaluation, and final acceptance remain external constraints. The ascending path illustrates possible progress through inherited meta-updates.}
  \label{fig:l5-meta-improvement}
\end{figure*}

The loop closes when a revised mechanism returns to govern the generation, evaluation, or selection of later successors. The inherited state includes the accepted code, prompts, evaluators, or research policy, together with evidence used by later revisions. Human designers still establish the overall mission, protected evaluation, editable components, and resource permissions, and may retain veto and deployment authority. Some of these decisions are enforced by fixed infrastructure; they do not require a person to approve every iteration. Figure~\ref{fig:l5-meta-improvement} illustrates how evidence from earlier attempts can guide a revision of the improvement process, which successors then inherit and reuse.

These conditions distinguish \emph{structural L5}, which demonstrates that an AI-directed change persists and controls a later improvement round, from \emph{effective L5}, which demonstrates that the revised mechanism produces or selects better successors under comparable budgets and independent assessment. Self-modifying task code provides insufficient evidence when the process responsible for later revisions remains unchanged. Conversely, an evolved successor evaluator can establish structural L5 even with unchanged task-agent code. Classification therefore applies to the particular mechanism examined. Table~\ref{tab:l5-recursive-ascent} summarizes the loop, inherited state, and external controls of representative systems.

\begin{table*}[t]
  \centering
  \caption{L5 mechanisms and external controls.}
  \label{tab:l5-recursive-ascent}
  \scriptsize
  \setlength{\tabcolsep}{3pt}
  \renewcommand{\arraystretch}{1.12}
  \renewcommand{\tabularxcolumn}[1]{m{#1}}
  \begin{tabularx}{\textwidth}{@{}>{\raggedright\arraybackslash}m{0.19\textwidth}>{\raggedright\arraybackslash}m{0.23\textwidth}>{\raggedright\arraybackslash}m{0.24\textwidth}>{\raggedright\arraybackslash}X@{}}
    \toprule
    \textbf{Work} & \textbf{Loop closes at} & \textbf{Inherited state} & \textbf{External controls} \\
    \midrule
    \multicolumn{4}{@{}l}{\textbf{Search and self-revision}} \\
    STOP~\cite{zelikman2024stop} & Next program search & Improver code & Utility; base LM; budget \\
    \addlinespace[2pt]
    G\"odel Agent~\cite{yin2025godelagent} & Next self-revision & Task and update code & Task objective; runtime access \\
    \addlinespace[2pt]
    DGM~\cite{zhang2026dgm} & Descendant search & Agent code; archive & Parent selection; benchmark \\
    \addlinespace[2pt]
    HyperAgents~\cite{zhang2026hyperagents} & Next agent generation & Task and meta-agent code & Main-study selection; evaluation \\
    \midrule
    \multicolumn{4}{@{}l}{\textbf{Successor evaluation}} \\
    RQGM~\cite{iacob2026redqueen} & Next-epoch selection & Evaluator; agent code & Anchor; replacement schedule \\
    \midrule
    \multicolumn{4}{@{}l}{\textbf{Research policies and harnesses}} \\
    A-Evolve-Training~\cite{shi2026aevolvetraining} & Next research round & Search policy; discovery log & Constitution; benchmark; substrate \\
    \addlinespace[2pt]
    AIRA$_2$~\cite{hambardzumyan2026aira2} / AAR~\cite{chen2026automatedresearchers} & Task experiments & Research artifacts & Research harness; evaluation \\
    \addlinespace[2pt]
    AIDE$^{2}$~\cite{weco2026aide2} & Later research runs & Research-agent harness & Private scores; cost budget \\
    \bottomrule
  \end{tabularx}
  \par\vspace{3pt}{\scriptsize\raggedright\textit{Note:} L5 claims are qualified in the text.\par}
\end{table*}

\subsubsection{Improving the Search Procedure}
\noindent\textbf{The AI revises how future candidate improvements are generated and explored.} A fixed optimizer can spend its budget on unproductive proposals, even when its underlying model can generate better search algorithms. The challenge is to evaluate a proposed optimizer through its ability to improve other programs.

STOP~\cite{zelikman2024stop} makes this problem executable. Its improver is a Python program that calls a fixed language model, generates candidate code, evaluates a supplied utility, and returns a selected candidate. The current improver receives its own source as the optimization target. Successor improvers are scored by the quality of programs they produce on downstream tasks, and the selected improver performs the next self-improvement round. The inherited artifact is thus a search procedure that may introduce beam search, alternative sampling, or other candidate-selection logic. The study reports that a selected fourth-generation improver outperformed the seed on all five transfer tasks excluded from self-improvement. The task distribution, utility, base model, and resource limits remained externally specified. Weaker-model runs regressed on average, and some generated programs evaded soft budgets or exploited evaluation bugs.

G\"odel Agent~\cite{yin2025godelagent} gives the AI access to both its task policy and recursive update logic in a shared Python program. Execution feedback informs a rewrite, and the revised program runs in the next recursive call, including its modified self-update procedure. This broadens the search beyond a developer's fixed editing routine. It also makes errors inheritable: the study reports that 14 of 100 MGSM optimization trials ended below the initial policy's performance, and unrestricted runs could call stronger models. Task objectives and execution permissions therefore remain consequential external controls.

The Darwin G\"odel Machine (DGM)~\cite{zhang2026dgm} retains coding-agent variants and explores their descendants, but keeps archive management and parent selection outside self-modification. Its lineage provides a transition case; task gains alone do not establish a better improvement procedure. HyperAgents~\cite{zhang2026hyperagents} directly exposes both task-agent and meta-agent code to revision. Evolved meta-agents build performance trackers and persistent memory to guide later modifications. In its preprint, meta-agents evolved on paper review and robotics generated improved agents for previously unseen mathematics grading, supporting transfer of the improvement procedure itself. The longer, 200-iteration experiment did not establish a statistically significant final advantage for transferred initialization. Parent selection and evaluation also remained fixed in the main experiments.

\subsubsection{Improving Successor Evaluation}
\noindent\textbf{The AI improves the evaluator that determines which successors are retained.} Repeated optimization places pressure on a judge's blind spots. Keeping that judge unchanged can favor proposals that exploit its errors. Unrestricted judge revision creates the opposite problem: successive scores may reflect changing standards.

The Red Queen G\"odel Machine (RQGM)~\cite{iacob2026redqueen} addresses this tension through evaluator co-evolution. Its learned evaluator remains frozen within an epoch. At a scheduled boundary, challenger evaluators are compared against an independent ground-truth anchor; the selected evaluator then governs the next epoch. Scores dependent on the replaced evaluator are discarded and affected agents are re-evaluated when revisited. The loop carries forward an improved evaluator and agent code while resetting incompatible evaluation records. On held-out Polyglot coding tasks, the preprint reports a 71.7\% pass rate against 69.9\% for HGM-H, with lower search-token use. The anchor, replacement schedule, and orchestration remain externally fixed. These controls preserve a reference against which evaluator changes can be judged; the formal stability argument applies within each frozen epoch, not to unrestricted changes of objective.

\subsubsection{Revising Research Goals and Policies}
\noindent\textbf{The AI uses accumulated experimental evidence to revise what later research should pursue.} Choosing a higher target is difficult when the available proxy rewards progress on the wrong bottleneck. In model post-training, testing that hypothesis requires comparing training runs, identifying failed search directions, and deciding whether to revise the research policy itself. A changing curriculum reaches L5 when its persistent selection procedure is also revised and reused.

A-Evolve-Training~\cite{shi2026aevolvetraining} provides a concrete case. Its preprint reports autonomous post-training of a 30B model across four rounds. Workers return recipe changes, checkpoint evaluations, and failures; a collector consolidates these results, and a meta-agent revises the next round's search policy. The policy carries a standing recipe, promoted or retired search directions, and a registry of unsuccessful approaches. When development scores increased without corresponding external gains, the revised policy directed workers toward interventions that could improve the external target despite lowering the proxy. The study reports a final leaderboard score of 0.86, compared with 0.87 for the top human submission. Cross-round inheritance resides in the research policy and discovery log; the underlying worker substrate and constitution remain fixed. The case supports policy-level L5 within a human-defined objective, with no demonstrated autonomous revision of that objective.

A spiral of improvement emerges when an inherited procedure helps expose a new limitation and is revised again to address it. The next goal should identify a testable capability gap, preserve established capabilities, and justify further expenditure. Goal revision alone supplies limited evidence: the evaluation must establish that the revised policy changes subsequent research and improves its outcomes.

\subsubsection{Industrial Practice}
Industrial implementations expose different portions of the improvement process to revision. Their relevance to L5 depends on whether the retained artifact changes later research decisions.

\emph{Research infrastructure.} Meta's AIRA$_2$~\cite{hambardzumyan2026aira2} uses asynchronous experimentation and hidden consistent evaluation; Anthropic's Automated Alignment Researchers~\cite{chen2026automatedresearchers} generate and test mitigations for specified alignment failures. These systems automate experiment execution within researcher-designed workflows. Their reported results concern research outputs, without establishing inherited changes to the procedures conducting the research.

\emph{Research-agent revision.} Weco's AIDE$^{2}$~\cite{weco2026aide2} applies a research agent to improving a research-agent harness. Candidate harnesses run downstream tasks under private scoring and a fixed cost budget. Accepted versions are retained for subsequent research; a separate experiment installs an evolved harness as the outer improver. Weco reports seven accepted improvements over 100 unattended steps and transfer to external tasks. Its stronger test of whether the evolved agent improves the outer search faster found no statistically significant efficiency advantage. These company-reported results support bounded harness improvement. Private evaluation, task families, and cost constraints remain human-designed, while reliable acceleration of successive improvers remains unestablished.

Across these configurations, the potential saving comes from reusing a better research procedure. Evidence of reduced human development effort requires accounting for the work spent designing evaluators, investigating failures, and maintaining the evolved code. Task performance or unattended runtime alone cannot quantify that saving.

\subsubsection{Evaluating L5 Recursive Improvement}
Evaluation must distinguish a stronger current agent from a procedure that reliably produces stronger successors. SEA-Eval~\cite{jiang2026sea} and SEAGym~\cite{zheng2026seagym} contribute trajectory-based assessment: performance across rounds, held-out transfer, retention, and resource use. Table~\ref{tab:l5-evaluation} summarizes these measurements and the additional mechanism tests needed for L5.

\begin{table*}[!t]
  \centering
  \caption{Evaluating recursive improvement.}
  \label{tab:l5-evaluation}
  \scriptsize
  \setlength{\tabcolsep}{3pt}
  \renewcommand{\arraystretch}{1.12}
  \renewcommand{\tabularxcolumn}[1]{m{#1}}
  \begin{tabularx}{\textwidth}{@{}>{\raggedright\arraybackslash}m{0.14\textwidth}>{\raggedright\arraybackslash}m{0.48\textwidth}>{\raggedright\arraybackslash}X@{}}
    \toprule
    \textbf{Dimension} & \textbf{Measurements} & \textbf{Purpose} \\
    \midrule
    Adaptivity & Gain; improvement trajectory; time to target~\cite{jiang2026sea,zheng2026seagym} & Detect progress and plateaus \\
    \addlinespace[2pt]
    Retention & Replay loss; tasks fixed or broken~\cite{zheng2026seagym} & Detect displaced capabilities \\
    \addlinespace[2pt]
    Transfer & Held-out gains within and across domains~\cite{zheng2026seagym} & Test reuse beyond update tasks \\
    \addlinespace[2pt]
    Efficiency & Tokens; time; cost per validated gain~\cite{jiang2026sea,zheng2026seagym} & Account for improvement expense \\
    \addlinespace[2pt]
    Stability & Harmful updates; largest temporary decline~\cite{jiang2026sea,zheng2026seagym} & Expose unreliable trajectories \\
    \addlinespace[2pt]
    Meta-recursion & Mechanism reuse; successor quality~\cite{zelikman2024stop,zhang2026hyperagents,iacob2026redqueen}; goal and stopping decisions & Test inherited improvement capacity \\
    \bottomrule
  \end{tabularx}
  \par\vspace{3pt}{\scriptsize\raggedright\textit{Note:} Trajectory measures synthesize the cited benchmarks. The meta-recursion row adds our L5 tests: cited systems illustrate mechanism reuse and improvement capacity; goal selection and stopping remain evaluation requirements.\par}
\end{table*}

A mechanism audit should identify the revised artifact, record the evidence that motivated it, and verify its invocation in a later improvement round. To establish effectiveness, the original and revised mechanisms should start from comparable agents and evidence under matched budgets, including the cost of evaluating the mechanisms. Holding the revised mechanism fixed during transfer tests helps isolate what it learned about improvement. HyperAgents~\cite{zhang2026hyperagents} uses this design to distinguish transferable agent-generation ability from performance on the original task.

Goal selection needs complementary evaluation. Adaptive frontier tasks can test whether the system identifies useful next challenges; protected reference tasks preserve comparability and expose forgetting or reward hacking. Private tasks with previously undisclosed rules strengthen tests of discovery beyond familiar distributions. Evaluations should also record whether the system stops when expected benefit falls below cost or risk. These are proposed requirements for assessing fuller L5 autonomy, rather than capabilities established by existing RSI benchmarks.

Current results support bounded meta-improvement: accepted changes to improvers, evaluators, or research policies can govern subsequent rounds, and some changes transfer across tasks~\cite{zelikman2024stop,zhang2026hyperagents,iacob2026redqueen}. Statistically reliable accumulation across generations under comparable resources remains open~\cite{zhang2026hyperagents,weco2026aide2}. The practical objective is to increase validated improvement per unit of computation and human effort. Strong base models, execution infrastructure, and independent evaluation continue to supply the resources and constraints under which that improvement occurs.

\begin{center}
\fbox{
\parbox{0.92\linewidth}{
\textbf{Finding: L5 makes the improvement procedure inheritable.}
The loop closes through reuse of a revised improver, evaluator, or research policy. Successors inherit that procedure and its supporting evidence; humans retain the overall objective, protected acceptance criteria, and resource authority.
}}
\end{center}

\subsection{Cross-Level Synthesis}

Across B0--L5, the autonomy hierarchy can be understood as a progressive transfer of responsibility over the improvement loop. The levels differ not primarily in the particular algorithm or artifact being optimized, but in which improvement decisions are internalized by the AI system, what state persists into later rounds, and which acceptance conditions remain externally protected.

The central boundaries between adjacent levels can therefore be stated compactly. The transition from B0 to L1 is \textbf{persistence}: an accepted change must survive the current task. L1 to L2 transfers \textbf{strategy selection}: AI decides which intervention to attempt rather than merely executing a prescribed one. L2 to L3 adds control over the \textbf{future learning agenda}, so the evolving learner influences what experience is acquired next. L3 to L4 extends this feedback process into \textbf{persistent deployment and environmental interaction}, where system changes must remain useful under changing operational conditions. L4 to L5 finally introduces \textbf{recursive inheritance}, in which the mechanism responsible for later improvement itself becomes an inherited target of improvement.

Importantly, a higher autonomy level does not by itself imply a better improvement process. Greater delegated authority can coexist with inefficient search, unreliable feedback, regression, evaluator exploitation, or poor transfer. We therefore separate the scope of improvement responsibility internalized by AI from the quality and efficiency of the resulting improvement. Across the evidence reviewed here, lower and intermediate levels are supported by a comparatively broad range of systems, whereas L3--L4 capabilities remain more domain-dependent and end-to-end L5 evidence is concentrated in bounded prototypes and emerging industrial or research systems. This distinction motivates the application-specific analysis in the following section.


\section{RSI Across Applications}
\label{sec:rsi-applications}

\begin{figure*}[!t]
    \centering
    \includegraphics[width=1\linewidth]{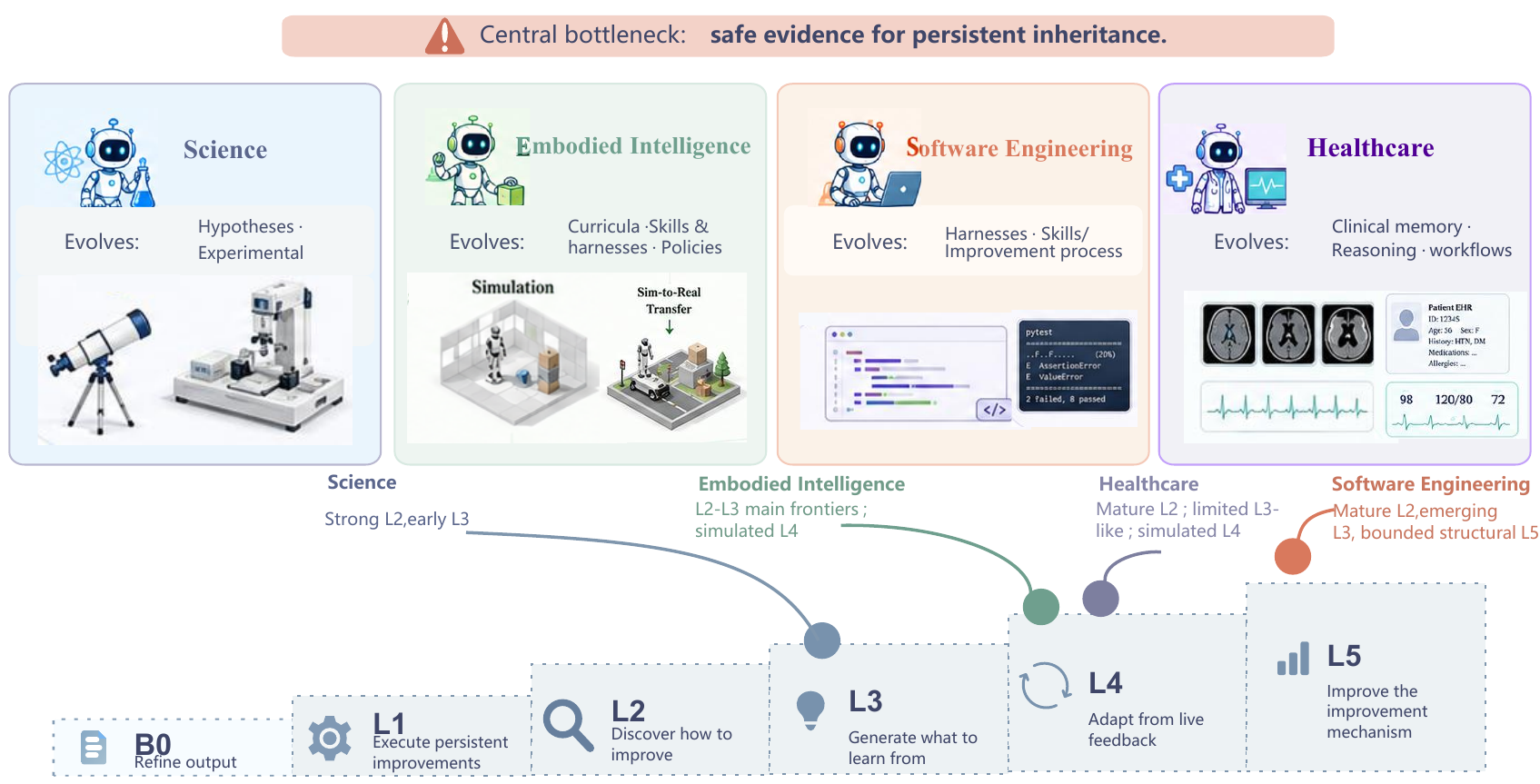}
    \caption{Application regimes differ mainly in the kind of feedback needed to validate and inherit an improvement.}
    \label{fig:rsi-applications}
\end{figure*}

Building on the B0--L5 taxonomy, this section summarizes how recursive self-improvement is instantiated under four domain-specific feedback regimes, including science, embodied intelligence, software engineering, and healthcare. Across these regimes, repeated refinement of an external artifact is insufficient: the feedback must produce a persistent change to the AI system that is reused in later tasks. Current evidence is therefore strongest for bounded, component-level improvement, whereas adaptation of the improvement mechanism itself remains rare.
A detailed review of the evolving components, supporting evidence, and remaining limitations in each application is provided in Appendix~\ref{app:rsi-applications}. 

\subsection{S1: RSI for Science}
\label{subsec:science-main}

Science is a consequential setting for RSI because progress depends not only on solving individual problems, but also on improving how future hypotheses are generated, tested, and revised. Its characteristic loop uses experimental evidence or falsification to update the AI4Sci system: a research agent proposes hypotheses, executes computational or physical experiments, attributes the observed outcomes, and retains validated changes to its generation strategy, tools, skills, memory, or scaffold. HypoForge~\cite{qian2026hypoforge}, for example, distills critiques and execution outcomes into reusable skills for hypothesis generation and testing. Test-Time Tool Evolution~\cite{lu2026tool} synthesizes and validates executable scientific tools for reuse, while DrugSAGE~\cite{zhang2026drugsage} transfers verified modeling procedures and error-recovery strategies across tasks. SIA~\cite{hebbar2026sia} extends the editable surface to agent scaffolds, tools, search logic, and model weights. These systems demonstrate persistent improvement of scientific-agent components, but their evaluators, update-selection rules, and promotion criteria remain largely fixed; improving a molecule, equation, or hypothesis alone therefore remains RSI-adjacent rather than evidence that the scientific agent has improved itself.

\subsection{S2: RSI for Embodied Intelligence}
\label{subsec:embodied-main}

Embodied intelligence grounds improvement in physical interaction and causal feedback. An embodied RSI loop selects or generates experience, executes a policy, evaluates the resulting trajectory, and commits validated updates to the policy, world model, skill library, harness, or curriculum used in subsequent interactions. POET~\cite{wang2019poet} co-evolves environments and agents so that newly generated challenges expose and extend current capabilities. Voyager~\cite{wang2023voyager} converts successful behaviors into an executable skill library that supports later exploration, and World-VLA-Loop~\cite{liu2026worldvlaloop} alternates between improving a world model from policy failures and optimizing the policy in the refined model. Together, these works move beyond within-episode correction toward persistent agent--environment co-adaptation. However, embodied experience is endogenous to the current policy, failures are difficult to attribute across perception, planning, control, and hardware, and real-world trials are costly and unsafe to reset. Most reported loops consequently rely on simulation or externally specified rewards, evaluators, interfaces, and safety constraints, leaving autonomous real-world validation and evolution of the embodied improver unresolved.

\subsection{S3: RSI for Software Engineering}
\label{subsec:software-main}

Software engineering offers unusually direct feedback because both the developed artifact and the coding agent can be represented, executed, versioned, and rolled back as code. Its RSI loop proceeds from a development challenge to an agent-level modification, repository-grounded execution, regression-aware selection, and versioned inheritance. SWE-Exp~\cite{chen2026sweexp} extracts lessons from successful and failed issue-resolution trajectories for reuse across repositories. Self-Harness~\cite{zhang2026selfharness} uses execution traces to propose modifications to its own harness and retains them only after regression testing. Darwin G\"odel Machine~\cite{zhang2026dgm} broadens this process by maintaining lineages of self-modifying coding agents and selecting variants according to executable task performance, thereby allowing accepted descendants to become the starting points for further improvement. These systems provide some of the clearest evidence of persistent source-, harness-, and experience-level evolution. Nevertheless, repository tests are incomplete proxies for user intent, security, maintainability, and descendant quality; benchmarks, utility functions, sandboxes, and selection rules also remain externally defined. Software-engineering RSI is therefore comparatively auditable, but still bounded by a fixed outer improvement contract.

\subsection{S4: RSI for Healthcare}
\label{subsec:healthcare-main}

Healthcare requires RSI to be grounded in longitudinal feedback from experts, clinical guidelines, diagnostic evidence, and patient outcomes. A clinical RSI loop performs a task, obtains validated feedback, updates a persistent component such as memory, reasoning policy, tool repertoire, or workflow, and reuses the approved update across later cases. MedAgent-Zero in Agent Hospital~\cite{li2024agenthospital} stores successful cases and derives reusable diagnostic rules from failures. EvoClinician~\cite{he2026evoclinician} uses process-level grading to revise how an agent acquires and integrates diagnostic evidence, while MACRO~\cite{fan2026macro} composes verified medical-imaging trajectories into reusable tools. HealthFlow~\cite{zhu2026healthflow} similarly retains validated safeguards, workflows, and code from completed health-data analyses. These approaches show how clinical experience can improve future agent behavior without necessarily changing foundation-model weights. Their evidence, however, remains dominated by simulated or retrospective settings, while real outcomes are delayed, heterogeneous, and confounded. Persistent clinical updates must therefore remain provenance-aware, population-specific, prospectively monitored, reversible, and subject to expert and institutional approval; autonomous evolution of the clinical improver is not yet established.


\section{Industry Landscape and Preliminary Practices}

\subsection{Theseus: Environment--Data--Model Co-Evolution for RSI Intelligence}
\label{sec:practice_theseus}

\newcommand{\theseuslogo}[1]{\raisebox{-2.2pt}{\includegraphics[height=7.5pt]{#1}}\hspace{0.25em}}
\newcommand{\openailogo}{\theseuslogo{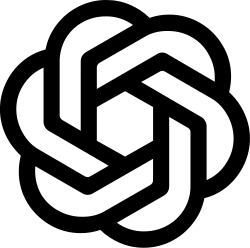}}
\newcommand{\claudelogo}{\theseuslogo{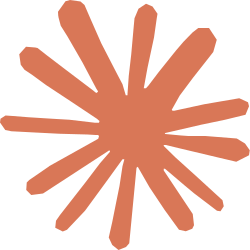}}
\newcommand{\deepseeklogo}{\theseuslogo{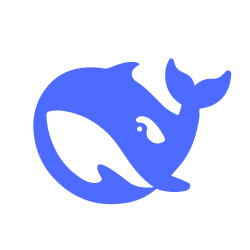}}
\newcommand{\pilogo}{\theseuslogo{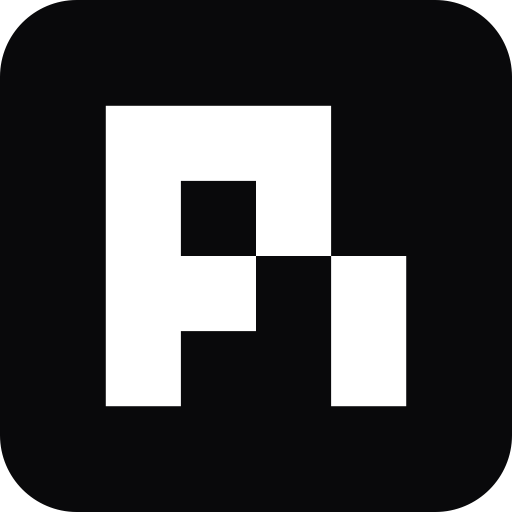}}
\newcommand{\glmlogo}{\theseuslogo{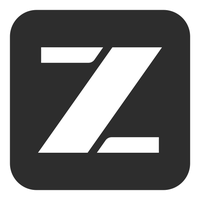}}
\newcommand{\kimilogo}{\theseuslogo{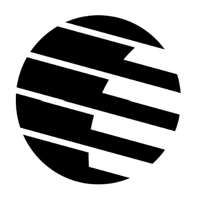}}
\newcommand{\groklogo}{\theseuslogo{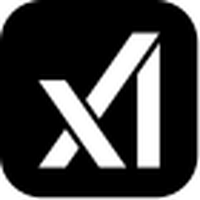}}
\newcommand{\metalogo}{\theseuslogo{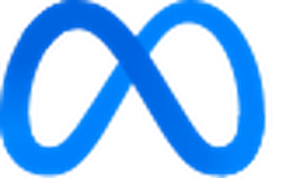}}

\begin{table*}[!t]
    \centering
    \footnotesize
    \renewcommand{\arraystretch}{1.22}
    \caption{{Early workspace experiments in Theseus's environment stage. \textbf{(a)} Clean-versus-noise pilot: pass rates (\%) of eight frontier model--harness configurations on 30 workspace tasks with 1,280 rubrics, comparing a clean workspace against a noise-laden workspace. \textbf{(b)} Productivity study: rubric scores of five fixed harness--model pairings on 30 tasks with 547 rubrics, comparing the bare workspace against the reconstructed environment (a shared package of a Collection Map and an Event Log); ``Env.\ tasks'' reports wins/draws/losses at the task level, and ``Rubric flips'' counts rubrics that changed from fail to pass versus pass to fail.}}
    \label{tab:theseus_results}
    \vspace{0.3em}

    \textbf{(a) Clean-versus-noise pilot} --- clean vs.\ noise-laden workspace\\[0.35em]
    {\setlength{\tabcolsep}{6pt}
    \begin{tabularx}{\textwidth}{@{}>{\raggedright\arraybackslash}X l c c c@{}}
        \toprule
        \textbf{Model} & \textbf{Harness} & \textbf{Clean} & \textbf{Noise} & $\mathbf{\Delta}$ \textbf{(pp)} \\
        \midrule
        \deepseeklogo DeepSeek-V4-Pro   & \deepseeklogo DSH        & 98.2 & 46.6 & $+51.6$ \\
        \deepseeklogo DeepSeek-V4-Flash & \deepseeklogo DSH        & 90.1 & 63.7 & $+26.4$ \\
        \openailogo GPT-5.6-Sol         & \openailogo Codex        & 89.1 & 54.1 & $+35.1$ \\
        \glmlogo GLM-5.3                & \claudelogo ClaudeCode   & 88.8 & 57.2 & $+31.6$ \\
        \kimilogo Kimi-K3               & \claudelogo ClaudeCode   & 87.6 & 53.7 & $+33.9$ \\
        \groklogo Grok-4.6              & \claudelogo ClaudeCode   & 84.7 & 63.0 & $+21.7$ \\
        \openailogo GPT-5.6-Luna        & \openailogo Codex        & 84.6 & 38.1 & $+46.5$ \\
        \metalogo Muse-Spark-1.2        & \claudelogo ClaudeCode   & 84.5 & 53.3 & $+31.2$ \\
        \bottomrule
    \end{tabularx}}

    \vspace{0.6em}
    \textbf{(b) Productivity-workspace study} --- bare vs.\ reconstructed environment\\[0.35em]
    {\setlength{\tabcolsep}{3.2pt}
    \begin{tabularx}{\textwidth}{@{}>{\raggedright\arraybackslash}X l c c c c c@{}}
        \toprule
        \textbf{Harness} & \textbf{Model} & \textbf{Bare} & \textbf{Reconstructed} & $\mathbf{\Delta}$ \textbf{(pp)} & \textbf{Env.\ tasks} & \textbf{Rubric flips} \\
        \midrule
        \openailogo Codex CLI  & \openailogo GPT-5.6-Sol        & 60.51\% & 92.50\% & $+31.99$ & 24/3/3 & 190/15 \\
        \openailogo Codex CLI  & \deepseeklogo DeepSeek-V4-Flash & 54.30\% & 84.83\% & $+30.53$ & 24/3/3 & 201/34 \\
        \claudelogo Claude Code & \deepseeklogo DeepSeek-V4-Flash & 61.06\% & 79.71\% & $+18.65$ & 23/2/5 & 140/38 \\
        \pilogo PI             & \deepseeklogo DeepSeek-V4-Flash & 62.52\% & 89.03\% & $+26.51$ & 22/6/2 & 155/10 \\
        \pilogo PI             & \openailogo GPT-5.6-Sol        & 50.27\% & 89.95\% & $+39.67$ & 23/2/5 & 240/23 \\
        \bottomrule
    \end{tabularx}}

\end{table*}

\begin{figure*}[!t]
    \centering
    \includegraphics[width=\textwidth]{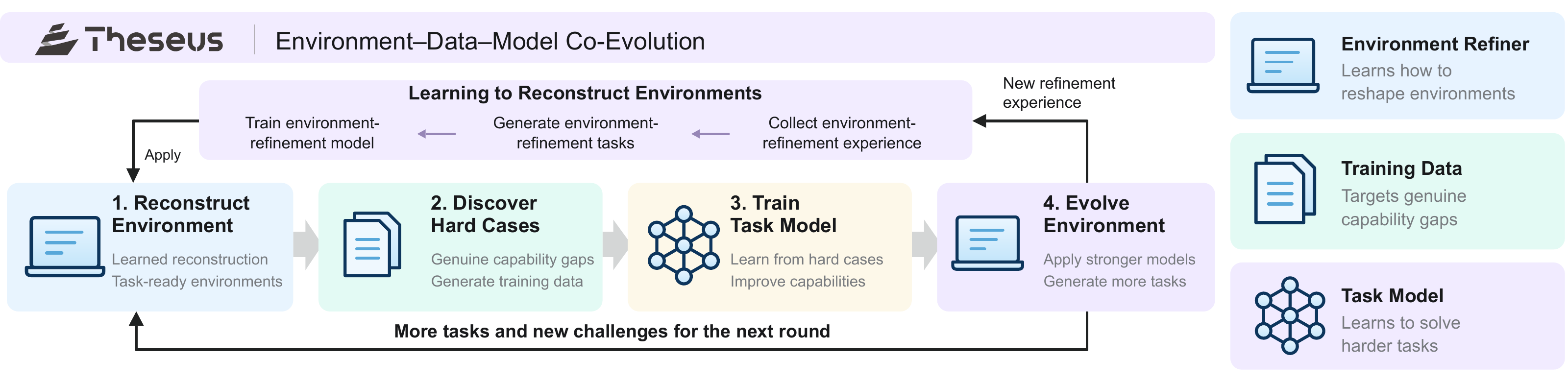}
    \caption{{Theseus's proposed four-stage co-evolution loop: reconstruct environments, uncover genuine capability gaps and generate training data, train task models, and iterate environments to produce more tasks. The upper loop learns an environment-refinement model from refinement experience and generated tasks, while successive iterations supply new experience for further refinement.}}
    \label{fig:bt_theseus}
\end{figure*}

{Theseus positions environment--data--model co-evolution as a path toward next-generation RSI intelligence. Its central premise is that environments make knowledge accessible and actions verifiable, task execution produces evidence for learning, and improved models expand the range of problems that can be explored. Under this vision, each round of work should contribute reusable capabilities to subsequent rounds, making real-world practice a continuing source of intelligence growth.} 

{As shown in Figure~\ref{fig:bt_theseus}, Theseus realizes this vision in four steps. First, refinement experience is converted into environment-refinement tasks to train an environment-refinement model. Second, agents use reconstructed environments to identify genuine task-solving difficulties and generate targeted training data. Third, these data train the task model to improve its capabilities. Fourth, the stronger model supports further environment iteration and the generation of more tasks. Subsequent rounds yield both new task-training data and new environment-refinement experience, linking progress in solving tasks to progress in constructing the conditions for future learning.} 

{Early results provide an initial empirical basis for this direction in the workspace stage of the loop. A pilot with eight frontier model--harness configurations on 30 workspace tasks adapted from Workspace-Bench~\cite{tang2026workspacebench}, scored by 1,280 rubrics, first measured sensitivity to workspace quality: relative to a noise-laden workspace, a clean workspace raised pass rates by 21.7--51.6 percentage points across all eight configurations (Table~\ref{tab:theseus_results}a), suggesting that workspace state, rather than model capability alone, can bound agent performance. Building on this observation, a follow-up productivity study asked whether enriching the environment improves task performance while keeping the underlying model and harness fixed: five fixed model--harness pairings solved 30 tasks scored by 547 rubrics in total under a bare workspace and a reconstructed environment consisting of a Collection Map and an Event Log; the reconstructed environment raised aggregate rubric scores by 18.65--39.67 percentage points across all tested configurations (Table~\ref{tab:theseus_results}b), although a few tasks showed scope-creep regressions, where agents produced more extensive changes than the task required. The team has also completed training a reusable file-verification module. These early results mark the first steps of the co-evolution loop, and successive rounds will extend environment reconstruction, data generation, and model training into a sustained cycle of compounding improvement.}


\subsection{Lark: Building the Data Foundation for Reliable Enterprise-Level RSI}

\begin{figure*}[!t]
    \centering
    \includegraphics[width=\textwidth]{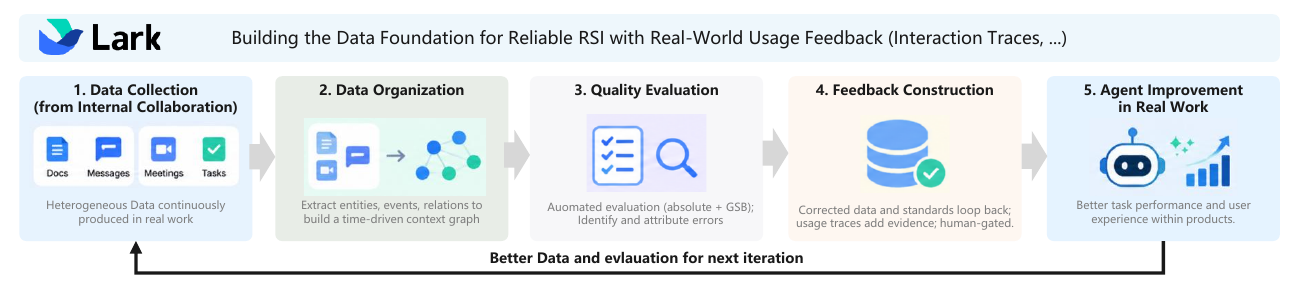}
    \caption{Lark’s data-foundation loop for RSI, where collaboration data are continuously structured, evaluated, and refined into high-quality knowledge and training signals, while real-world agent usage feeds new evidence back into subsequent improvement cycles.}
    \label{fig:bt_lark}
\end{figure*}

Figure~\ref{fig:bt_lark} summarizes Lark's data-foundation loop. Lark approaches RSI from a prerequisite that is often overlooked: before an agent can improve itself, it needs a continuously evolving data substrate on which improvement can be grounded. In enterprise collaboration, documents, messages, meetings, and tasks continuously generate new information, while agents consume these data and produce new interaction traces. Lark therefore treats data production, quality evaluation, and failure attribution as the foundation of its RSI pipeline. The key objective is not merely to accumulate more data, but to ensure that each iteration has fresh experience to learn from, a reliable criterion for judging progress, and sufficiently fine-grained diagnostics to determine what should be changed.

\textbf{Enterprise Knowledge Graph.} A representative example is the construction and continual refinement of an enterprise knowledge graph. Information scattered across documents, messages, and meeting records is linked into entities, events, people, and temporal relations so that agents can answer questions requiring cross-source reasoning. Because graph errors directly propagate into downstream answers, graph construction is itself treated as an iterative data-quality problem. Lark addresses scalability through batched graph construction and delegated authorization checks, while continuously evaluating whether graph-based retrieval improves downstream question answering. In internal evaluations, it reports that human-rated task usability increased from 52\% to 65\%, while automated-evaluation usability increased from 47\% to 56\% when using its graph-based pipeline compared with the RAG-based baseline.

\textbf{Automated Data-Quality Evaluation.} Another example is an automated data-quality evaluation system that serves as the measurement layer for subsequent improvement. Rather than relying entirely on expensive and inconsistent manual review, Lark first applies absolute judgments to identify clearly unusable outputs and then uses comparative GSB evaluation to determine whether a new system variant should be promoted. Importantly, evaluation is coupled with explicit failure attribution. Bad cases are categorized into issues (e.g., temporal inconsistency, distorted queries, or poor source quality), so that each failure maps to a concrete improvement direction. For example, time-sensitive questions are constrained by query-time filtering to avoid using future information, and low-quality synthetic queries are rewritten before re-entering the evaluation pipeline. The initial automated evaluator achieved approximately 84\% agreement with human judgments, where most disagreements arose because the automated evaluator was stricter than human reviewers, providing a concrete signal for subsequent calibration.

From an RSI perspective, the significance of these trials lies in the feedback structure they create for future agent improvement. Enterprise activity continuously produces new graph data and interaction traces, automated evaluation determines which outputs and data are reliable, error attribution identifies where the pipeline failed, and corrected data and evaluation standards then become inputs to later iterations. The resulting cycle turns production data from a passive resource into an evolving improvement substrate. At the same time, Lark's current practice remains deliberately human-gated: agents perform much of the repetitive data processing and evaluation, while humans retain responsibility for defining standards, reviewing critical samples, and approving important changes.


\subsection{Xiaohongshu: Dual-Timescale RSI for Recommendation}

\begin{figure*}[t]
    \centering
    \includegraphics[width=\textwidth]{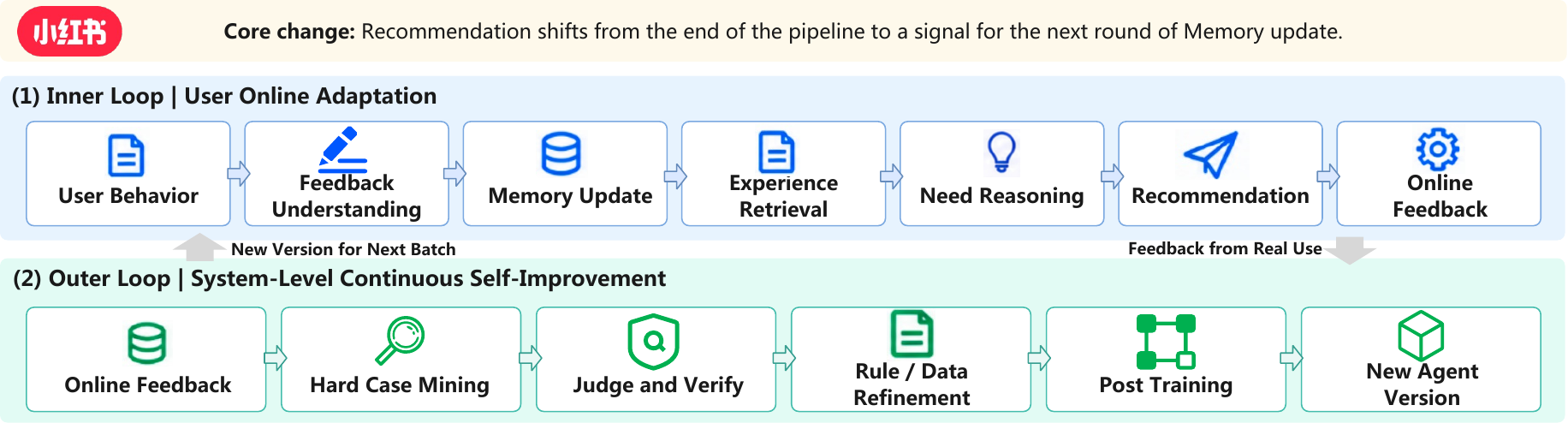}
    \caption{Xiaohongshu’s dual-timescale RSI loop for recommendation. The Intent-Memory Agent (IMA) uses online behavioral feedback to update structured user memory and guide subsequent recommendations. At a slower timescale, hard cases undergo automated judging, human spot checks, and data reconstruction, supporting post-training updates that consolidate validated experience into agent parameters before redeployment.}
    \label{fig:bt_xiaohongshu}
\end{figure*}

\begin{table}[t]
    \centering
    \caption{Performance comparison between the online baseline
    and the RSI model at Xiaohongshu.}
    \label{tab:xiaohongshu_rsi}

    \small
    \setlength{\tabcolsep}{7pt}
    \renewcommand{\arraystretch}{1.2}

    \begin{tabular}{@{}llccc@{}}
        \toprule
        Module
        & Metric
        & \makecell{A: Online baseline}
        & \makecell{B: RSI model}
        & Improvement \\
        \midrule

        Discovery Feed
        & \texttt{score\_mean@16}
        & 2.211
        & 2.366
        & $+7.0\%$ \\

        Discovery Feed
        & \texttt{score\_median@16}
        & 1.587
        & 1.663
        & $+4.8\%$ \\

        \addlinespace[3pt]

        In-video Feed
        & \texttt{score\_mean@16}
        & \makecell{1.537 ($n=470$)}
        & \makecell{1.739 ($n=447$)}
        & $\approx +13.1\%$\textsuperscript{*} \\

        \bottomrule

    \end{tabular}
\end{table}

Xiaohongshu’s commercialization recommendation system explores feedback-driven self-improvement through an Intent-Memory Agent (IMA). Rather than treating user understanding as a one-off inference from historical behavior, IMA connects feedback interpretation, memory updating, experience retrieval, need reasoning, and recommendation into a continuous cycle. Subsequent clicks, dwell time, saves, searches, conversions, and negative feedback serve not only as new behavioral inputs, but also as observations against which earlier estimates of user state and intent are checked. These observations guide the addition, reinforcement, replacement, or decay of memory, as well as decisions to ignore uninformative evidence. The persistent improvement target is therefore the user representation that informs future recommendations, rather than only the current recommendation output.

\textbf{Structured Memory and Experience Reuse.} IMA organizes its intermediate state into three semantic layers: Content, describing the content topic; Who, representing the user’s state; and Need, specifying the inferred requirement. These layers also structure knowledge-base storage and retrieval, allowing historical experience to be recalled through content similarity, user-state similarity, or need similarity. This decomposition is intended to reduce retrieval errors caused by mixing distinct semantic factors in a single caption representation. It also enables subsequent feedback to correct particular state variables rather than indiscriminately rewriting the entire user profile.

\textbf{Fast Adaptation and Slower Capability Consolidation.} The system combines two complementary improvement cycles. At the user level, real-time feedback updates memory to accommodate changing interests and temporary needs. At the system level, high-value hard cases collected from online interactions undergo automated judging, human spot checks, rule correction, and training-data reconstruction. The resulting examples support post-training updates to IMA, and the updated agent returns to the recommendation pipeline. Long-tail experience can thus become immediately available through the knowledge base, while repeatedly validated patterns are gradually consolidated into model parameters. This dual-timescale design connects rapid, non-parametric adaptation with more persistent model-level learning.

\textbf{Reported Observations.} In the company-provided comparison, Discovery Feed’s score\_mean@16 increases from 2.211 to 2.366, a reported 7.0\% improvement, while score\_median@16 increases from 1.587 to 1.663, or 4.8\%. For the in-video feed, score\_mean@16 increases from 1.537 to 1.739, approximately 13.1\%. The latter result comes from a small offline evaluation with 470 baseline samples and 447 RSI-model samples. These findings should be interpreted as preliminary, company-reported ranking-score improvements; the supplied material does not establish corresponding gains in click-through or conversion rates.

The case illustrates an industrial RSI pattern in which recommendation outcomes become evidence for revising the system that produces subsequent recommendations. Online feedback corrects persistent user memory, while reviewed hard cases improve later agent versions through post-training. Its practical contribution is the coupling of immediate state correction, structured experience reuse, and longer-term capability consolidation, with human inspection retained in the training-data refinement process.

\subsection{Humanlaya: Delivery-Driven RSI for Data Quality Assurance}
\label{subsec:humanlaya}

\begin{figure*}[!t]
    \centering
    \includegraphics[width=\textwidth]{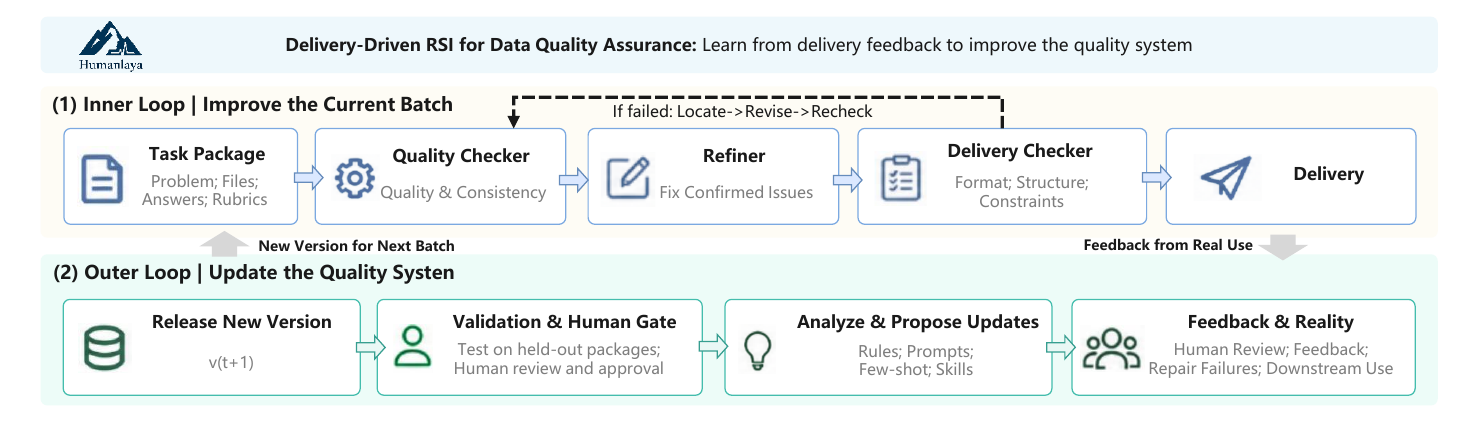}
    \caption{Humanlaya's delivery-driven RSI loop for data-quality assurance.}
    \label{fig:bt_humanlaya}
\end{figure*}

Humanlaya provides training and evaluation data for foundation-model developers, with its production workflow centered on complex multi-file task packages consisting of task descriptions, input attachments, reference answers, scoring rubrics, and executable or procedural verification requirements. In such settings, quality failures are often subtle, where an individual artifact may appear correct in isolation while remaining inconsistent with the task specification, scoring logic, or downstream evaluation protocol. As task types and customer requirements evolve, repeatedly repairing individual defective samples is insufficient, and the quality-control system itself must learn from recurring failures. Humanlaya therefore treats its data-quality agents and their associated rules, prompts, few-shot examples, and skills as persistent improvement targets, including an inner loop and an outer loop, as shown in Figure~\ref{fig:bt_humanlaya}. 

\textbf{The Data-Centric Inner Loop.} The inner loop operates within the current delivery batch and focuses on repairing the data being produced. Each task package first passes through a Quality Checker that inspects content quality and cross-component consistency, followed by a Refiner that repairs confirmed defects, and finally a Delivery Checker that verifies formatting, structure, and delivery constraints. Failed checks trigger repeated localization, revision, and re-verification before delivery. For example, if several independently gradable results are incorrectly merged into a single coarse rubric item, the checker may identify the rubric as insufficiently atomic and request restructuring before the package is accepted. This loop improves the current artifact, but by itself does not constitute persistent system improvement.

\textbf{The Improvement-Centric Outer Loop.} The outer loop operates across delivery batches and is the central RSI mechanism. After delivery, the system aggregates evidence like internal review, customer feedback, and downstream usage.  An agent analyzes these cases for recurrent causes and proposes versioned modifications to the quality-control system, including components like Refiner instructions, prompts, few-shot examples, and reusable skills. Candidate updates are evaluated on task packages that are not involved in producing the modification and are additionally reviewed by humans before promotion. Once approved, the new quality-system version replaces the previous one and is used for subsequent production batches. New tasks then generate new evidence, which again enters the next improvement cycle.

A representative case involved a three-page scanned PDF that the extraction tool failed to parse, causing the Quality Checker to mark the attachment as unusable. Manual review shows that the scan itself is clear, while the error comes from conflating extraction failure with poor document quality. The system therefore generates a new decision rule and few-shot examples to distinguish the two cases. After held-out validation and human review, the update is incorporated into the next system version and reused on future tasks.
The inner loop repairs the current task package, whereas the outer loop updates the method used to inspect and repair future packages. Humanlaya therefore represents a practical form of scaffold-level RSI driven by production feedback.

\textbf{Measured Improvement.} Humanlaya reports an internal comparison between V0 and V4 after four feedback-driven updates. Using the same base model, tools, and processing budget, both versions were evaluated on 600 task packages excluded from the update process. The rate of packages containing key defects after automated repair decreased from 9.0\% to 3.7\%, while average human handling time fell from 48 to 27 minutes per task. 

At the current stage, when model capabilities are still advancing rapidly, some tasks cannot yet be completed fully autonomously by agents. Humanlaya therefore provides a practical example of combining RSI with human oversight, where agents perform most of the self-improvement process, while humans provide partial ground truth, review, and final approval at critical stages, rather than allowing an unconstrained self-improvement loop.

\subsection{ModelBest: Zero-Human Industrial AI Engineering}

\begin{figure*}[t]
    \centering
    \includegraphics[width=\textwidth]{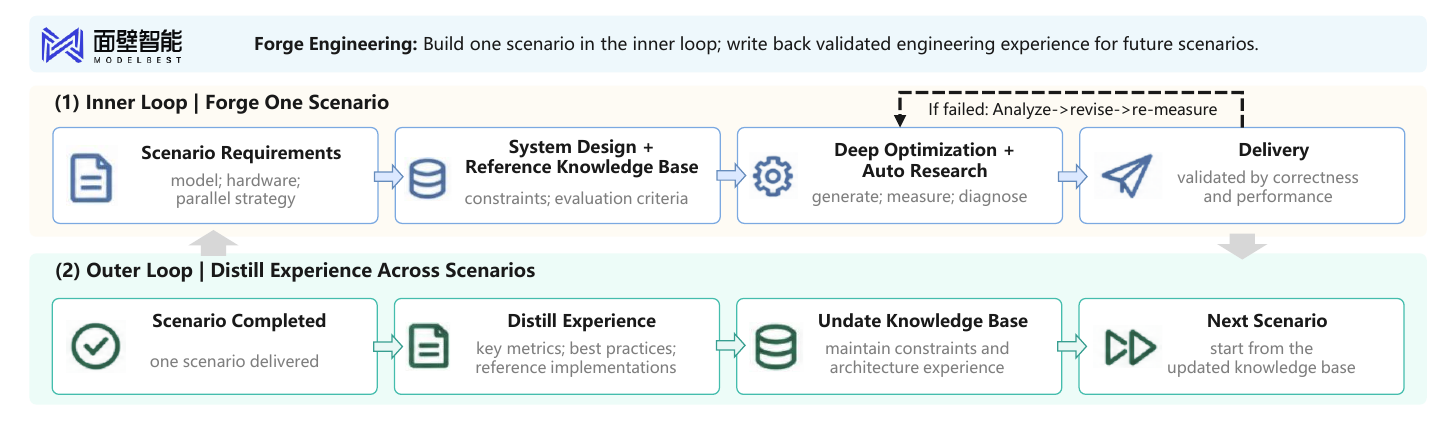}
    \caption{
    ModelBest: Forge Engineering as a two-level RSI loop.}
    \label{fig:bt_modelbest}
\end{figure*}

ModelBest aims to advance RSI by automating AI engineering, based on the premise that \textit{AI engineering is likely to become autonomous earlier than open-ended AI research}. Engineering tasks (e.g., implementing training frameworks, optimizing kernels, configuring parallelism, running benchmarks) require substantial human effort but offer comparatively inexpensive, objective feedback through execution. To this end, to produce a production-ready implementation without human intervention, ModelBest proposes \textit{{Forge Engineering}}, an end-to-end paradigm in which an AI system takes a target model, hardware platform, and parallelization requirements as inputs, then builds, tests, and refines an implementation from an initially empty codebase. Figure~\ref{fig:bt_modelbest} shows how its within-project optimization and cross-project experience reuse form a two-level loop. 

\textbf{The Industrial RSI Loop.} Within a target scenario, the system combines constrained architecture design with autonomous low-level optimization. Humans and AI maintain a knowledge base containing specifications, evaluation criteria, and accumulated engineering experience. The agent first uses this knowledge to establish a reasonable high-level architecture and constrain the search space. An AutoResearch loop then repeatedly generates implementations, measures correctness and performance, diagnoses failures, and repairs bottlenecks. Successful optimizations, including kernels such as GEMM and FlashAttention, are progressively integrated into the main execution path. This design keeps global architectural constraints stable while allowing aggressive autonomous optimization at the implementation level.

Across scenarios, successful and failed engineering experience is written back into the shared knowledge base. Performance measurements, effective optimization strategies, reference implementations, and failure rules from one project become starting knowledge for subsequent projects. The same engineering procedure can therefore be reused across different models, hardware platforms, and workloads. ModelBest reports applying the paradigm across pre-training frameworks, operator libraries, reinforcement-learning infrastructure, inference engines, fine-tuning, compression, quantization, and edge deployment on several hardware ecosystems.

\textbf{Measured Improvement.} Starting from an empty directory together with reference scripts and model specifications, ForgeTrain~\cite{forgetrain2026} generated a pre-training framework that matched Megatron-LM v0.15 within roughly 8 hours and reportedly surpassed it within 1.5–2.5 days, compared with an estimated 3–5 engineers working for 6–12 months for comparable manual development. Reported model FLOPs utilization increased from 40.1\% to 44.1\% for MiniCPM4-0.5B and from 47.0\% to 50.9\% for the 8B model. ForgeStencil~\cite{forgestencil} extends the same approach to scientific-computing kernels, with the company reporting 1.15–1.9× speedups over public state-of-the-art implementations and a median 1.41× end-to-end speedup.


\subsection{Tencent Hunyuan: Experience-Driven Self-Improvement}

\begin{figure*}[t]
    \centering
    \includegraphics[width=\textwidth]{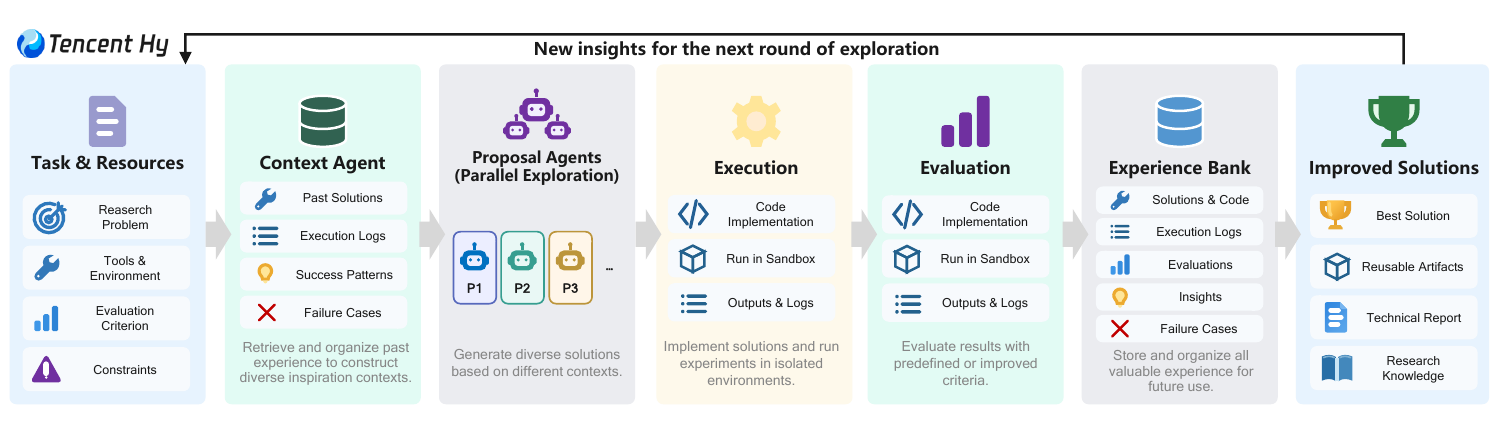}
    \caption{Experience-driven self-improvement in Tencent Hunyuan Hyra. A Context Agent organizes prior experience to guide parallel solution proposals, which are executed in isolated environments and evaluated. The resulting code, execution logs, and evaluation feedback are retained in an Experience Bank to inform subsequent exploration and support evaluator refinement when needed.}
    \label{fig:bt_hunyuan}
\end{figure*}

Tencent Hunyuan's Hyra (Hunyuan Research Agent) explores RSI in performance-oriented research and engineering tasks. Figure~\ref{fig:bt_hunyuan} outlines its experience-driven search loop. Its design follows a deliberately lightweight philosophy: rather than encoding increasingly elaborate human-designed workflows, Hyra gives agents a broad solution space and repeatedly uses experimental evidence to guide subsequent search. Given a task, the system continues exploring until it decides to terminate or exhausts its resource budget, returning the best solution discovered during the process.

A central component is the Experience Bank, which retains substantially richer state (e.g., previous solution code, generated artifacts, execution logs, scores, and evaluator feedback) than a conventional textual memory. A Context Agent recombines this evidence into diverse inspiration contexts, while multiple Proposal Agents asynchronously consume these contexts, construct new solutions, and execute them in fresh isolated sandboxes. The resulting artifacts and evaluation outcomes are written back into the Experience Bank, allowing later proposals to directly reuse, combine, or avoid patterns discovered in earlier attempts.

Hyra further extends this idea to evaluation itself. For open-ended tasks where the initial evaluator is incomplete or becomes exploitable, accumulated search experience can be used to revise the evaluation mechanism—for example, by increasing its granularity, strengthening comparison baselines, or closing reward-hacking loopholes—before subsequent search continues under the improved criterion. This is particularly relevant to RSI because the persistent update is no longer limited to a better solution: the mechanism that determines which future solutions are considered improvements can also change.

\begin{table}[t]
    \centering
    \caption{Comparison of Recursive and hyra-1.0 on AI R\&D benchmarks.}
    \label{tab:hyra_benchmarks}
    \small
    \setlength{\tabcolsep}{5pt}
    \renewcommand{\arraystretch}{1.1}
    \renewcommand{\tabularxcolumn}[1]{m{#1}}

    \begin{tabularx}{\linewidth}{
        @{}
        >{\centering\arraybackslash}X
        >{\centering\arraybackslash}X
        >{\centering\arraybackslash}X
        c
        c
        @{}
    }
        \toprule
        \textbf{Benchmark}
        & \textbf{Task}
        & \textbf{Metric}
        & \textbf{Recursive}
        & \textbf{hyra-1.0} \\
        \midrule

        NanoChat Autoresearch~\cite{nanogpt_ar}
        & Model training
        & Validation BPB $\downarrow$
        & 0.9109
        & \textbf{0.9015} \\

        \addlinespace
        NanoGPT Speedrun~\cite{modded_nanogpt_2024}
        & Model training acceleration
        & Time to 3.28 loss $\downarrow$
        & 77.5\,s
        & \textbf{76.4\,s} \\

        \addlinespace
        SOL-ExecBench~\cite{lin2026solexecbench}
        & GPU kernel optimization
        & Mean SOL $\uparrow$
        & 0.754
        & \textbf{0.771}$^{\dagger}$ \\

        \bottomrule
    \end{tabularx}
\end{table}

\textbf{Measured Improvement.} As shown in Table~\ref{tab:hyra_benchmarks}, the released companion artifacts report improvements across both AI-for-AI and AI-for-Science tasks. For example, Hyra reports validation BPB of 0.9015 on nanochat AutoResearch~\cite{nanogpt_ar} compared with a cited previous best of 0.9109, 76.4 s on nanoGPT Speedrun~\cite{modded_nanogpt_2024} compared with 77.5 s, and 0.771 on SOL-ExecBench~\cite{lin2026solexecbench} compared with 0.754. As an industrial RSI practice, Hyra highlights the value of retaining executable experience rather than only final answers. 


\subsection{Agent-Native Research Lab: Verifiable Research Infrastructure for RSI}

\begin{figure*}[t]
    \centering
    \includegraphics[width=\textwidth]{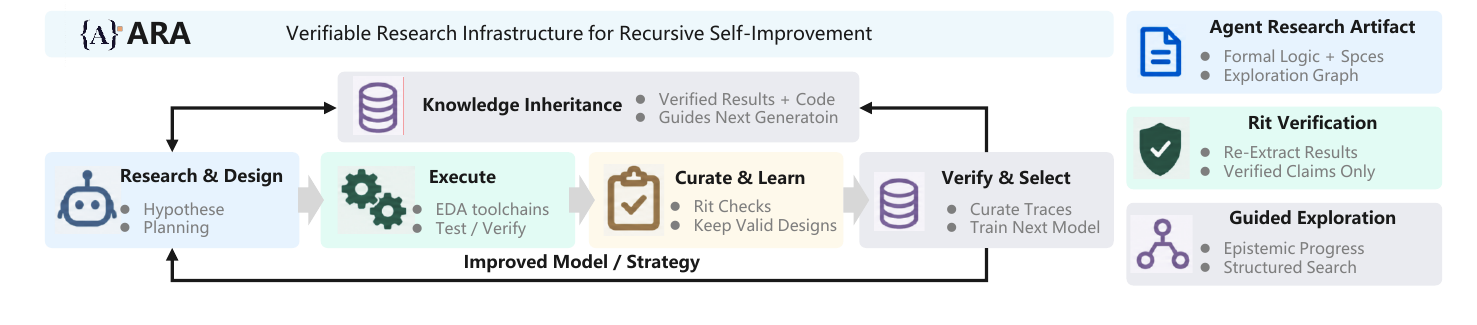}
    \caption{Agent-Native Research Lab's agent-native RSI loop for verifiable engineering discovery, integrating guided exploration, deterministic verification, knowledge inheritance, and next-generation learning to support persistent improvement across research cycles.}
    \label{fig:bt_amber}
\end{figure*}

Achieving sustained recursive self-improvement requires more than increasingly autonomous research agents, while it also requires an infrastructure through which research experience can be reliably recorded, verified, inherited, and reused across generations. Agent-Native Research Lab focuses on this infrastructural layer, summarized in Figure~\ref{fig:bt_amber}. 

$\bullet$ \textbf{(1) Human-Centric vs. Agent-Native Research Artifact.} Conventional scientific papers are designed for human communication and would discard much of the operational state that autonomous agents need to continue previous work (e.g., failed experiments, intermediate evidence, executable specifications, and implementation details). For RSI, such information loss directly weakens cross-generation knowledge accumulation.

To address this, they propose Agent-Native Research Artifact (ARA) as an executable alternative to conventional scientific documentation. ARA jointly preserves formal scientific logic, executable code and specifications, the exploration graph of both successful and abandoned branches, and raw empirical evidence. Rather than inheriting only a polished final narrative, a successor agent can therefore inspect how a result was obtained, which alternatives failed, and which assumptions or parameters were actually used. The reported evaluation shows 93.7\% question-answering accuracy over prior work using ARA compared with 72.4\% using conventional papers, while RE-Bench reproduction success increases from 57.4\% to 64.4\%. The important RSI contribution is thus not merely better documentation, but a richer inheritance substrate for later research cycles.

$\bullet$ \textbf{(2) Deterministic Verification for Research Claims.} Additionally, as autonomous agents generate experiments and claims at machine speed, \textit{verification rather than generation becomes the main bottleneck}. Its rit protocol anchors empirical claims directly to execution traces, re-extracting reported results from logs, while analytical claims can be checked through formal verification systems such as Lean 4. Only claims that pass these machine-verifiable gates are admitted into the shared research state. This prevents hallucinated or incorrectly reported results from being recursively inherited and amplified by later generations. They also augment sparse outcome-based optimization with epistemic-progress guidance and structured priors derived from human scientific reasoning. Instead of rewarding only the final benchmark score, agents are encouraged to prefer experiments that reduce uncertainty and reveal useful structure in the problem. This is intended to improve research efficiency in large hypothesis spaces, where naive hill climbing can repeatedly exploit familiar parameters without producing deeper understanding.

\textbf{Measured Improvement.} These ideas are instantiated in silicon design, where the feedback loop is unusually fast and deterministic. Autonomous agents generate synthesizable SystemVerilog, construct testbenches, and propose microarchitectural modifications; candidates are then evaluated using industrial EDA tools for synthesis, place-and-route, timing analysis, and formal equivalence. Invalid designs are filtered automatically, while verified implementations, execution traces, failed branches, and superior power-performance-area trajectories are retained as training and research evidence for subsequent generations. On Chip-Bench, they report a Level-3 CPU optimization score of 5.8416, with a design that is 2.7\% faster and 21.5\% smaller in area than standard-agent baselines on the same model foundation while passing all 97 directed tests and 350 random differential programs.

\subsection{Frontis.AI: Enterprise Agent Evolution and Cross-Task Meta-Improvement}

\begin{figure*}[t]
    \centering
    \includegraphics[width=\textwidth]{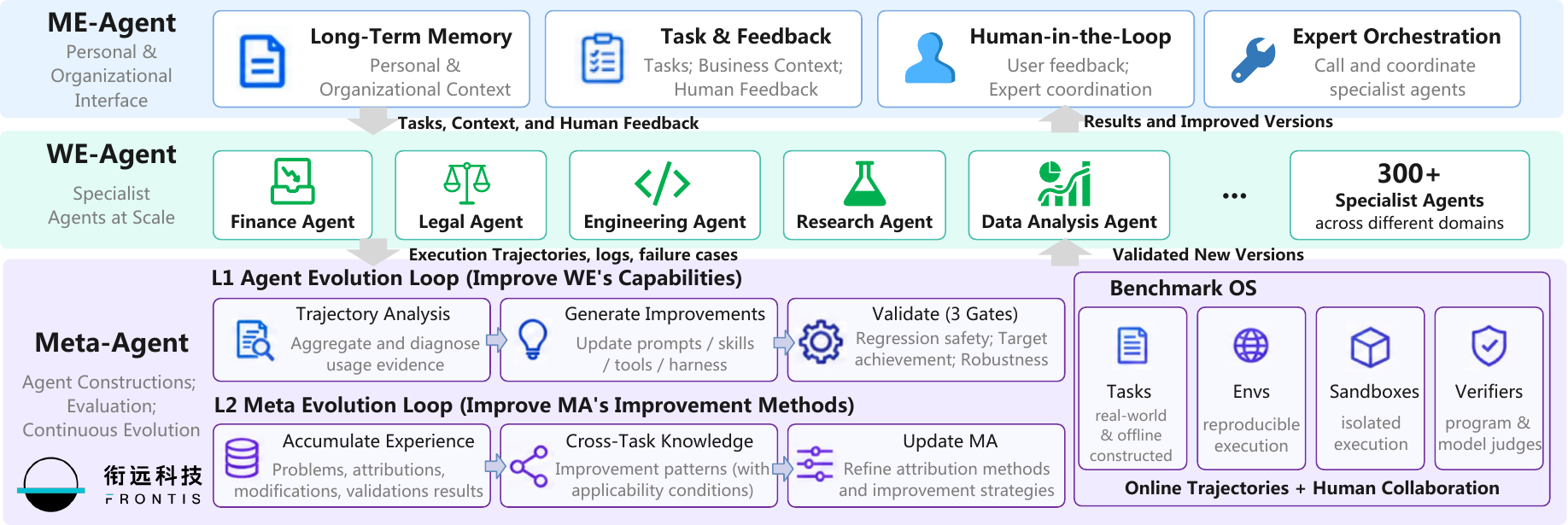}
    \caption{Frontis Horizon’s ME–WE–MA architecture combines specialist-agent evolution and cross-task meta-improvement with shared evaluation and human-approved release.}
    \label{fig:bt_frontis}
\end{figure*}

Frontis Horizon provides an enterprise-owned infrastructure for improving both specialist agents and the methods used to improve them. Its ME–WE–MA architecture separates human interaction, contextual memory, and release approval (ME); task execution by more than 300 heterogeneous specialist agents across finance, legal, engineering, and human resources (WE); and agent construction, evaluation, and continuous evolution (MA). The central objective is to make improvement experience reusable across tasks rather than allowing it to remain isolated within individual agents.

\textbf{Business-Grounded Expert Evolution.} The platform’s Benchmark OS organizes enterprise tasks, reproducible environments, and verifiers into shared evaluation infrastructure. Programmatic checks are preferred, while semantic quality is assessed by model judges isolated from the evaluated agents. Given operational feedback, MA attributes failures to routing, execution mechanisms, or specialist capabilities, defines bounded improvement objectives and acceptance criteria, and obtains human confirmation before modifying prompts, skills, tools, or harnesses. Candidate versions are compared with their predecessors through three gates: regression safety, target achievement, and robustness in exceptional scenarios. Approved updates return to production with versioned test reports supporting audit and rollback.

\textbf{Cross-Task Meta-Improvement.} Beyond improving individual specialists, the platform retains each improvement episode’s problem context, diagnostic rationale, candidate modifications, validation results, and applicability conditions. Agent-specific experience supports subsequent versions, while transferable lessons inform MA’s diagnosis and improvement strategies on new tasks. Transfer requires adaptation and revalidation: a lesson about prioritizing real-time evidence, for example, still requires identifying authoritative sources and their temporal validity in the receiving task. This distinguishes learning to perform a business task from learning how to improve agents more effectively.

\textbf{Reported Outcomes.} Over two months, Frontis reports initiating 216 improvement tasks involving 117 specialist agents, of which 63 completed at least one full evolution cycle. Across 76 comparable tasks evaluated with the same evaluator, average scores increased from 5.25 to 5.87, with a median machine-processing time of 27.8 minutes per completed cycle. Separately, after accumulating meta-evolution experience from more than 100 tasks, the company reports approximately 20\% faster evolution on previously unseen internal test tasks relative to a process without cross-task experience. Its evaluation protocol distinguishes specialist performance from improvement efficiency by comparing identical initial agent versions and tracking iterations, time, token consumption, and regressions. These findings remain company-reported evidence rather than independent replication.

\textbf{Learning Improvement in Model Weights.} Frontis-MA1 separately explores training improvement capabilities into a 35B model. The supplied report records an MLE-Bench Lite score increase from 39.39\% to 60.61\% under a fixed harness, reaching 71.21\% with additional search. These results concern machine-learning and deep-learning engineering; extending the approach to learn directly from the platform’s enterprise-agent evolution trajectories remains future work.

The case illustrates a practical RSI pattern: retain not only improved agents, but also validated knowledge about how to improve them. Shared evaluation infrastructure, context-sensitive experience transfer, versioned updates, and human-controlled release provide the governance framework for this accumulation.

\section{Challenges and Future Directions}

The preceding chapters show progress in automating improvement execution, strategy selection, experience acquisition, deployment adaptation, and the revision of improvement mechanisms. They also establish that greater autonomy, durable capability gains, and a more effective improvement process are distinct properties. The three bottlenecks motivating this survey therefore remain only partially resolved: foundation-model development requires substantial resources and coordination; scalable learning depends on useful experience and reliable verification; and deployed systems require continuing effort to diagnose, validate, and release updates. Building on the autonomy hierarchy, the four application regimes, and the industrial practices reviewed above, we identify eight research directions for connecting these partial advances into sustained recursive improvement.

\noindent\textbf{Cross-Component Diagnosis and Coordinated Improvement.}
The transition from L1 to L2 exposes a diagnostic problem: an observed failure rarely identifies the component that should change. An incorrect answer may originate in defective source data, missing context, a tool interface, the model, or the evaluator. The Humanlaya case in Section~\ref{subsec:humanlaya} makes this ambiguity concrete: a failed extraction was initially treated as evidence of poor document quality, while the appropriate intervention concerned the quality checker's decision rule. In science and embodied intelligence, the same problem extends to experimental protocols, perception, control, and environmental conditions. As several components adapt, changing one can also invalidate assumptions on which another relies.

Future systems should turn diagnoses into testable intervention hypotheses and use controlled comparisons to estimate individual effects and interactions. Selective component freezing, targeted ablations, and explicit records of dependencies could help distinguish a useful repair from compensation for an upstream defect. Coordinated search should then allocate effort across data, models, harnesses, and environments as constraints shift. Theseus's workspace studies in Section~\ref{sec:practice_theseus} illustrate the importance of separating environment improvements from changes to the task model; the reported component-level gains leave the full learning cycle to be established. The research goal is to identify which intervention improves the overall system, under what conditions, and whether the resulting knowledge guides later improvement decisions.

\noindent\textbf{Learner-Conditioned Experience Acquisition and Reliable Learning Signals.}
The L3 analysis distinguishes three properties of experience: whether it is valid, how difficult it is for the current learner, and whether learning from it produces a durable benefit. AZR uses executable checks alongside learner-dependent task proposal, whereas R-Zero uses solver consistency as a proxy for difficulty~\cite{zhao2025absolutezero,huang2026rzero}. Neither correctness nor estimated difficulty alone establishes learning value. An adaptive curriculum may repeatedly select ambiguous tasks, reinforce evaluator errors, or concentrate on a narrow region where progress is easy to measure. Such errors can enter both the learner and the state that determines its next experiences.

Future acquisition mechanisms should estimate learning value across multiple rounds while preserving validity, diversity, and coverage of previously acquired capabilities. This includes selecting from existing data, generating new tasks, seeking interactions, and requesting stronger external feedback when internal judgments are insufficient. A central challenge is deciding when expensive verification is worth its cost and how uncertain experience should influence subsequent training. Evaluation should allow the acquired distributions to differ while matching data-source access and total acquisition and learning budgets. Freezing the learner-state input to the acquisition mechanism, or replacing adaptive selection with a learner-independent schedule, can test whether learner conditioning improves transfer beyond the effects of additional training.

\noindent\textbf{Persistent-State Management and Reliable Reuse.}
L4 systems inherit parameters, memories, skills, tools, and harness code, but retaining an artifact does not establish that later agents benefit from it. The L4 analysis identifies failures in both activation and execution: a relevant skill may never be retrieved, or an agent may retrieve it without following it correctly. Accumulation creates an additional difficulty. Library Drift shows that an expanding skill library can degrade retrieval and stall improvement, while overly aggressive retirement can also be harmful~\cite{zhang2026library}. Persistence therefore requires managing the continuing applicability and interaction of retained changes.

Future work should associate inherited artifacts with their supporting evidence, applicability conditions, dependencies, and observed effects on later tasks. Admission tests such as HDSO's paired evaluations provide a starting point~\cite{shang2026hypothesis}, but validation must continue as the task distribution, executor, and surrounding system change. Studies should separately measure update quality, activation, faithful use, and downstream benefit, then investigate when to merge, revise, retire, or revalidate artifacts. Cross-model reuse introduces a further question: a skill useful to its author may be unsuitable for a different executor. Resolving these issues would connect successful update generation to reliable capability transfer across sessions and successors.

\noindent\textbf{Governed Adaptation under Domain-Specific Feedback.}
The four application regimes show why a common improvement loop requires different validation strategies. In science, an unsuccessful experiment may reflect the hypothesis, protocol, or instrument, and evidence must retain its assumptions and uncertainty. In embodied intelligence, the current policy determines which states are observed, while physical trials consume resources and can have irreversible consequences. Software offers executable feedback, but passing tests remains conditional on the specification and test coverage. In healthcare, outcomes are delayed and confounded, and an update validated in one population or institution may not transfer to another. These differences constrain both what a system can learn autonomously and what evidence supports deployment.

Future research should develop update policies that distinguish transient failures from recurring limitations and select a validation process appropriate to the proposed change. Simulation, retrospective replay, and sandboxed execution can support initial screening, followed by supervised or staged evaluation under the intended operating conditions. Monitoring should test whether benefits persist as requirements, populations, tools, and environments change. Restoring a software checkpoint cannot undo physical or clinical consequences, making pre-release validation essential in those settings. Human authority over consequential releases, access permissions, and safety constraints should remain explicit. The open problem is how to increase useful adaptation while controlling the propagation of errors across future interactions.

\noindent\textbf{Trustworthy Evolution of Improvement Mechanisms.}
At L5, changes to the improver, evaluator, or research policy alter how subsequent successors are produced and accepted. This creates a coupled validation problem: a stronger solver can expose weaknesses in its evaluator, but changing the evaluator can also make scores incomparable or reward exploitation. RQGM addresses part of this problem by freezing its evaluator within an epoch, validating replacements against an independent anchor, and revisiting scores affected by replacement~\cite{iacob2026redqueen}. The evaluator refinement described in Hyra raises the same broader question of how adaptive measurement can remain credible as search progresses.

Future systems should distinguish editable internal feedback mechanisms from independently maintained acceptance criteria and preserve evidence linking each mechanism revision to later decisions. Controlled comparisons should assess changes to candidate generation and evaluation separately before attributing gains to their joint evolution. A-Evolve-Training provides an example of policy inheritance within a fixed high-level objective~\cite{shi2026aevolvetraining}; whether such policies transfer beyond the research setting in which they evolved remains to be tested. Open questions include how to maintain independent assessment under repeated adaptive access, detect coordinated proposer--evaluator errors, and reverse harmful mechanism changes without losing useful experience. Revising operational research priorities need not transfer authority over the system's overall mission.

\noindent\textbf{Long-Horizon Evaluation of Inherited Improvement Capacity.}
The distinction between structural and effective L5 should guide evaluation. Structural evidence establishes that a revised mechanism is inherited and invoked; effectiveness requires showing that it produces or selects better subsequent improvements. This remains a substantive empirical gap. For example, the AIDE$^{2}$ experiment reviewed in this survey did not establish a statistically significant efficiency advantage when an evolved harness was installed as the outer improver~\cite{weco2026aide2}. Software lineage studies further motivate separating current task performance from the ability to produce useful descendants, since temporarily weaker variants may enable later progress.

Future benchmarks should compare original and revised mechanisms from comparable initial systems and evidence under matched total budgets, including the cost of developing and evaluating the mechanisms. Holding an evolved mechanism fixed on fresh tasks can test whether its improvement strategy transfers, as illustrated by HyperAgents~\cite{zhang2026hyperagents}. Longer studies should report complete trajectories across repeated runs, including rejected updates, regressions, recovery, retained capabilities, and resource use. Protected reference tasks can preserve comparability, while fresh tasks and unfamiliar constraints test transfer and discovery. These protocols should establish whether inherited changes alter subsequent improvement capacity, when gains plateau, and whether apparent acceleration survives accounting for additional search and evaluation effort.

\noindent\textbf{Resource-Aware Improvement and Human Collaboration.}
The efficiency objective motivating RSI requires accounting for the entire improvement process. A faster kernel, a better training recipe, or a longer unattended run may still require substantial candidate generation, evaluation, infrastructure maintenance, and human review. Meta's regression-analysis workflow retains engineering review of proposed changes~\cite{meta_capacity_efficiency_2026}; Humanlaya also retains human validation and reports handling time alongside quality. These practices show why autonomy should be assessed together with the amount and type of human effort required. Stronger base models and more parallel trials can further confound comparisons between improvement strategies.

Future work should study how to allocate a total budget among diagnosis, experience acquisition, candidate search, verification, and deployment monitoring. Cheap screening can reduce expensive trials, provided its predictions remain calibrated against independent outcomes. Reusing validated tools and prior failures may reduce repeated work, but its maintenance cost must also be counted. Evaluations should report time and total cost to a validated capability target, alongside review effort and rework. Adaptive stopping is another research priority: systems should justify continued experimentation by its expected learning value and uncertainty, with explicit conditions for pausing, escalating to a human, or terminating an unproductive search.

\noindent\textbf{Reproducible Infrastructure for Cross-Round Inheritance.}
Successor systems need access to how an improvement was obtained, including unsuccessful alternatives and the conditions under which its evidence holds. The industrial cases expose complementary requirements: Lark emphasizes temporally grounded data and failure attribution; Humanlaya retains versioned quality-system updates; Hyra stores executable experience; and Agent-Native Research Lab proposes research artifacts that preserve code, specifications, exploration history, and empirical traces. These practices motivate a shared infrastructure for carrying research evidence across rounds, while the full longitudinal benefit of such inheritance remains to be established.

A reusable artifact format should connect the parent state, proposed change, motivating evidence, evaluation configuration, acceptance decision, and subsequent use. Versioned data, environments, and evaluators would make results easier to replay and clarify which comparisons remain valid after an update. Deterministic re-extraction can check reported measurements, and formal verification can establish properties relative to a specification; neither alone establishes that the experiment or specification captures the intended capability. Public evaluation should therefore pair inspectable artifacts with replay studies and report result provenance, artifact availability, and independent replication separately. Where data are confidential, shareable evaluation subsets and controlled replay interfaces could support partial verification while preserving the limits of what can be checked.

Across these directions, progress toward genuine RSI should be demonstrated through repeated, attributable, and transferable improvements in the capacity to improve. The decisive evidence is that inherited changes help later rounds acquire more useful experience, discover better interventions, or validate successors more effectively under explicit resource and authority constraints.

\section{Conclusion}

In this report, we present an autonomy-centered roadmap for recursive self-improvement, spanning five stages: improvement execution, strategy selection, experience acquisition, environmental adaptation, and recursive meta-improvement. Taking the improvement loop as the unit of analysis, we clarify what systems change, what successors inherit, and which decisions remain human-controlled. We connect this roadmap to scientific discovery, embodied intelligence, software engineering, and healthcare, showing how feedback availability, verification costs, and deployment constraints shape progress toward RSI in each domain. Industry practices and preliminary empirical findings further ground the roadmap in demonstrated capabilities and current limitations. Long-horizon evaluation is needed to establish whether these mechanisms deliver transferable gains under explicit resource constraints and human oversight. Ultimately, the promise of RSI lies in enabling each generation of AI to make future improvement more reliable, efficient, and conducive to novel discoveries.

\clearpage
\bibliographystyle{IEEEtran}
\bibliography{reference}

\clearpage
\appendix

\phantomsection
\addcontentsline{toc}{section}{Appendix}

\newcommand{\appsection}[1]{%
    \refstepcounter{section}%
    \section*{Appendix \thesection: #1}%
    \addcontentsline{toc}{subsection}{%
        \protect\numberline{\thesection}#1%
    }%
}

\pagestyle{appendixAstyle}

\appsection{RSI Landscape}
\label{app:rsi-landscape}

\begin{figure}[!htbp]
    \centering
    \includegraphics[
        width=\linewidth,
        height=0.72\textheight,
        keepaspectratio
    ]{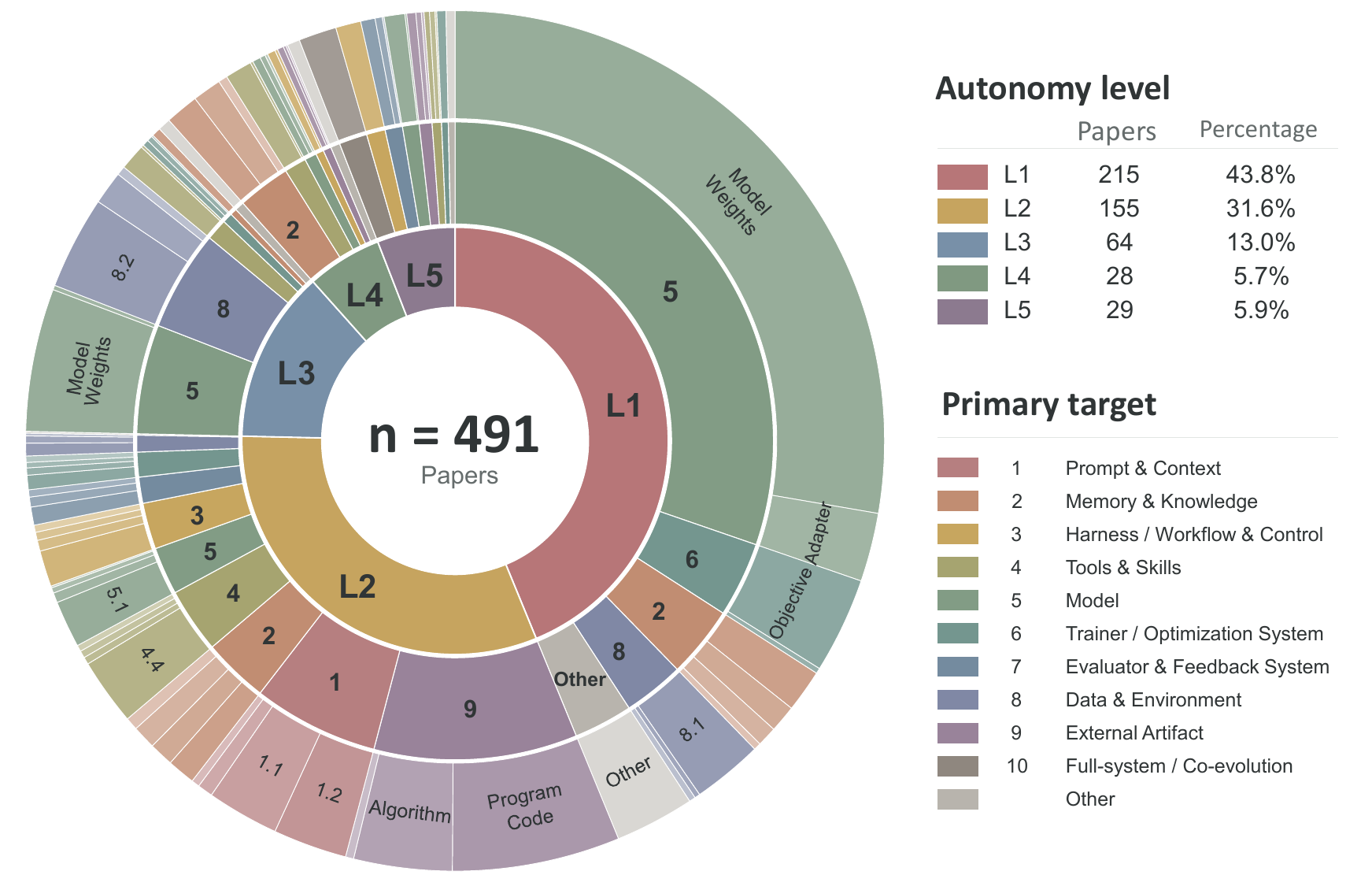}
    \caption{
        \textbf{RSI Landscape: Autonomy Levels and Improvement Targets.}
        Distribution of 491 surveyed papers.
        The inner, middle, and outer rings represent autonomy
        levels (L1--L5), primary improvement targets, and
        sub-targets, respectively.
        Autonomy-level percentages are based on paper counts.
        Papers associated with multiple improvement targets
        contribute fractional weights to the target categories.
    }
    \label{fig:rsi-landscape}
\end{figure}

Figure~\ref{fig:rsi-landscape} summarizes the surveyed papers by autonomy level and improvement target.

Table~\ref{tab:rsi-target-taxonomy} details the primary targets and sub-targets represented in Figure~\ref{fig:rsi-landscape}.

\captionsetup[longtable]{
    justification=centering,
    singlelinecheck=false
}
\begingroup
\small   
\renewcommand{\arraystretch}{1.08}
\setlength{\tabcolsep}{4pt}

\begin{longtable}{
    >{\RaggedRight\arraybackslash}m{0.18\linewidth}
    >{\RaggedRight\arraybackslash}m{0.27\linewidth}
    >{\RaggedRight\arraybackslash}m{0.49\linewidth}
}

\caption{
\textbf{Taxonomy of Improvement Targets in Recursive Self-Improvement Systems.}
The taxonomy decomposes the improvement targets represented in Fig.~\ref{fig:rsi-landscape} into primary targets and sub-targets.
For each sub-target, the table specifies the concrete system component subject to modification.
}
\label{tab:rsi-target-taxonomy}\\

\toprule
\textbf{Primary Target}
& \textbf{Sub-target}
& \textbf{Modified Component}
\\
\midrule
\endfirsthead

\multicolumn{3}{c}{\textit{Table~\ref{tab:rsi-target-taxonomy} continued}}\\
\toprule
\textbf{Primary Target}
& \textbf{Sub-target}
& \textbf{Modified Component}
\\
\midrule
\endhead

\midrule
\multicolumn{3}{r}{\textit{Continued on next page}}\\
\endfoot

\bottomrule
\endlastfoot

1. Prompt \& Context
& \multicolumn{2}{@{}l@{}}{%
\begin{tabular}{
    >{\RaggedRight\arraybackslash}m{0.27\linewidth}
    >{\RaggedRight\arraybackslash}m{0.49\linewidth}
}
1.1 Instruction
& System prompts, role specifications, behavioral rules, constraints, and high-level instructions \\
\subsep
1.2 Task Prompt / Template
& Task descriptions, problem formulations, prompt templates, and reusable task specifications \\
\subsep
1.3 Reasoning Prompt / Protocol
& Explicit reasoning instructions and protocols, including chain-of-thought, reflection, critique, and debate \\
\subsep
1.4 Demonstrations / Exemplars
& Few-shot examples, demonstrations, successful trajectories, and worked solutions included in context \\
\subsep
1.5 Context Composition
& Context selection, ordering, compression, summarization, and allocation of the available context budget \\
\end{tabular}%
}\\
\midrule

2. Memory \& Knowledge
& \multicolumn{2}{@{}l@{}}{%
\begin{tabular}{
    >{\RaggedRight\arraybackslash}m{0.27\linewidth}
    >{\RaggedRight\arraybackslash}m{0.49\linewidth}
}
2.1 Episodic / Experience Memory
& Records of prior interactions, execution trajectories, successful and failed attempts, and task-specific reflections \\
\subsep
2.2 Semantic / Knowledge Memory
& Persistent factual, conceptual, and domain knowledge, including structured or external knowledge stores \\
\subsep
2.3 Procedural Memory
& Reusable strategies, experience-derived rules, heuristics, procedures, and playbooks \\
\subsep
2.4 Memory Representation / Organization
& Memory schemas, hierarchical structures, summaries, graphs, indexes, and vector representations \\
\subsep
2.5 Memory Operations
& Policies governing memory writing, retrieval, updating, consolidation, prioritization, and forgetting \\
\end{tabular}%
}\\
\midrule

3. Harness / Workflow \& Control
& \multicolumn{2}{@{}l@{}}{%
\begin{tabular}{
    >{\RaggedRight\arraybackslash}m{0.27\linewidth}
    >{\RaggedRight\arraybackslash}m{0.49\linewidth}
}
3.1 Workflow / Graph
& Agent graphs, pipelines, nodes, edges, and execution dependencies between computational stages \\
\subsep
3.2 Planning / Decomposition
& Task decomposition, planning modules, subgoal structures, and execution order \\
\subsep
3.3 Routing / Scheduling
& Model routing, agent routing, tool routing, execution scheduling, and computational resource allocation \\
\subsep
3.4 Verification / Reflection Loop
& Control loops for checking and revising intermediate or final outputs, including critic--revise, verify--retry, and reflection procedures \\
\subsep
3.5 Multi-agent Structure / Protocol
& Agent population, role assignment, interaction topology, coordination mechanisms, and communication protocols \\
\subsep
3.6 Harness Implementation / Scaffold Code
& Source code implementing the agent scaffold, orchestration logic, workflow controller, and runtime coordination mechanisms \\
\end{tabular}%
}\\
\midrule

4. Tools \& Skills
& \multicolumn{2}{@{}l@{}}{%
\begin{tabular}{
    >{\RaggedRight\arraybackslash}m{0.27\linewidth}
    >{\RaggedRight\arraybackslash}m{0.49\linewidth}
}
4.1 Tool Set / Inventory
& The collection of tools, APIs, external services, and callable capabilities available to the system \\
\subsep
4.2 Tool Definition / Interface
& Tool descriptions, function signatures, schemas, API wrappers, and input--output specifications \\
\subsep
4.3 Tool Implementation
& Executable code and internal logic implementing tools or callable external functions \\
\subsep
4.4 Skill / Macro Library
& Reusable skills, macros, subroutines, procedures, and higher-level behavioral modules \\
\subsep
4.5 Skill Composition
& Rules and structures for combining lower-level skills into higher-level procedures or capabilities \\
\end{tabular}%
}\\
\midrule

5. Model
& \multicolumn{2}{@{}l@{}}{%
\begin{tabular}{
    >{\RaggedRight\arraybackslash}m{0.27\linewidth}
    >{\RaggedRight\arraybackslash}m{0.49\linewidth}
}
5.1 Model Weights
& Trainable parameters of the backbone model \\
\subsep
5.2 Adapter / Auxiliary Module
& Parameters of LoRA modules, adapters, memory modules, and other trainable auxiliary components \\
\subsep
5.3 Architecture
& Network architecture, module composition, connectivity patterns, and computational structure \\
\subsep
5.4 Inference Policy / Configuration
& Persistent decoding strategies, test-time policies, inference configurations, and runtime model settings \\
\end{tabular}%
}\\
\midrule

6. Trainer / Optimization System
& \multicolumn{2}{@{}l@{}}{%
\begin{tabular}{
    >{\RaggedRight\arraybackslash}m{0.27\linewidth}
    >{\RaggedRight\arraybackslash}m{0.49\linewidth}
}
6.1 Training Objective
& Loss functions, training rewards, learning objectives, and optimization criteria \\
\subsep
6.2 Optimizer / Update Rule
& Optimizers, parameter-update algorithms, learning-rate policies, and associated hyperparameters \\
\subsep
6.3 Training Schedule / Pipeline
& Ordering, configuration, and interaction of training stages such as fine-tuning, reinforcement learning, and distillation \\
\subsep
6.4 Data Selection / Curriculum
& Training-sample selection rules, data mixtures, difficulty schedules, and curriculum policies \\
\subsep
6.5 Search / Meta-optimization Procedure
& Evolutionary, search, selection, and meta-optimization procedures used to generate and select candidate improvements \\
\end{tabular}%
}\\
\midrule

7. Evaluator \& Feedback System
& \multicolumn{2}{@{}l@{}}{%
\begin{tabular}{
    >{\RaggedRight\arraybackslash}m{0.27\linewidth}
    >{\RaggedRight\arraybackslash}m{0.49\linewidth}
}
7.1 Evaluator / Judge
& Evaluators, critics, graders, judge models, and related components for assessing candidate outputs or system variants \\
\subsep
7.2 Reward / Fitness Function
& Reward models, fitness functions, utility functions, preference models, and scoring rules \\
\subsep
7.3 Verifier
& Correctness verifiers, test generators, proof checkers, consistency checks, and validation mechanisms \\
\subsep
7.4 Feedback / Credit Assignment
& Mechanisms that transform observed outcomes into localized, aggregated, or temporally assigned improvement signals \\
\end{tabular}%
}\\
\midrule

8. Data \& Environment
& \multicolumn{2}{@{}l@{}}{%
\begin{tabular}{
    >{\RaggedRight\arraybackslash}m{0.27\linewidth}
    >{\RaggedRight\arraybackslash}m{0.49\linewidth}
}
8.1 Training / Experience Data
& Training datasets, replay buffers, experience pools, interaction trajectories, and other data used for subsequent learning \\
\subsep
8.2 Task / Curriculum Generator
& Components that generate tasks, problems, challenges, or training episodes for subsequent improvement cycles \\
\subsep
8.3 Environment / Simulator
& Environment dynamics, simulators, interaction rules, task worlds, and structures governing system--environment interaction \\
\subsep
8.4 World Model
& Learned models used to represent, predict, generate, or simulate environmental states and dynamics \\
\end{tabular}%
}\\
\midrule

9. External Artifact
& \multicolumn{2}{@{}l@{}}{%
\begin{tabular}{
    >{\RaggedRight\arraybackslash}m{0.27\linewidth}
    >{\RaggedRight\arraybackslash}m{0.49\linewidth}
}
9.1 Program / Solution Code
& Task-level programs, patches, or solution code produced by the system, excluding code implementing the agent itself \\
\subsep
9.2 Algorithm
& Algorithms, kernels, mathematical procedures, symbolic methods, and computational techniques \\
\subsep
9.3 Scientific / Design Artifact
& Scientific hypotheses, experimental designs, architectures, circuits, engineering designs, and related research artifacts \\
\end{tabular}%
}\\
\midrule

10. Full-system / Co-evolution
& \multicolumn{2}{@{}l@{}}{%
\begin{tabular}{
    >{\RaggedRight\arraybackslash}m{0.27\linewidth}
    >{\RaggedRight\arraybackslash}m{0.49\linewidth}
}
10.1 Multi-target Harness
& Multiple harness-level components jointly modified within the same improvement process, such as prompts, memory, workflows, and tools \\
\subsep
10.2 Model--Harness Co-evolution
& Model parameters and surrounding harness or scaffold components jointly modified across improvement cycles \\
\subsep
10.3 Improvement-loop / Meta-RSI
& The mechanism that generates, evaluates, selects, and applies candidate changes across successive improvement cycles \\
\end{tabular}%
}\\

\end{longtable}
\endgroup




\begin{landscape}

\pagestyle{appendixBstyle}

\appsection{Industry Landscape}
\label{app:industry-landscape}

Table~\ref{tab:industry-landscape} organizes the surveyed industrial systems by company archetype and improvement target.

\newcolumntype{P}[1]{>{\RaggedRight\arraybackslash\hspace{0pt}}p{#1}}
\newcolumntype{I}[1]{>{\Centering\arraybackslash\hspace{0pt}}p{#1}}
\newcommand{\src}[1]{\href{#1}{\textcolor{blue!55!black}{\faGlobe}}}
\newcommand{\sectionrow}[2]{%
  \multicolumn{7}{@{}p{\dimexpr\linewidth-2\tabcolsep\relax}@{}}{%
    \cellcolor{black!7}\strut\textbf{#1}\hspace{0.8em}\textcolor{black!65}{\scriptsize #2}}\\[-0.1ex]
}
\newcommand{\innerline}{\arrayrulecolor{black!13}\cline{2-7}\arrayrulecolor{black}}
\setlength{\emergencystretch}{2em}
\urlstyle{same}
\raggedbottom
\pagestyle{plain}
\begin{center}
{\large\bfseries Company Archetypes in the Emerging RSI / Self-Improvement Landscape}\\[2pt]
{\footnotesize One row per product or representative work; 72 distinct companies/teams; public-source snapshot: September 2026}
\end{center}
\vspace{-0.35em}
\noindent
{\scriptsize\textit{Organization principle: company archetype, not geography. The RSI tag is a compact mapping to the B0--L5 scheme and is not a claim that the company itself uses that label. \faGlobe\ = clickable primary source.}}
\vspace{0.35em}
\begingroup
\fontsize{7.15}{8.20}\selectfont
\setlength{\tabcolsep}{2.0pt}
\renewcommand{\arraystretch}{1.02}
\begin{longtable}{
  P{2.72cm}
  P{4.69cm}
  P{3.10cm}
  P{4.15cm}
  P{4.45cm}
  P{2.06cm}
  I{0.78cm}}
\caption{RSI / self-improvement company landscape organized by company archetype.}\label{tab:industry-landscape}\\
\toprule
\textbf{Company} &
\textbf{Product / representative work} &
\textbf{Sub-scenario} &
\textbf{Improvement target / artifact} &
\textbf{AI-controlled part} &
\textbf{RSI relation} &
\textcolor{blue!55!black}{\faGlobe} \\
\midrule
\endfirsthead
\multicolumn{7}{c}{\tablename\ \thetable\ -- continued}\\
\toprule
\textbf{Company} &
\textbf{Product / representative work} &
\textbf{Sub-scenario} &
\textbf{Improvement target / artifact} &
\textbf{AI-controlled part} &
\textbf{RSI relation} &
\textcolor{blue!55!black}{\faGlobe} \\
\midrule
\endhead
\midrule
\multicolumn{7}{r}{\footnotesize continued on next page}\\
\endfoot
\bottomrule
\endlastfoot
\sectionrow{A. Frontier foundation-model \& general-agent labs}{Broad model/platform labs; RSI appears as one capability frontier rather than the sole company thesis.}
\multirow[t]{7}{2.72cm}{\RaggedRight\textbf{OpenAI}} & Research acceleration / automated research intern & AI-for-AI research & research code; experiments; evals & code; experiments; candidate integration & L2 & \src{https://openai.com/index/research-acceleration-view-inside-openai/} \\*
\innerline
 & GPT-Red & self-play robustness & red-team policy; adversarial data & attack; defend; train; evaluate & L2 & \src{https://openai.com/index/unlocking-self-improvement-gpt-red/} \\*
\innerline
 & Self-improving tax agents & production adaptation & agent code; prompts; eval set & mine failures; patch; evaluate & L4 cand. & \src{https://openai.com/index/building-self-improving-tax-agents-with-codex/} \\*
\innerline
 & Harness Engineering & agent-first software R\&D & repo; CI; agent instructions & code; test; PR; repair & L1-L2 adj. & \src{https://openai.com/index/harness-engineering/} \\*
\innerline
 & Symphony & agent orchestration & task state; repo; workflows & schedule; execute; handoff & L1 adj. & \src{https://github.com/openai/symphony} \\*
\innerline
 & AgentKit / prompt optimizer / RFT & agent optimization stack & prompts; graders; weights & optimize; grade; fine-tune & L1-L2 & \src{https://openai.com/index/introducing-agentkit/} \\*
\innerline
 & Deep Research & autonomous research & task-local evidence & search; browse; synthesize & B0 & \src{https://openai.com/index/introducing-deep-research/} \\
\addlinespace[0.85pt]
\multirow[t]{9}{2.72cm}{\RaggedRight\textbf{Anthropic}} & When AI builds itself & RSI roadmap & AI-development process & code; experiments; future successor design & L5 target & \src{https://www.anthropic.com/institute/recursive-self-improvement} \\*
\innerline
 & Automated Weak-to-Strong Researcher & automated alignment research & hypotheses; code; experiment logs & propose; train; evaluate; share & L2 & \src{https://alignment.anthropic.com/2026/automated-w2s-researcher/} \\*
\innerline
 & Tool optimization with Claude & tool self-optimization & tool specs; implementations & analyze traces; rewrite; evaluate & L2 & \src{https://www.anthropic.com/engineering/writing-tools-for-agents} \\*
\innerline
 & Harness design for long-running apps & autonomous software R\&D & harness; evaluator; app code & plan; generate; evaluate; iterate & L2 adj. & \src{https://www.anthropic.com/engineering/harness-design-long-running-apps} \\*
\innerline
 & Effective long-running agent harnesses & cross-context persistence & progress files; git state & initialize; code; handoff & L1 adj. & \src{https://www.anthropic.com/engineering/effective-harnesses-for-long-running-agents} \\*
\innerline
 & Agent Skills & persistent skill substrate & skills; scripts; resources & discover; load; reuse & L1 enabler & \src{https://www.anthropic.com/engineering/equipping-agents-for-the-real-world-with-agent-skills} \\*
\innerline
 & Multi-agent Research & autonomous research & task-local findings & plan; spawn; search; synthesize & B0 & \src{https://www.anthropic.com/engineering/multi-agent-research-system} \\*
\innerline
 & Managed Agents & long-horizon agent infrastructure & stable interface; harness & run; resume; supervise & L1 enabler & \src{https://www.anthropic.com/engineering/managed-agents} \\*
\innerline
 & Parallel Claude compiler project & autonomous software engineering & shared codebase; tests & decompose; code; test; coordinate & B0-L1 adj. & \src{https://www.anthropic.com/engineering/building-c-compiler} \\
\addlinespace[0.85pt]
\multirow[t]{3}{2.72cm}{\RaggedRight\textbf{Google DeepMind}} & AlphaEvolve & task-specific program search & algorithms; kernels; system code & generate; mutate; evaluate; select & B0 & \src{https://deepmind.google/blog/alphaevolve-a-gemini-powered-coding-agent-for-designing-advanced-algorithms/} \\*
\innerline
 & AI Co-Scientist & scientific hypothesis search & hypotheses; research proposals & generate; debate; rank; refine & L2 adj. & \src{https://research.google/blog/accelerating-scientific-breakthroughs-with-an-ai-co-scientist/} \\*
\innerline
 & AlphaChip & AI-for-hardware co-design & chip layouts; design policy & place; score; learn; transfer & L2 adj. & \src{https://deepmind.google/blog/how-alphachip-transformed-computer-chip-design/} \\
\addlinespace[0.85pt]
\multirow[t]{2}{2.72cm}{\RaggedRight\textbf{NVIDIA}} & Eureka & reward / policy design & reward code; robot policies & reward synthesis; simulation; policy training & L2 & \src{https://arxiv.org/abs/2310.12931} \\*
\innerline
 & ASPIRE & reward / policy design & reward code; robot policies & reward synthesis; simulation; policy training & L2 & \src{https://research.nvidia.com/labs/gear/aspire/} \\
\addlinespace[0.85pt]
\multirow[t]{5}{2.72cm}{\RaggedRight\textbf{Microsoft}} & Agent Lightning & agent reinforcement learning & policy weights; experience traces & collect; credit; train; evaluate & L2 & \src{https://www.microsoft.com/en-us/research/project/agent-lightning/} \\*
\innerline
 & Agent Lightning v1.0 & harnessed agentic RL & harness traces; policy weights & interact; retokenize; train; benchmark & L2 & \src{https://www.microsoft.com/en-us/research/publication/agent-lightning-v1-0-towards-harnessed-agentic-rl/} \\*
\innerline
 & SkillOpt & skill optimization & skill files; instructions & edit; evaluate; optimize & L2 & \src{https://www.microsoft.com/en-us/research/blog/skillopt-agent-skills-as-trainable-parameters/} \\*
\innerline
 & ReVeal & self-verifying code agents & code; tests; verifier policy & generate; verify; revise; scale & L2 & \src{https://www.microsoft.com/en-us/research/publication/reveal-self-evolving-code-agents-via-reliable-self-verification/} \\*
\innerline
 & Universal Verifier / auto-research & agent verification R\&D & rubrics; verifier; eval pipeline & design; test; compare; refine & L2 adj. & \src{https://www.microsoft.com/en-us/research/articles/the-art-of-building-verifiers-for-computer-use-agents/} \\
\addlinespace[0.85pt]
\multirow[t]{2}{2.72cm}{\RaggedRight\textbf{Meta}} & Self-Taught Evaluator & evaluator self-training & judge model; synthetic preferences & generate; judge; train; iterate & L2 & \src{https://ai.meta.com/blog/fair-news-segment-anything-2-1-meta-spirit-lm-layer-skip-salsa-lingua/} \\*
\innerline
 & HyperAgents & meta-agent self-modification & task agent; meta agent; program & solve; self-edit; evaluate; archive & L5 cand. & \src{https://ai.meta.com/research/publications/hyperagents/} \\
\addlinespace[0.85pt]
\multirow[t]{3}{2.72cm}{\RaggedRight\textbf{Alibaba / Qwen}} & Qwen-Agent & agent training / synthetic environments & training data; environments; post-training & simulation; data synthesis; tool use; RL & L1-L2 & \src{https://qwenlm.github.io/blog/qwen-agent-2405/} \\*
\innerline
 & AgentWorld & agent training / synthetic environments & training data; environments; post-training & simulation; data synthesis; tool use; RL & L1-L2 & \src{https://qwen.ai/blog?id=qwen-agentworld} \\*
\innerline
 & Qwen-Scope & agent training / synthetic environments & training data; environments; post-training & simulation; data synthesis; tool use; RL & L1-L2 & \src{https://arxiv.org/abs/2605.11887} \\
\addlinespace[0.85pt]
\multirow[t]{2}{2.72cm}{\RaggedRight\textbf{DeepSeek}} & R1 & reasoning self-bootstrapping & reasoning policy; verifier & self-generated reasoning; RL; verifier iteration & L2 & \src{https://github.com/deepseek-ai/DeepSeek-R1} \\*
\innerline
 & Math-V2 & reasoning self-bootstrapping & reasoning policy; verifier & self-generated reasoning; RL; verifier iteration & L2 & \src{https://github.com/deepseek-ai/DeepSeek-Math-V2} \\
\addlinespace[0.85pt]
\RaggedRight\textbf{Tencent AI Lab} & R-Zero & self-play reasoning & tasks; pseudo-labels; policy weights & challenge generation; solve; vote; RL & L2-L3 & \src{https://arxiv.org/abs/2508.05004} \\
\addlinespace[0.85pt]
\multirow[t]{2}{2.72cm}{\RaggedRight\textbf{ByteDance Seed}} & Seed-Thinking & model / agent post-training & data; reward; policy weights & data filtering; reward verification; RL & L1-L2 & \src{https://github.com/ByteDance-Seed} \\*
\innerline
 & Seed1.5-VL & model / agent post-training & data; reward; policy weights & data filtering; reward verification; RL & L1-L2 & \src{https://seed.bytedance.com/zh/blog/bytedance-s-latest-thinking-model-seed-thinking-v1-5-technical-details-disclosed} \\
\addlinespace[0.85pt]
\multirow[t]{3}{2.72cm}{\RaggedRight\textbf{MiniMax}} & M2 & persistent cloud agents & memory; skills; agent policy & tool use; long-run execution; skill reuse & L1-L2 & \src{https://www.minimax.io/news/minimax-m2} \\*
\innerline
 & MaxHermes & persistent cloud agents & memory; skills; agent policy & tool use; long-run execution; skill reuse & L1-L2 & \src{https://www.maxhermes.dev/en} \\*
\innerline
 & MaxClaw & persistent cloud agents & memory; skills; agent policy & tool use; long-run execution; skill reuse & L1-L2 & \src{https://agent.minimaxi.com/activity/max-claw} \\
\addlinespace[0.85pt]
\multirow[t]{2}{2.72cm}{\RaggedRight\textbf{Moonshot AI}} & Kimi K3 & long-horizon agents & agent orchestration; tool policy & planning; tool use; swarm coordination & L1-L2 adj. & \src{https://www.moonshot.cn/} \\*
\innerline
 & Agent Swarm & long-horizon agents & agent orchestration; tool policy & planning; tool use; swarm coordination & L1-L2 adj. & \src{https://www.moonshot.cn/} \\
\addlinespace[0.85pt]
\multirow[t]{2}{2.72cm}{\RaggedRight\textbf{Zhipu AI / Z.AI}} & AutoGLM & computer-use agent training & policy weights; virtual-phone environments & perception; planning; action; RL & L2 & \src{https://autoglm.z.ai/blog/} \\*
\innerline
 & AgentRL & computer-use agent training & policy weights; virtual-phone environments & perception; planning; action; RL & L2 & \src{https://docs.z.ai/guides/vlm/autoglm-phone-multilingual} \\
\addlinespace[0.85pt]
\multirow[t]{2}{2.72cm}{\RaggedRight\textbf{Deep Cogito}} & Cogito v2 & reasoning post-training & reasoning policy; weights & search; distill; iterative alignment & L1-L2 & \src{https://www.deepcogito.com/research/cogito-v2-preview} \\*
\innerline
 & IDA & reasoning post-training & reasoning policy; weights & search; distill; iterative alignment & L1-L2 & \src{https://www.deepcogito.com/research/cogito-v2-1} \\
\addlinespace[0.85pt]
\RaggedRight\textbf{Poolside} & Model Factory & automated model R\&D & synthetic data; RL; architecture & eval; code-exec RL; ablations; data mix & L1-L2 & \src{https://poolside.ai/research} \\
\addlinespace[0.85pt]
\multirow[t]{2}{2.72cm}{\RaggedRight\textbf{Thinking Machines Lab}} & Tinker Agent RL & model customization / agent RL & policy weights; tool-use policy & RL; LLM-judge grading; tool discovery & L1-L2 & \src{https://tinker-docs.thinkingmachines.ai/cookbook/recipes/agent-rl/} \\*
\innerline
 & Inkling & model customization / agent RL & policy weights; tool-use policy & RL; LLM-judge grading; tool discovery & L1-L2 & \src{https://thinkingmachines.ai/news/introducing-inkling/} \\
\addlinespace[0.85pt]
\multirow[t]{2}{2.72cm}{\RaggedRight\textbf{Nous Research}} & Hermes Agent & persistent agents & skills; memory; tool gateway & memory; skill reuse; tool orchestration & L1-L2 & \src{https://github.com/NousResearch/hermes-agent} \\*
\innerline
 & skills / memory & persistent agents & skills; memory; tool gateway & memory; skill reuse; tool orchestration & L1-L2 & \src{https://hermes-agent.nousresearch.com/docs} \\
\addlinespace[0.85pt]
\sectionrow{B. RSI-native / AI4AI companies}{Self-improvement, AI-for-AI, self-evolution, or recursive improvement is central to the company/research thesis.}
\RaggedRight\textbf{Ricursive Intelligence} & AI-chip co-design & AI systems / chips & EDA designs; compute stack & design search; verify; iterate & L2; L5 vision & \src{https://www.ricursive.com/} \\
\addlinespace[0.85pt]
\RaggedRight\textbf{Recursive} & Automated AI Research & model-training research & training recipes; kernels; code & ideas; experiments; branch merge & L2 & \src{https://www.recursive.com/articles/first-steps-toward-automated-ai-research} \\
\addlinespace[0.85pt]
\multirow[t]{2}{2.72cm}{\RaggedRight\textbf{Imbue}} & Catalyst & research search & recipes; code; hypotheses & population search; experiments; interpretation & L2-L3 & \src{https://github.com/imbue-ai/catalyst} \\*
\innerline
 & Darwinian Evolver & research search & recipes; code; hypotheses & population search; experiments; interpretation & L2-L3 & \src{https://imbue.com/blog/2026-07-20-imbue-catalyst-nanochat} \\
\addlinespace[0.85pt]
\RaggedRight\textbf{Weco AI} & AIDE2 & meta-improvement of researcher & research harness; improver code & rewrite improver; eval; inheritance & L5 cand. & \src{https://www.weco.ai/blog/first-evidence-of-recursive-self-improvement} \\
\addlinespace[0.85pt]
\multirow[t]{2}{2.72cm}{\RaggedRight\textbf{Sakana AI}} & Darwin Godel Machine & self-editing agents / AI research & agent source; research pipeline & self-modify; benchmark; archive; experiments & L5 cand. & \src{https://arxiv.org/abs/2505.22954} \\*
\innerline
 & AI Scientist & self-editing agents / AI research & agent source; research pipeline & self-modify; benchmark; archive; experiments & L5 cand. & \src{https://github.com/SakanaAI/AI-Scientist} \\
\addlinespace[0.85pt]
\multirow[t]{3}{2.72cm}{\RaggedRight\textbf{Evolvent AI}} & Org self-evolving agents & software-agent evolution & skills; memory; code; environments & tasks; feedback; refactor; skill updates & L2-L3 & \src{https://evolvent.co/zh/blog/org-self-evolving-agents} \\*
\innerline
 & RSIBench & software-agent evolution & skills; memory; code; environments & tasks; feedback; refactor; skill updates & L2-L3 & \src{https://evolvent.co/zh/blog/RSIBench-Data} \\*
\innerline
 & Terrarium & software-agent evolution & skills; memory; code; environments & tasks; feedback; refactor; skill updates & L2-L3 & \src{https://evolvent.co/en/blog/terrarium} \\
\addlinespace[0.85pt]
\multirow[t]{2}{2.72cm}{\RaggedRight\textbf{MetaCircle}} & ComfyResearch & AI4AI / autoresearch & training workflows; research hypotheses & experiment compose; code; hypothesis; scoring & L2; L5 vision & \src{https://meta-circle.com/blog/comfyresearch-see-how-learning-happens} \\*
\innerline
 & OPHIS & AI4AI / autoresearch & training workflows; research hypotheses & experiment compose; code; hypothesis; scoring & L2; L5 vision & \src{https://meta-circle.com/blog/ophis-a-new-paradigm-for-autoresearch} \\
\addlinespace[0.85pt]
\multirow[t]{2}{2.72cm}{\RaggedRight\textbf{Frontis AI / Xianyuan}} & OpenRSI & AI4AI / self-improving agents & skills; memory; harness; weights & experience; update search; eval; post-training & L2-L3 & \src{https://github.com/FrontisAI/OpenRSI} \\*
\innerline
 & Frontis-MA1 & AI4AI / self-improving agents & skills; memory; harness; weights & experience; update search; eval; post-training & L2-L3 & \src{https://arxiv.org/abs/2607.28568} \\
\addlinespace[0.85pt]
\multirow[t]{3}{2.72cm}{\RaggedRight\textbf{EvoMap}} & Evolver & shared code evolution & genes / capsules; code assets & generate; test; publish; reuse; adapt & L2-L3 & \src{https://github.com/EvoMap/evolver} \\*
\innerline
 & GEP & shared code evolution & genes / capsules; code assets & generate; test; publish; reuse; adapt & L2-L3 & \src{https://evomap.ai/wiki/16-gep-protocol} \\*
\innerline
 & GeneBench & shared code evolution & genes / capsules; code assets & generate; test; publish; reuse; adapt & L2-L3 & \src{https://evomap.ai/wiki/36-gene-bench-report} \\
\addlinespace[0.85pt]
\RaggedRight\textbf{Endless Frontier} & BigBang-v1 & AI-research data generation & synthetic programs; training data; critic & generate; execute; critique; meta-critique & L2-L3 & \src{https://github.com/endless-frontier/BigBang-v1} \\
\addlinespace[0.85pt]
\multirow[t]{2}{2.72cm}{\RaggedRight\textbf{Mirendil}} & Self-accelerating AI & AI R\&D automation & model code; experiments; research stack & experiment generation; execution; iteration & L2 & \src{https://mirendil.com/news/scaling-self-accelerating-ai-with-google/} \\*
\innerline
 & AI Scientist & AI R\&D automation & model code; experiments; research stack & experiment generation; execution; iteration & L2 & \src{https://www.devvrit.com/} \\
\addlinespace[0.85pt]
\multirow[t]{3}{2.72cm}{\RaggedRight\textbf{Chaoyan Intelligence*}} & TUMIX & self-evolving research models & research policy; tool-use policy & questioning; experiments; code; self-check & L2-L3 & \src{https://arxiv.org/abs/2510.01279} \\*
\innerline
 & R1-Code-Interpreter & self-evolving research models & research policy; tool-use policy & questioning; experiments; code; self-check & L2-L3 & \src{https://arxiv.org/abs/2505.21668} \\*
\innerline
 & AI Scientist & self-evolving research models & research policy; tool-use policy & questioning; experiments; code; self-check & L2-L3 & \src{https://www.qbitai.com/2026/07/455041.html} \\
\addlinespace[0.85pt]
\multirow[t]{2}{2.72cm}{\RaggedRight\textbf{Theseus}} & Argus & long-horizon self-evolving agents & experience; memory; environment setup & plan; code; review; experience writeback & L2-L3 & \src{https://arxiv.org/abs/2608.05144} \\*
\innerline
 & SetupX & long-horizon self-evolving agents & experience; memory; environment setup & plan; code; review; experience writeback & L3 & \src{https://arxiv.org/abs/2605.26186} \\
\addlinespace[0.85pt]
\multirow[t]{2}{2.72cm}{\RaggedRight\textbf{Adaption Labs}} & AutoScientist & AI-research automation & training recipes; datasets; code & research plan; experiments; selection & L2 & \src{https://docs.adaptionlabs.ai/autoscientist/overview/} \\*
\innerline
 & Forge & AI-research automation & training recipes; datasets; code & research plan; experiments; selection & L2 & \src{https://docs.adaptionlabs.ai/api/resources/autoscientist} \\
\addlinespace[0.85pt]
\sectionrow{C. Autonomous R\&D \& scientific-discovery companies}{Primary product is automated research/discovery; RSI relevance comes from automating experiment and knowledge-production loops.}
\RaggedRight\textbf{Periodic Labs} & Autonomous laboratory & AI for science & hypotheses; experiment data; models & experiment design; lab run; learning & L3 cand. & \src{https://periodic.com/} \\
\addlinespace[0.85pt]
\RaggedRight\textbf{FutureHouse} & BixBench & AI-scientist evaluation & research tasks; benchmark frontier & bioinformatics workflows; open-ended eval & B0-L2 adj. & \src{https://github.com/Future-House/BixBench} \\
\addlinespace[0.85pt]
\RaggedRight\textbf{Edison Scientific} & Kosmos & autonomous science & hypotheses; code; scientific artifacts & literature; experiments; synthesis & L2-L3 & \src{https://arxiv.org/abs/2511.02824} \\
\addlinespace[0.85pt]
\multirow[t]{2}{2.72cm}{\RaggedRight\textbf{Axiom Math}} & Putnam 2025 & formal mathematics & proofs; verifier traces & conjecture; proof search; formal verification & L2 & \src{https://github.com/AxiomMath/putnam2025} \\*
\innerline
 & IMO 2026 & formal mathematics & proofs; verifier traces & conjecture; proof search; formal verification & L2 & \src{https://github.com/AxiomMath/IMO2026} \\
\addlinespace[0.85pt]
\RaggedRight\textbf{Harmonic} & Aristotle & theorem proving & formal proofs & translate; prove; verify & L1-L2 & \src{https://harmonic.fun/pdf/Aristotle_IMO_Level_Automated_Theorem_Proving.pdf} \\
\addlinespace[0.85pt]
\RaggedRight\textbf{Core Automation} & AI systems-research stack & automated systems research & systems code; research hypotheses & design; code; benchmark; iterate & L2 cand. & \src{https://www.coreauto.com/blog} \\
\addlinespace[0.85pt]
\RaggedRight\textbf{Discovery Loop} & Autonomous discovery loop & automated experimentation & protocols; findings & hypothesis; experiment; analysis; next-step & L3 cand. & \src{https://www.discoveryloop.com/} \\
\addlinespace[0.85pt]
\RaggedRight\textbf{Lila Sciences} & Autonomous Science platform & scientific discovery & hypotheses; experiments; post-training data & design; lab execution; real-time learning & L3 cand. & \src{https://www.lila.ai/} \\
\addlinespace[0.85pt]
\RaggedRight\textbf{Karpathy / autoresearch} & autoresearch & narrow automated research & training code; configs & edit; train; measure; keep & L2 & \src{https://github.com/karpathy/autoresearch} \\
\addlinespace[0.85pt]
\multirow[t]{2}{2.72cm}{\RaggedRight\textbf{Prime Intellect}} & Autonomous research & model R\&D & training recipe; optimizer; code & hypothesis; GPU runs; ablation & L2 & \src{https://www.primeintellect.ai/blog/measuring-autonomous-research} \\*
\innerline
 & Speedrun Frontier & model R\&D & training recipe; optimizer; code & hypothesis; GPU runs; ablation & L2 & \src{https://github.com/PrimeIntellect-ai/experiments-autonomous-speedrunning} \\
\addlinespace[0.85pt]
\RaggedRight\textbf{Analemma} & FARS & autonomous research & proposals; code; logs; papers & topic; experiment; analysis; writing & L2-L3 & \src{https://fars-live.analemma.ai/blog/introducing-fars/} \\
\addlinespace[0.85pt]
\multirow[t]{3}{2.72cm}{\RaggedRight\textbf{Novix}} & AutoAgent & AI research agent & research workflow; eval harness & idea; tool use; experiments; reporting & L2 & \src{https://github.com/HKUDS/AutoAgent} \\*
\innerline
 & OpenHarness & AI research agent & research workflow; eval harness & idea; tool use; experiments; reporting & L2 & \src{https://github.com/HKUDS/OpenHarness} \\*
\innerline
 & AI-Researcher & AI research agent & research workflow; eval harness & idea; tool use; experiments; reporting & L2 & \src{https://github.com/HKUDS/AI-Researcher} \\
\addlinespace[0.85pt]
\multirow[t]{3}{2.72cm}{\RaggedRight\textbf{UniPat}} & UniScientist & research / evaluation agents & research traces; benchmarks & experiments; coding; multi-agent eval & L1-L2 & \src{https://www.unipat.ai/blog/UniScientist} \\*
\innerline
 & UniSwarm & research / evaluation agents & research traces; benchmarks & experiments; coding; multi-agent eval & L1-L2 & \src{https://www.unipat.ai/blog/UniSwarm} \\*
\innerline
 & UniMath & research / evaluation agents & research traces; benchmarks & experiments; coding; multi-agent eval & L1-L2 & \src{https://www.unipat.ai/blog/UniMath} \\
\addlinespace[0.85pt]
\RaggedRight\textbf{Kai Chen / venture TBD*} & Intern-S1 & AI for science models & scientific model; training data & model training; scientific reasoning & adj.; entity TBD & \src{https://air.tsinghua.edu.cn/info/1008/2492.htm} \\
\addlinespace[0.85pt]
\sectionrow{D. Agent optimization, evaluation \& learning infrastructure}{Infrastructure for feedback, skills, RL, evaluation, simulators, training data, or production-agent improvement.}
\RaggedRight\textbf{Warp} & Skill optimization loop & coding-agent skills & skills; instructions; examples & feedback mining; skill rewrite; eval & L2 & \src{https://www.warp.dev/blog/self-improvement-loop-for-skills} \\
\addlinespace[0.85pt]
\multirow[t]{2}{2.72cm}{\RaggedRight\textbf{Factory}} & Signals & production coding agents & agent behavior; product code & session mining; failure clusters; patch PRs & L4 & \src{https://factory.com/news/factory-signals} \\*
\innerline
 & Software Factory & production coding agents & agent behavior; product code & session mining; failure clusters; patch PRs & L4 & \src{https://factory.ai/news/software-factory} \\
\addlinespace[0.85pt]
\RaggedRight\textbf{LangChain} & LangSmith self-improving evaluators & evaluator alignment & judge prompts; evaluator model & feedback ingest; judge update; eval & L1-L2 & \src{https://docs.langchain.com/langsmith/improve-judge-evaluator-feedback} \\
\addlinespace[0.85pt]
\RaggedRight\textbf{Replit} & Agent 3 & software engineering & application code; tests & build; test; repair; long runs & L1-L2 & \src{https://replit.com/blog/introducing-agent-3-our-most-autonomous-agent-yet} \\
\addlinespace[0.85pt]
\RaggedRight\textbf{MorphMind} & Caliper & agent calibration / org memory & agent skills; shared experience & measure; route; reuse experience & L1-L2 adj. & \src{https://morphmind.ai/products/caliper} \\
\addlinespace[0.85pt]
\multirow[t]{2}{2.72cm}{\RaggedRight\textbf{DatologyAI}} & BeyondWeb & data optimization & training corpus; synthetic data & curate; filter; synthesize; benchmark & L1-L2 & \src{https://www.datologyai.com/blog/beyondweb} \\*
\innerline
 & DatBench & data optimization & training corpus; synthetic data & curate; filter; synthesize; benchmark & L1-L2 & \src{https://www.datologyai.com/blog/datbench-discriminative-faithful-and-efficient-vision-language-model-evaluations} \\
\addlinespace[0.85pt]
\RaggedRight\textbf{Ineffable Intelligence} & RL infrastructure & RL / experience infrastructure & training environments; trajectories & rollouts; reward; distributed RL & L1-L3 enabler & \src{https://www.ineffable.ai/} \\
\addlinespace[0.85pt]
\multirow[t]{2}{2.72cm}{\RaggedRight\textbf{Goodfire}} & Ember & interpretability-driven training & feature activations; reward signals & feature discovery; feedback; RL & L1-L2 & \src{https://www.goodfire.com/research/rlfr} \\*
\innerline
 & RLFR & interpretability-driven training & feature activations; reward signals & feature discovery; feedback; RL & L1-L2 & \src{https://www.goodfire.com/research/rlfr} \\
\addlinespace[0.85pt]
\multirow[t]{2}{2.72cm}{\RaggedRight\textbf{Patronus AI}} & Generative Simulators & evaluation / simulation & simulators; eval suites & scenario generation; grading; failure analysis & L2-L3 enabler & \src{https://www.patronus.ai/blog/introducing-generative-simulators} \\*
\innerline
 & Percival & evaluation / simulation & simulators; eval suites & scenario generation; grading; failure analysis & L2-L3 enabler & \src{https://www.patronus.ai/blog/percival-chat-an-eval-copilot-for-agentic-systems} \\
\addlinespace[0.85pt]
\multirow[t]{2}{2.72cm}{\RaggedRight\textbf{Braintrust}} & Loop & production feedback / eval & prompts; evals; datasets & trace ingest; scoring; prompt optimization & L1-L2 & \src{https://www.braintrust.dev/docs/loop} \\*
\innerline
 & Autoevals & production feedback / eval & prompts; evals; datasets & trace ingest; scoring; prompt optimization & L1-L2 & \src{https://www.braintrust.dev/docs/cookbook/recipes/Loop} \\
\addlinespace[0.85pt]
\multirow[t]{2}{2.72cm}{\RaggedRight\textbf{Mechanize}} & RL environments & experience / eval infrastructure & RL tasks; graders; environments & task design; grading; training signal & L3 enabler & \src{https://www.mechanize.work/} \\*
\innerline
 & GBA Eval & experience / eval infrastructure & RL tasks; graders; environments & task design; grading; training signal & L3 enabler & \src{https://www.mechanize.work/} \\
\addlinespace[0.85pt]
\RaggedRight\textbf{Kando AI} & Early company signal & undisclosed / early-stage & undisclosed & undisclosed & unverified & \src{https://www.preqin.com/data/profile/asset/kando-ai/811935} \\
\addlinespace[0.85pt]
\RaggedRight\textbf{Entropy Order*} & Data-expert platform & high-quality training data & expert data; eval assets & data production; benchmarking & L1-L3 enabler & \src{https://emergeia.com/en/project/2025F-020} \\
\addlinespace[0.85pt]
\RaggedRight\textbf{Compounding Intelligence / CORAL*} & CORAL & multi-agent research & shared notes; skills; logs & parallel experiments; knowledge sharing; eval & L2-L3 & \src{https://human-agent-society.github.io/CORAL} \\
\addlinespace[0.85pt]
\RaggedRight\textbf{Naive.ai*} & Research-lab signal & early RSI lab & undisclosed & undisclosed & RSI claim; unverified & \src{https://naive.ai/} \\
\addlinespace[0.85pt]
\sectionrow{E. Embodied, world-model \& continual-adaptation companies}{Persistent adaptation is tied to world models, robotics, environments, or test-time/continual learning.}
\multirow[t]{2}{2.72cm}{\RaggedRight\textbf{AI2 Robotics}} & FiS-VLA & embodied policy learning & VLA policy; action data & data; training; planning; control & L2-L3 adj. & \src{https://arxiv.org/abs/2506.01953} \\*
\innerline
 & Video2Act & embodied policy learning & VLA policy; action data & data; training; planning; control & L2-L3 adj. & \src{https://ai2robotics.com/about/} \\
\addlinespace[0.85pt]
\multirow[t]{2}{2.72cm}{\RaggedRight\textbf{X Square Robot}} & WALL & world models / skill acquisition & world model; skills; action policy & video skill capture; world prediction; control & L2-L3 & \src{https://x2robot.com/research} \\*
\innerline
 & HOST & world models / skill acquisition & world model; skills; action policy & video skill capture; world prediction; control & L2-L3 & \src{https://x2robot.com/pages/host} \\
\addlinespace[0.85pt]
\multirow[t]{2}{2.72cm}{\RaggedRight\textbf{Galaxea AI}} & G0 / G0.5 & VLA R\&D platform & VLA model; data; deployment stack & training; eval; real-robot deployment & L2 adj. & \src{https://arxiv.org/abs/2509.00576} \\*
\innerline
 & GForge & VLA R\&D platform & VLA model; data; deployment stack & training; eval; real-robot deployment & L2 adj. & \src{https://galaxea-ai.com/cn/platform} \\
\addlinespace[0.85pt]
\multirow[t]{2}{2.72cm}{\RaggedRight\textbf{TARS Robotics}} & AWE & tactile world models & tactile world model; manipulation policy & multimodal sensing; prediction; control & L2-L3 adj. & \src{https://capital.lenovo.com/news/detail/id/1092/s/1.html} \\*
\innerline
 & TacForeSight & tactile world models & tactile world model; manipulation policy & multimodal sensing; prediction; control & L2-L3 adj. & \src{https://www.webdisclosure.com/press-release/tars-etr-tars-debuts-at-waic-2026-as-its-awe-embodied-foundation-model-wins-prestigious-sail-award-169FMrhkUg1} \\
\addlinespace[0.85pt]
\multirow[t]{3}{2.72cm}{\RaggedRight\textbf{Synapx Dynamics}} & SYNWorld & embodied data / model stack & data; world model; post-training & data synthesis; training; eval; control & L2-L3 & \src{https://www.synapxdynamics.ai/news/25} \\*
\innerline
 & OctoMind & embodied data / model stack & data; world model; post-training & data synthesis; training; eval; control & L2-L3 & \src{https://www.synapxdynamics.ai/news/46} \\*
\innerline
 & OctoSense & embodied data / model stack & data; world model; post-training & data synthesis; training; eval; control & L2-L3 & \src{https://www.synapxdynamics.ai/news/46} \\
\addlinespace[0.85pt]
\multirow[t]{2}{2.72cm}{\RaggedRight\textbf{MirrOS}} & Code as Worlds & world models / test-time adaptation & world code; fast weights & environment modeling; eval; test-time update & L1-L2 & \src{https://mirros-lab.github.io/code-as-world/} \\*
\innerline
 & Spatial-TTT & world models / test-time adaptation & world code; fast weights & environment modeling; eval; test-time update & L1-L2 & \src{https://arxiv.org/abs/2608.27549} \\
\addlinespace[0.85pt]
\multirow[t]{3}{2.72cm}{\RaggedRight\textbf{Wuya Zhiyuan*}} & SHINE & continual / test-time learning & LoRA; skill vectors; weights & context adaptation; skill transfer & L1-L2 & \src{https://arxiv.org/abs/2602.06358} \\*
\innerline
 & LIFT & continual / test-time learning & LoRA; skill vectors; weights & context adaptation; skill transfer & L1-L2 & \src{https://arxiv.org/abs/2412.13626} \\*
\innerline
 & PaST & continual / test-time learning & LoRA; skill vectors; weights & context adaptation; skill transfer & L1-L2 & \src{https://arxiv.org/abs/2601.11258} \\
\addlinespace[0.85pt]
\RaggedRight\textbf{Singularity Escape / Nexus*} & Nexus & collaborative self-evolving agents & shared state; org knowledge; harness & collaboration; experience capture; harness adaptation & L3-L4 cand. & \src{https://www.iimedia.cn/c1099/113820.html} \\
\addlinespace[0.85pt]
\sectionrow{F. Persistent-memory \& personal-AI companies}{Cross-task memory, identity, or reusable skills are the main substrate of adaptation.}
\RaggedRight\textbf{Engram} & Knowledge Cartridges & persistent enterprise memory & knowledge modules; memory & extract; store; retrieve; adapt & L1-L2 adj. & \src{https://engram.com/blog/legal-agents-with-memory} \\
\addlinespace[0.85pt]
\RaggedRight\textbf{Lemon AI / Hexdo} & LemonAI Evolving & local persistent agents & workspace; memory; experience library & web / files / code; persistence; reuse & L1-L2 & \src{https://github.com/hexdocom/lemonai} \\
\addlinespace[0.85pt]
\multirow[t]{3}{2.72cm}{\RaggedRight\textbf{EverMind}} & EverOS & agent memory / OS & long-term memory; skills & retrieve; skill forge; active tasks; eval & L2-L3 & \src{https://github.com/EverMind-AI/EverOS} \\*
\innerline
 & Raven & agent memory / OS & long-term memory; skills & retrieve; skill forge; active tasks; eval & L2-L3 & \src{https://github.com/EverMind-AI/Raven} \\*
\innerline
 & EverMemOS & agent memory / OS & long-term memory; skills & retrieve; skill forge; active tasks; eval & L2-L3 & \src{https://arxiv.org/abs/2601.02163} \\
\addlinespace[0.85pt]
\multirow[t]{2}{2.72cm}{\RaggedRight\textbf{Mindverse}} & Second Me & personal memory / adaptation & identity model; memory; LoRA & memory modeling; personalization; retrieval & L1-L2 & \src{https://github.com/Mindverse/Second-Me} \\*
\innerline
 & Me.bot & personal memory / adaptation & identity model; memory; LoRA & memory modeling; personalization; retrieval & L1-L2 & \src{https://www.mindverse.com/about-us.html} \\
\addlinespace[0.85pt]

\end{longtable}
\endgroup
\vspace{-0.3em}
{\scriptsize
\textit{Reading guide.} B0 denotes task-local autonomy only. L1--L5 follow the review's increasing autonomy scale.
``adj.'' marks RSI-adjacent infrastructure or behavior; ``cand.'' marks a plausible but not fully demonstrated level;
``target'' marks an explicit future objective rather than a demonstrated current capability.
Asterisks indicate entries whose company identity or public technical boundary still requires stronger verification.
}

\end{landscape}

\clearpage

\appsection{RSI Across Applications}

Having classified systems from B0 to L5 according to the improvement functions placed under AI control, we now examine how RSI is applied across different scenarios. As illustrated in Fig.~\ref{fig:rsi-applications}, applications differ in both what they improve and the feedback required to validate and retain an update. We organize the evidence into four application regimes, including AI for science (S1), embodied intelligence (S2), software engineering (S3), and healthcare (S4). These scenarios are not mutually exclusive, as a healthcare AI scientist may fall under both S1 and S4.

Each subsection follows a common structure. We first characterize the typical AI workflow in the target scenario, identify how an RSI loop can emerge from its tasks and available feedback, and review the challenges of the RSI for a specific scenario. We then group representative systems according to the components or workflows they improve, and compare where they intervene in the RSI loop, what they update, and how these improvements are retained. Finally, we assess the gap between current research and the full RSI loop envisioned for each scenario, highlighting the strength of existing evidence, scenario-specific risks, and the key barriers to greater improvement.






\subsection{S1: RSI for Science}
\label{subsec:s1}

Science is a particularly consequential setting for RSI, as scientific progress depends not only on solving isolated problems, but also on improving the mechanisms by which future questions, hypotheses, and experiments are generated. 
Scientific RSI therefore differs from automated discovery that merely searches for a better molecule, equation, or hypothesis. A scientific RSI loop starts from a research challenge, generates candidate hypotheses and experiments, evaluates them through computation or interaction with the physical world, and uses the resulting evidence or falsification to diagnose the AI-for-science (AI4Sci) system. 

RSI for science is mainly challenged by three characteristics. 
First, scientific discovery is open-ended. Neither the space of relevant hypotheses nor the set of necessary tools and experiments is known in advance, and costly experiments make exploration itself part of the improvement problem. 
Second, scientific feedback is often sparse, delayed, and non-identifying. For example, a failed experiment may indicate an incorrect hypothesis, but it may also result from an inappropriate model, an invalid protocol, or an unreliable instrument. 
Third, scientific validity is conditional on assumptions, domains, and measurement procedures. An update that succeeds in one setting cannot be inherited safely without recording its provenance, scope of validity, and uncertainty. 
These characteristics make scientific RSI simultaneously a problem of evidence acquisition, capability evolution, and epistemic control. Accordingly, we organize existing work around three components that may evolve, including the scientific hypothesis module, the experimental agent, and the reflection system or improver.

\subsubsection{Evolving Scientific Hypothesis Modules}

The hypothesis module determines which explanations or candidate methods are proposed and prioritized. Its evolution requires feedback to change the mechanism used for future hypothesis generation, rather than merely selecting a better hypothesis for the current problem. HypoForge~\cite{qian2026hypoforge} continually distills scientific experience into reusable procedural skills for hypothesis generation, experimental design, and execution. It updates generation skills from discriminator critiques when direct empirical feedback is unavailable, and refines testing skills from real execution outcomes and attributed failures.
EvoScientist~\cite{lyu2026evoscientist} instead maintains an ideation memory of promising and failed research directions, which is retrieved in later tasks to adjust subsequent proposals. This produces persistent memory-based adaptation, although the underlying models, memory schema, and update procedure remain fixed.

A more direct approach updates the candidate generator itself. TTT-Discover~\cite{yuksekgonul2026discover} applies reinforcement learning at test time, allowing scientific feedback to modify model weights while solving a new problem. The self-improvable polymer discovery framework~\cite{khajeh2024polymer} similarly adds simulation-evaluated candidates to the training data and retrains its generative model. 
The evolution of these systems remain task- or domain-specific and are produced by fixed training and selection procedures. 
Current evidence therefore supports bounded hypothesis-level improvement, not a general mechanism that becomes progressively better at formulating scientific hypotheses across domains.

\subsubsection{Evolving Experimental Agents}

An experimental agent turns a hypothesis into an executable procedure by selecting tools, writing code, or operating laboratory equipment. Its evolution occurs when execution feedback is converted into reusable actions. Test-Time Tool Evolution~\cite{lu2026tool} synthesizes, verifies, and reuses executable scientific tools during inference, whose SciEvo benchmark contains 1,590 tasks supported by 925 evolved tools. 
CASCADE~\cite{huang2025cascade} creates, repairs, and accumulates executable skills for chemistry and materials science, while S1-NexusAgent~\cite{s1nexusagent2026framework} distills complete research trajectories into reusable Scientific Skills. 
SkillFoundry~\cite{shen2026skillfoundry} builds and continually maintains a library of reusable scientific skills by mining heterogeneous resources such as papers, repositories, documentation, and APIs, compiling them into executable skill packages, and validating, merging, or pruning these skills based on execution and downstream utility.
These methods make the agent's action space persistent and extensible, although their tool-generation and admission rules remain externally specified.

Other systems evolve broader experimental strategies and workflows. 
DrugSAGE~\cite{zhang2026drugsage} stores verified modeling procedures, training strategies, and recurring error fixes in a cross-task memory. Experience collected from 16 tasks improves performance on 17 held-out tasks, providing direct evidence that inherited experience changes behavior on new problems. 
STELLA~\cite{jin2025stella} updates reasoning templates and expands a Tool Ocean for biomedical research, EarthLink~\cite{guo2025earthlink} retains successful scripts and analytical workflows for climate science, and OriGene~\cite{zhang2025origene} refines thinking templates, tool compositions, and analytical protocols using human and experimental feedback. 
In physical simulation, the self-evolving fluid-control agent~\cite{sun2026fluidcontrol} jointly revises its agent definition and the white-box controller it produces. This case also illustrates an important evaluation boundary, where improvements to the experimental agent must be measured separately from improvements to the scientific artifact, such as the controller.

Across these systems, the strongest evidence comes from validated reuse on later tasks rather than growth in the size of a tool or memory library. Reliable experimental-agent evolution therefore requires explicit skill preconditions, execution tests, provenance, versioning, and rollback. Without these controls, a locally successful workflow may be inherited outside the conditions under which it was validated.

\subsubsection{Evolving Reflection Systems and Improvers}

The reflection system interprets evidence and identifies the cause of failure, whereas the improver decides what component to modify, how to modify it, and whether the update should be retained. A fixed critic that revises the current output executes a predefined improvement procedure, while an evolving improver must instead update its own diagnostic, intervention, or promotion policy from the outcomes of earlier improvements.

SIA~\cite{hebbar2026sia} provides the broadest component-level update loop among current AI4Sci systems. Its Feedback-Agent uses task outcomes to choose between modifying the agent scaffold, tools, retry and search logic, and low-rank adapter (LoRA) weights with different strategies. 
CORAL~\cite{qu2026coral} replaces fixed evolutionary-search heuristics with long-running asynchronous agents that autonomously inspect prior attempts, decide when to test or invoke evaluators, and externalize discoveries as shared attempts, notes, and reusable skills, with heartbeat-triggered reflection and redirection sustaining long-horizon exploration.
SAGA~\cite{du2025saga} extends the editable surface to the objective layer. Under a fixed high-level scientific goal, it diagnoses failure modes in optimized candidate populations, proposes or reweights concrete objectives, and implements them as executable scoring functions that steer subsequent search.

\subsubsection{Toward Recursively Improving Science Agents}

Taken together, these lines of work point toward a broader vision of scientific RSI. Its ultimate form is not merely a system that produces better hypotheses or experimental results, but one that becomes progressively better at conducting science with original innovation. It should jointly improve how hypotheses are generated, experiments are executed, and evidence is interpreted, such that each research cycle produces both scientific findings and reusable improvements to future discovery.

Current AI4Sci systems realize this vision only partially. B0-style output refinement is already widespread, while L1 persistent updates have been demonstrated through model retraining and memory accumulation~\cite{yuksekgonul2026discover,khajeh2024polymer,lyu2026evoscientist}. L2 represents the strongest current frontier, with systems autonomously selecting or applying updates to model weights, tools, skills, workflows, and scaffolds~\cite{hebbar2026sia,lu2026tool,huang2025cascade}. A few systems exhibit aspects of L3 through adaptive exploration and cross-task experience reuse~\cite{qu2026coral,zhang2026drugsage}, whereas end-to-end L4 environment adaptation and L5 evolution of the improver itself remain largely unexplored.

\begin{tcolorbox}[
    width=\linewidth,
    colback=gray!7,
    colframe=gray!45,
    boxrule=0.9pt,
    arc=2mm,
    outer arc=2mm,
    left=7pt,
    right=7pt,
    top=5pt,
    bottom=5pt,
    boxsep=0pt,
    before skip=5pt,
    after skip=5pt
]
\textbf{How far are we from True RSI for Science?}
Progress toward higher-level scientific RSI will mainly require more reliable attribution of experimental outcomes, more selective transfer of improvements across tasks, and stronger evaluation of whether successive system updates genuinely improve future research. Equally important is maintaining provenance, reproducibility, and stable scientific and safety constraints as increasingly powerful components of the improvement process become adaptive.
\end{tcolorbox}

\subsection{S2: RSI for Embodied Intelligence}
\label{subsec:s2}


For embodied intelligence, an agent's actions produce observable consequences in a physical or simulated environment, which provides a natural setting for RSI.
However, three main challenges remain.
First, experience is endogenous, as the current policy determines which states the agent reaches and therefore what evidence is available for further improvement. 
Second, capability is distributed across the sensorimotor system, including perception, planning, control, and the physical embodiment, making failures difficult to attribute to a single component. 
Third, physical trials cannot be freely reproduced, reversed, or reset, so candidate updates must be evaluated under safety, hardware, and resource constraints. 

Embodied RSI must therefore jointly address experience generation, cross-component credit assignment, and safe inheritance. Its interaction loop begins with environments and curricula that provide learning tasks, where the agent harness organizes relevant capabilities, the policy executes actions, and world models and evaluators interpret the resulting outcomes. Validated feedback is then written back to these components, allowing the updated system to participate in the next round of interaction. Accordingly, we review existing work on evolving environments and curricula, agent harnesses, policies and action models, and world models and evaluators.

\subsubsection{Evolving Environments and Curricula}

An embodied RSI loop begins by determining what the agent will experience and learn from. POET~\cite{wang2019poet} addresses the limitation of fixed, human-designed curricula by co-evolving a population of environment--agent pairs. It mutates environments, admits challenges that are neither trivial nor unsolvable through a minimal criterion, optimizes the corresponding policies, and periodically transfers policies across environments to reuse behavioral stepping stones. EnvGen~\cite{zala2024envgen} targets the high cost of directly deploying large language models as embodied agents and the inability of static curricula to address individual weaknesses. It instead employs an LLM as a curriculum designer, where the LLM generates simulator configurations, a smaller RL agent trains in them, and skill-level performance is returned to the LLM to produce environments targeting the agent's remaining weaknesses. 
GenEnv~\cite{guo2025genenv} further addresses the mismatch between static training data and an improving learner by parameterizing the environment simulator as a trainable curriculum policy. Its $\alpha$-Curriculum Reward favors tasks near the learner's current capability frontier, after which the learner is optimized with GRPO and the simulator is updated through reward-weighted regression.

Whereas these methods adapt tasks within an existing simulator, OMNI-EPIC~\cite{faldor2024omni} addresses the restricted search space imposed by manually predefined environment distributions. It uses LLMs to generate executable environment and reward code, while conditioning subsequent proposals on an archive of prior tasks and the current agent's learning progress to seek challenges that are both learnable and novel. 
SimWorld Studio~\cite{kang2026simworld} tackles a further obstacle, where generated three-dimensional content is often static, unverifiable, and unsuitable for embodied training. Its SimCoder agent writes executable Unreal Engine code, revises environments using compilation errors, physics checks, and vision-language-model critiques, and stores successful tools and skills for later generation. Learner performance then guides it toward progressively harder Gym-compatible worlds. Collectively, these methods convert learner performance into a mechanism for evolving the distribution and the generator of future embodied experience.


\subsubsection{Evolving Skills, Memory, and Agent Harnesses}

Given a task, the agent harness determines how memories, skills, tools, and execution rules are assembled into an actionable system. Voyager~\cite{wang2023voyager} writes successful Minecraft behaviors into an executable skill library, while LRLL~\cite{tziafas2024lrll} uses self-guided exploration and a wake--sleep process to grow and reorganize a library of composable robot programs. EmbodiSkill~\cite{ju2026embodiskill} further distinguishes between failures caused by an incorrect skill and failures caused by the agent not following valid guidance, updating a skill only when the trajectory provides relevant evidence. 

More recent systems expand the update target from individual skills to the surrounding agent architecture. SHAPER~\cite{wang2026shaper} jointly evolves reusable skills and the context-code harness through environment rollouts, while ASPIRE~\cite{lu2026aspire} diagnoses robot execution failures, validates candidate repairs, and retains successful fixes as transferable skills. ENPIRE~\cite{xiao2026enpire} goes further by allowing coding agents to revise robot policies, training procedures, and supporting infrastructure through repeated real-world experiments. 

These systems approach system-level evolution as experience changes not only what the agent knows, but also how its capabilities are selected, composed, and improved. The resulting harness ultimately shapes how the policy converts observations and instructions into actions.

\subsubsection{Evolving Policies and Action Models}

With the task and execution context in place, the policy determines how the agent acts in the environment. Self-Improving Embodied Foundation Models~\cite{ghasemipour2025selfimproving} derive reward and success signals from a pretrained model, allowing robots to practice tasks and improve their policies with limited human supervision. MEDAL++~\cite{sharma2023selfimproving} similarly supports autonomous practice by learning both how to complete and undo a task, reducing the need for manual environment resets. Other methods update more targeted components. Q-Planning~\cite{giridhar2026qplanning} keeps a large behavior-cloning policy fixed but continually trains a smaller value function from both successful and failed deployment trajectories, while SERP~\cite{li2026serp} adjusts its action model from recent navigation failures before replanning. In each case, interaction produced by the current policy becomes training evidence for its successor. 

However, the update algorithm, reward definition, and permitted model components remain specified in advance, making these systems primarily cases of policy-level self-improvement. Moreover, before such experience can be safely inherited, the system must predict and evaluate what its actions actually caused.

\subsubsection{Evolving World Models and Evaluators}

Once the policy acts, world models and evaluators turn the resulting interaction into predictions, outcome judgments, and learning signals. VLAW~\cite{guo2026vlaw} uses real-robot rollouts to improve an action-conditioned video world model, which then generates additional synthetic experience for training the vision-language-action policy. World-VLA-Loop~\cite{liu2026worldvlaloop} makes this relation iterative, where policy failures refine the world model, and the refined model provides a more reliable environment for the next round of policy optimization. Motus2~\cite{bi2026motus2} integrates policy, simulation, and evaluation functions into a shared model, using failed and suboptimal interactions to improve its dynamics and value predictions. SAIL~\cite{luo2025sail} follows a related loop in which a video planner executes its own generated plans and is fine-tuned on the resulting trajectories. 

These methods improve not only the policy but also the mechanisms that predict consequences and produce future training signals. Their feedback can then update the policy, harness, or curriculum used in the next round, closing the embodied RSI loop. Nevertheless, their objectives and update procedures remain externally fixed, so they instantiate bounded model--policy co-evolution rather than unrestricted recursive improvement.

\subsubsection{Toward Recursively Improving Embodied Agents}

Taken together, these lines of work suggest that embodied RSI should improve not only task performance but also the process through which an agent acquires and validates new capabilities. An environment generator should propose tasks near the agent's capability frontier; the harness should assemble relevant memories, skills, tools, and models; the policy should interact with the environment; and world models and evaluators should interpret the resulting outcomes. Verified improvements should then be inherited by the policy, world model, or harness, allowing each interaction cycle to improve both current behavior and future learning.

Current systems realize this loop only partially. B0-style within-episode reflection and replanning are already widespread. L1 persistent improvement has been demonstrated through policy and model updates based on embodied experience~\cite{ghasemipour2025selfimproving,sharma2023selfimproving,luo2025sail}. L2 represents a major current frontier, with systems selecting and applying updates to skills, prompts, control programs, and agent harnesses under fixed evaluation procedures~\cite{ju2026embodiskill,wang2026shaper,lu2026aspire}. Some systems exhibit L3 capabilities by generating experience according to the learner's state and retaining validated skills across tasks~\cite{wang2023voyager,tziafas2024lrll}. L4 agent--environment co-evolution has also emerged, primarily in simulation~\cite{wang2019poet,zala2024envgen,guo2025genenv,kang2026simworld}. ENPIRE~\cite{xiao2026enpire} further approaches bounded L5 by allowing coding agents to revise policies, training procedures, and supporting infrastructure. However, its environment interfaces, evaluators, and safety constraints remain externally specified. 

\begin{tcolorbox}[
    width=\linewidth,
    colback=gray!7,
    colframe=gray!45,
    boxrule=0.9pt,
    arc=2mm,
    outer arc=2mm,
    left=7pt,
    right=7pt,
    top=5pt,
    bottom=5pt,
    boxsep=0pt,
    before skip=5pt,
    after skip=5pt
]
\textbf{How far are we from True RSI for Embodied Intelligence?}
True embodied RSI requires agents to actively acquire informative experience, attribute outcomes across the full sensorimotor system, and validate updates under physical conditions that cannot be freely reset. Improvements must remain effective across tasks, environments, and embodiments, with explicit provenance, uncertainty, and rollback. The system must also learn to improve its own harness and evaluation process without weakening fixed safety constraints. Current methods demonstrate these abilities separately, but have not yet integrated them into a safe, persistent, and autonomous real-world improvement loop.
\end{tcolorbox}

\subsection{S3: RSI for Software Engineering}
\label{subsec:s3}

Software engineering provides an unusually concrete setting for studying RSI because both the product and the agent that develops it are represented as executable code. A typical loop begins when a coding agent attempts an issue, observes repository-grounded feedback such as test failures, runtime traces, or code review, and uses this evidence to modify a persistent part of itself. The modified component may be the agent's skills, workflow, or implementation, and accepted changes are reused in later tasks.

RSI for software engineering then faces three fundamental challenges. 
First, the system must continually generate software tasks that target its current capability gaps while remaining both verifiable and valuable for future learning.
Second, improvement for software engineering must assess both whether a modification is effective now and whether it increases the system's capacity for further improvement. A change may raise immediate benchmark performance yet cause subsequent search to stagnate, whereas a temporarily weaker variant may produce stronger descendants.
Third, coordinating the evolution of multiple components while preserving the outer contract (e.g., evaluation criteria, improvement objectives) remains a challenge. 
Accordingly, we organize existing work around three evolving components, including agent implementations and harnesses, software-engineering experience and collaboration, and the improvement process itself.


\subsubsection{Evolving Coding-Agent Implementations and Harnesses}

The most direct form of software-engineering RSI treats the coding agent itself as an editable software project. SICA~\cite{robeyns2025sica} allows an agent to inspect previous versions and benchmark results, modify its own Python codebase, and use the selected version in the next improvement round. 
Because the same system acts both as the software developer and as the object being developed, changes to file-editing tools, subagents, or context management can improve both ordinary coding and subsequent self-modification. 
However, its benchmark, utility function, and sandbox remain externally fixed.

More recent work narrows self-modification to the agent \emph{harness} (e.g., middleware, verification rules, and control logic). 
Self-Harness~\cite{zhang2026selfharness} uses execution traces to identify recurring weaknesses, asks the same underlying model to propose small changes to its current harness, and retains a proposal only after regression testing. 
Agentic Harness Engineering~\cite{lin2026ahe} exposes harness components as separately editable and revertible files, distills long trajectories into structured evidence, and checks whether each proposed change produces its predicted effect. Its separate evolver role makes the loop less directly self-referential than Self-Harness, but the accepted harness still persists across later executions. Ouroboros~\cite{razzhigaev2026ouroboros} extends this idea to deployment: reviewed changes to the agent's tools, context assembly, prompts, and core implementation become the runtime used for later work. 
Together, these systems show how version control, regression tests, and rollback can turn agent self-modification into a persistent and auditable process.

\subsubsection{Evolving Software-Engineering Experience, Skills, and Collaboration}

A lighter form of self-evolution changes how a coding agent uses prior development experience without rewriting its full implementation. SWE-Exp~\cite{chen2026sweexp} extracts both successful and failed issue-resolution trajectories into an experience bank and retrieves relevant lessons for later repository tasks. CODESKILL~\cite{li2026codeskill} goes beyond storing cases, where it learns a management policy that extracts procedural skills at multiple levels, updates or removes them as new trajectories arrive, and maintains a compact skill bank for future agents. 
In both cases, the persistent output is reusable development knowledge which supports cross-task adaptation. However, their fixed extraction and update procedures do not themselves become better through use.

Self-evolution can also change the organization of a software team. EvoMAC~\cite{hu2025evomac} represents a multi-agent coding workflow as a network of role prompts and communication links, then uses test feedback and textual back-propagation to revise both the agents and their connections. 
This improves the workflow used to construct the current program, but the reported updates occur during test time for each task.

\subsubsection{Evolving the Improvement Process}

The strongest systems do not follow a single sequence of accepted edits; they also search over alternative evolutionary paths. Darwin G\"odel Machine (DGM)~\cite{zhang2026dgm} maintains an archive of coding-agent variants, selects previous agents as parents, lets them modify their own implementations, and evaluates their descendants on executable coding tasks. Keeping multiple lineages prevents one harmful edit from permanently blocking progress and allows temporarily weak variants to become useful stepping stones. Huxley--G\"odel Machine (HGM)~\cite{wang2026hgm} further argues that current benchmark performance is not the same as the ability to produce better descendants. It therefore estimates \emph{metaproductivity} from an agent's descendant lineage and allocates evaluation effort toward agents with stronger long-term improvement potential.

Population-based methods broaden what can be inherited across lineages. Group-Evolving Agents~\cite{weng2026gea} allows several parent agents to share successful modifications and failure experience when producing a new group, so useful discoveries are not isolated within one branch. DarwinX~\cite{zhang2026darwinx} evolves populations of complete harnesses around a frozen model and admits variants only when they extend task coverage without regressing previously solved cases. HELIX~\cite{fan2026helix} proposes a further model--harness loop in which harness evolution improves current execution and generates verified trajectories for updating the model, after which the harness is rebuilt for the changed model.

\subsubsection{Toward Recursively Improving Software-Engineering Agents}

Taken together, these works suggest a software-engineering RSI loop of \emph{Development Challenge $\rightarrow$ Agent-Level Modification $\rightarrow$ Repository-Grounded Trial $\rightarrow$ Regression-Aware Selection $\rightarrow$ Versioned Inheritance}. Repository feedback triggers a persistent agent update, which is evaluated on current and held-out tasks before being inherited as a traceable and reversible version for future development and self-improvement.

Current systems realize this loop only partially. B0 task-local refinement is widespread, while L1 persistent updates appear in cross-task experience and skill learning~\cite{chen2026sweexp,li2026codeskill}. L2 is the strongest established level, with agents selecting changes to harnesses, tools, or workflows under fixed objectives and evaluators~\cite{zhang2026selfharness,lin2026ahe,zhang2026darwinx}. L3 is emerging through learner-conditioned task generation in Self-play SWE-RL and Socratic-SWE~\cite{wei2025selfplayswerl,xiao2026socraticswe}. L4 environment adaptation remains largely absent. SICA and DGM exhibit bounded L5 characteristics by modifying code that shapes later self-improvement~\cite{robeyns2025sica,zhang2026dgm}, while HGM introduces lineage-based meta-selection without adapting the selection rule itself~\cite{wang2026hgm}. Full L5 improvement of the software-development improver has not yet been demonstrated.

\begin{tcolorbox}[
   width=\linewidth,
   colback=gray!7,
   colframe=gray!45,
   boxrule=0.9pt,
   arc=2mm,
   outer arc=2mm,
   left=7pt,
   right=7pt,
   top=5pt,
   bottom=5pt,
   boxsep=0pt,
   before skip=5pt,
   after skip=5pt
]
\textbf{How far are we from True RSI for Software Engineering?}
True software-engineering RSI requires robust specifications, attributable improvements, and reliable transfer across repositories and software versions. As test generation and improvement strategies become adaptive, user intent, safety constraints, auditability, and rollback must remain externally protected.
\end{tcolorbox}


\subsection{S4: RSI for Healthcare}
\label{subsec:s4}



As one of the most prominent scenarios for AI application, healthcare spans tasks like diagnosis, medical image analysis, and treatment planning. A typical RSI loop begins with an agent performing a clinical task and receiving verifiable feedback from diagnostic results, clinician corrections, or clinical guidelines. The agent then persistently updates its components (e.g., clinical knowledge, reasoning strategies, workflows) with improvements retained in memory, skill libraries, or model parameters for reuse across subsequent cases. 

RSI for healthcare is mainly challenged by three characteristics. First, healthcare does not permit unrestricted trial and error, as diagnostic and treatment actions may harm patients and therefore require expert, ethical, and institutional oversight. Second, clinical feedback is often delayed, heterogeneous, and confounded, making it difficult to attribute patient outcomes to a particular agent decision or update. Third, clinical improvements are population- and institution-dependent, so an update validated in one setting cannot be safely inherited without recording its provenance, scope, and uncertainty. These characteristics make healthcare RSI a problem of safe feedback acquisition, reliable credit assignment, and governed capability inheritance. Accordingly, we organize existing work around three components that may evolve: clinical memory and knowledge, clinical reasoning strategies, and healthcare tools and workflows.

\subsubsection{Evolving Clinical Memory and Knowledge}

The most common form of RSI in healthcare evolves the agent's clinical memory, where the feedback from completed cases is converted into persistent knowledge that can influence decisions for subsequent patients. 
Early systems primarily retain case-level experience. MedAgent-Zero in Agent Hospital~\cite{li2024agenthospital} stores successful treatment cases and reflects on failed cases to derive reusable diagnostic rules, allowing a doctor agent with frozen model weights to retrieve these experiences in later simulated encounters. 
Similarly, Agent Mental Clinic~\cite{lan2024agentmentalclinic} compares the psychiatrist agent's diagnosis with labeled outcomes through a supervisor module and writes the resulting diagnostic experience into a tiered memory architecture. 
MedAgentSim~\cite{almansoori2025medagentsim} further combines medical records with experience records, enabling diagnostic agents to reuse knowledge accumulated from earlier simulated doctor--patient interactions. 
In these systems, the output of one encounter becomes part of the persistent context from which the agent reasons about future cases.

More recent methods evolve the structure and reliability of remembered knowledge rather than simply accumulating raw trajectories. DxEvolve~\cite{ren2026dxevolve} distills diagnostic encounters into \emph{diagnostic cognition primitives}, which encode reusable patterns for information acquisition and diagnostic reasoning and can be retrieved in new cases. GSEM~\cite{han2026gsem} organizes experience as a dual-layer graph and uses subsequent feedback to recalibrate node quality and inter-experience edge weights, thereby changing both what is remembered and which memories are considered applicable. Evo-MedAgent~\cite{shen2026evomedagent} maintains complementary stores for retrospective clinical episodes, procedural heuristics, and tool reliability. After each chest-radiography case, reflection updates these stores so that later cases inherit both clinical lessons and revised trust in diagnostic tools. 
Nevertheless, most evidence remains based on simulated or retrospective cases with externally supplied labels, and the validation of evolution in real‑world cases is still lacking.

\subsubsection{Evolving Clinical Reasoning Strategies}

Beyond evolving clinical facts, healthcare agents can evolve the strategies that determine how evidence is acquired, interpreted, and integrated. EvoClinician~\cite{he2026evoclinician} instantiates this process through a \emph{Diagnose--Grade--Evolve} loop, where an Actor sequentially asks questions and orders examinations, a Process Grader evaluates each action according to its clinical yield and resource cost, and an Evolver uses this feedback to revise the Actor's prompt and memory before the next case. The inherited evolution object is thus a diagnostic policy, for example, which symptom to clarify or which examination to prioritize. 

Reasoning evolution can also occur at multi-agent coordination level. EvoMDT~\cite{liu2026evomdt} decomposes oncology decision-making among role-specialized agents and uses expert assessments and outcome-oriented signals to update prompts, consensus weights, and retrieval scope. Thus, feedback can change not only an individual agent's reasoning but also how specialist opinions are weighted and reconciled in later cases. In psychological counseling, PsychAgent~\cite{yang2026psychagent} extracts practice-grounded skills from historical counseling trajectories, adds them to an evolving skill repertoire, and further internalizes selected skills through rejection fine-tuning. 

\vspace{-1em}
\subsubsection{Evolving Healthcare Tools and Workflows}

A third line of work evolves the agent's action repertoire by turning successful executions into reusable tools, skills, or clinical workflows. MACRO~\cite{fan2026macro} begins with a fixed collection of medical-imaging tools but identifies recurring multi-step patterns in verified execution trajectories, synthesizes them into composite tools, and registers these composites as new high-level actions for subsequent imaging tasks. The resulting agent therefore changes what it can directly invoke, rather than merely retrieving an earlier solution. SkeMex~\cite{sun2026skemex} follows a related approach at the level of procedural skills: it distills informative clinical trajectories into general, task-specific, and action-level skills, estimates their context-dependent utility from environmental feedback, and manages them through a \emph{Read--Write--Assess--Govern} lifecycle that can promote, merge, update, or remove entries. In both cases, inheritance operates through an evolving action space whose components are selected according to their demonstrated utility in later tasks.

Other systems evolve larger analytical workflows and their supporting components. TissueLab~\cite{li2025tissuelab} allows domain experts to inspect intermediate medical-imaging results and provide corrections that guide active learning, classifier refinement, and subsequent workflow construction across pathology, radiology, and spatial-omics tasks. HealthFlow~\cite{zhu2026healthflow} converts completed EHR analyses into persistent safeguards, reusable workflows, dataset-specific anchors, and code snippets; an evaluator first identifies execution or methodological defects, after which a reflector validates, updates, or retires experience before it is retrieved for later analyses. 
These systems move healthcare RSI from selecting predefined tools to constructing reusable procedures. However, because technical success or case-specific clinician feedback does not guarantee safety across patient populations, evolved workflows must be validated and reversible before being inherited across cases.

\vspace{-1em}
\subsubsection{Toward Recursively Improving Healthcare Agents}

The ultimate form of RSI for healthcare is not merely a system that produces better diagnoses, recommendations, or analytical outputs, but one that becomes progressively better at performing healthcare tasks while preserving clinical validity and safety. Each episode should produce both a task outcome and a validated improvement to the agent's memory, reasoning strategy, or action repertoire, such that the resulting capability can be safely reused and further evaluated in subsequent cases. 

B0-style within-case reflection and answer refinement are already widespread, but they do not persistently alter future behavior. L1 persistent inheritance has been demonstrated through case memories, diagnostic rules, and reusable clinical knowledge~\cite{lan2024agentmentalclinic,almansoori2025medagentsim,ren2026dxevolve}. L2 represents the strongest current frontier: systems can diagnose failures and autonomously select or apply updates to prompts, memories, reasoning policies, tools, workflows, and coordination mechanisms while their objectives, evaluators, and deployment gates remain externally fixed~\cite{he2026evoclinician,han2026gsem,shen2026evomedagent,liu2026evomdt,fan2026macro,sun2026skemex,zhu2026healthflow}. A few systems exhibit L3-like elements by using current uncertainty or accumulated experience to shape subsequent evidence acquisition, simulated encounters, or expert annotation~\cite{li2024agenthospital,li2025tissuelab}; however, their patient populations, task distributions, and admission criteria remain largely externally specified. EvoPatient further explores L4-style adaptation by co-evolving simulated patients and doctors~\cite{du2024evopatient}, but the simulator's clinical realism and evaluation criteria remain externally determined. End-to-end L4 adaptation from real longitudinal outcomes and L5 evolution of the clinical improver itself remain largely unexplored.

\begin{tcolorbox}[
   width=\linewidth,
   colback=gray!7,
   colframe=gray!45,
   boxrule=0.9pt,
   arc=2mm,
   outer arc=2mm,
   left=7pt,
   right=7pt,
   top=5pt,
   bottom=5pt,
   boxsep=0pt,
   before skip=5pt,
   after skip=5pt
]
\textbf{How far are we from True RSI for Healthcare?}
Progress toward higher-level healthcare RSI will mainly require reliable attribution of longitudinal patient outcomes, safe acquisition of improvement feedback without unrestricted clinical exploration, and selective transfer of updates across populations and institutions. Equally important is a governed capability-commitment process in which every persistent update is clinically validated, prospectively monitored, traceable to its supporting evidence, and reversible when its assumptions or safety guarantees no longer hold.
\end{tcolorbox}

\end{document}